%% file: main.tex
\documentclass{article}

\usepackage[preprint]{neurips_2026}

\usepackage[utf8]{inputenc} 
\usepackage[T1]{fontenc}    
\usepackage{hyperref}       
\usepackage{url}            
\usepackage{booktabs}       
\usepackage{amsfonts}       
\usepackage{nicefrac}       
\usepackage{microtype}      
\usepackage{xcolor}         

\usepackage{graphicx}
\usepackage{subcaption}
\usepackage{booktabs} 
\usepackage{makecell} 
\usepackage{pifont} 
\usepackage{tabularx}
\usepackage{booktabs}
\usepackage{multirow}
\usepackage{array}
\usepackage[export]{adjustbox} 
\usepackage{comment}
\usepackage{nicematrix}
\usepackage[capitalise]{cleveref}

\definecolor{royalblue}{rgb}{0.2549,0.4118,0.8824}

\hypersetup{
    colorlinks=true,  %
    citecolor=royalblue,
    linkcolor=royalblue,
    filecolor=royalblue,
    urlcolor=royalblue,
}

\renewcommand{\tabularxcolumn}[1]{m{#1}}%

\newcolumntype{S}{>{\centering\arraybackslash\scriptsize}X}

\newcolumntype{C}{>{\centering\arraybackslash}X}

\input{defs}

\title{The \acron~Dataset: Bridging the Data Gap for Open-Vocabulary Object Counting}

\author{%
  Corentin Dumery\thanks{Equal contribution.} \\
  EPFL \\
  \And
  Niki Amini-Naieni\footnotemark[1] \\
  University of Oxford \\
  \And
  Shervin Naini \\
  Northwestern University \\
  \And
  Pascal Fua \\
  EPFL \\
}

\begin{document}

\maketitle

\input{figs/teaser}

\input{sec/0_abstract}

\input{sec/1_intro}

\input{sec/2_rw}

\input{sec/3_dataset}

\input{sec/4_experiments}
\input{sec/5_conclusion}

\bibliographystyle{plain}
\bibliography{main}


\newpage
\input{sec/6_supplementary}

\end{document}

%% file: defs.tex
\newif\ifdraft
\draftfalse

\definecolor{burntorange}{rgb}{0.8, 0.33, 0.0}
\definecolor{orange}{rgb}{1,0.5,0}
\definecolor{green0}{rgb}{0.1,0.7,0.1}

\ifdraft
\newcommand{\PF}[1]{{\color{red}{\bf PF: #1}}}

\newcommand{\CD}[1]{{\color{blue}{\bf CD: #1}}}

\newcommand{\NA}[1]{{\color{green0}{\bf NA: #1}}}

\newcommand{\NE}[1]{{\color{violet}{\bf NE: #1}}}

\newcommand{\RL}[1]{{\color{orange}{\bf RL: #1}}}

\newcommand{\HL}[1]{{\color{burntorange}{\bf HL: #1}}}

\newcommand{\JX}[1]{{\color{brown}{\bf JX: #1}}}

\newcommand{\TODO}[1]{\textbf{\color{red}[TODO: #1]}}
\else
\newcommand{\PF}[1]{{\color{red}{}}}

\newcommand{\CD}[1]{{\color{blue}{}}}

\newcommand{\NA}[1]{{\color{green}{}}}

\newcommand{\NE}[1]{{\color{violet}{}}}

\newcommand{\RL}[1]{{\color{orange}{}}}

\newcommand{\HL}[1]{{\color{pink}{}}}

\newcommand{\JX}[1]{{\color{brown}{}}}

 \newcommand{\TODO}[1]{}
 
\fi

\newcommand{\acron}{MixCount}

\definecolor{darkgreen}{RGB}{0,100,0}
\definecolor{darkred}{RGB}{140,0,0}

\newcommand{\cmark}{%
    {\color{darkgreen}\checkmark}
}
\newcommand{\xmark}{%
    {\color{darkred}\ding{55}}
}

%% file: figs/teaser.tex
\begin{figure*}[htbp]
    \centering
    \begin{subfigure}{\textwidth}
        \centering
        \includegraphics[width=0.197\textwidth]{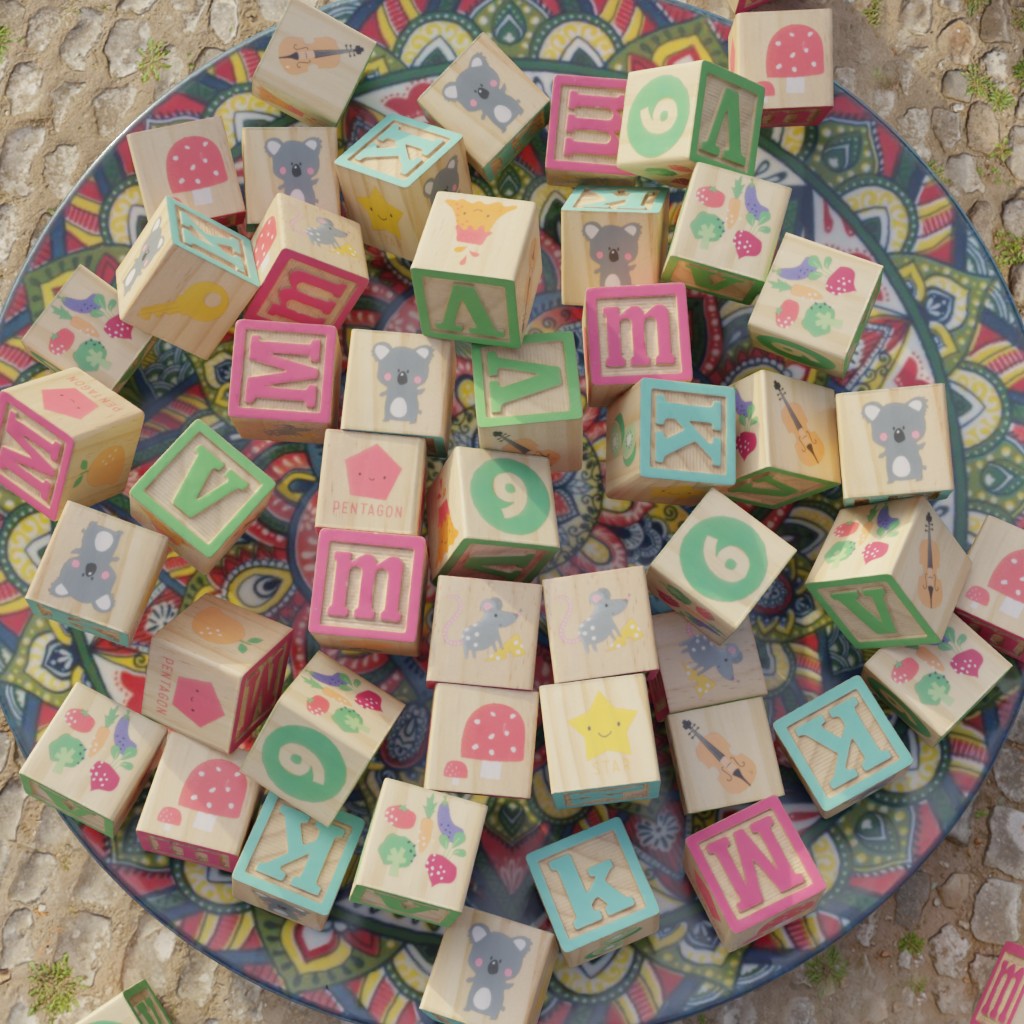}\hfill %
        \includegraphics[width=0.197\textwidth]{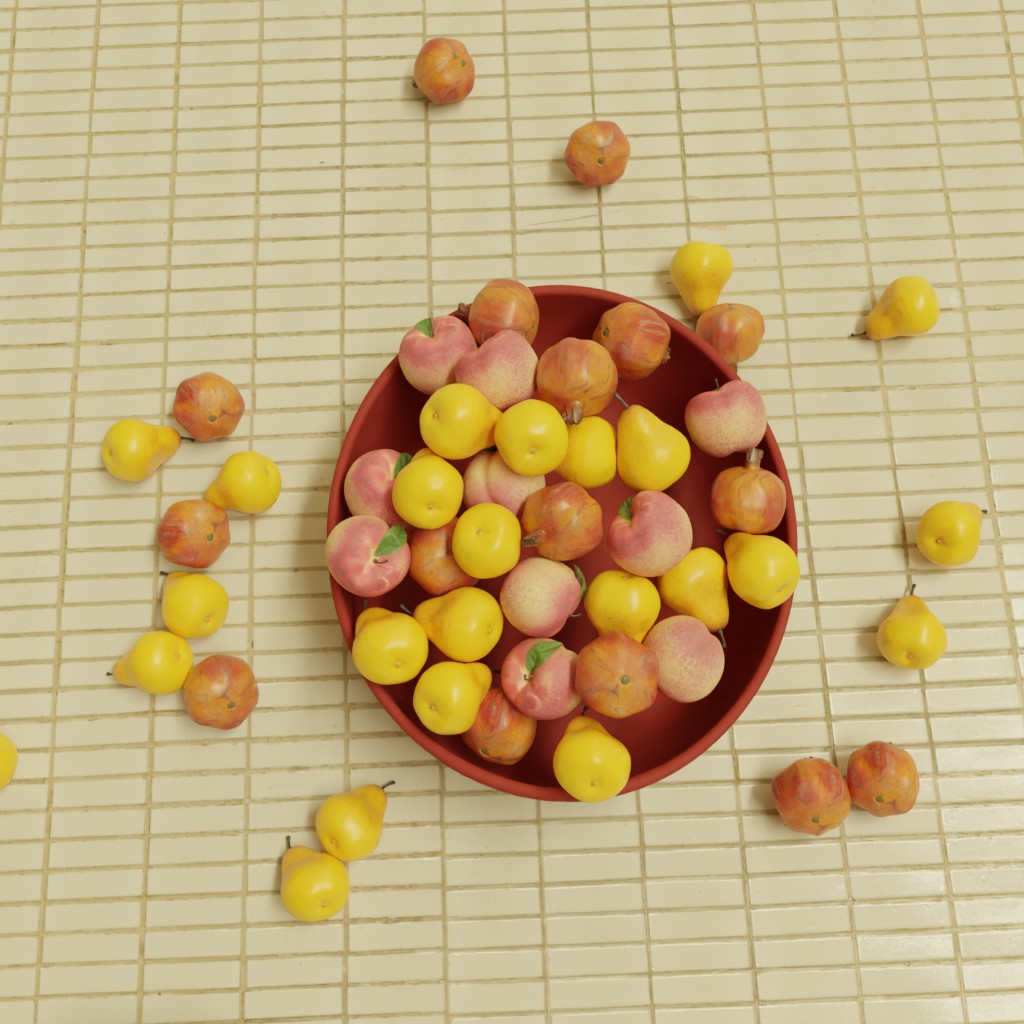}\hfill %
        \includegraphics[width=0.197\textwidth]{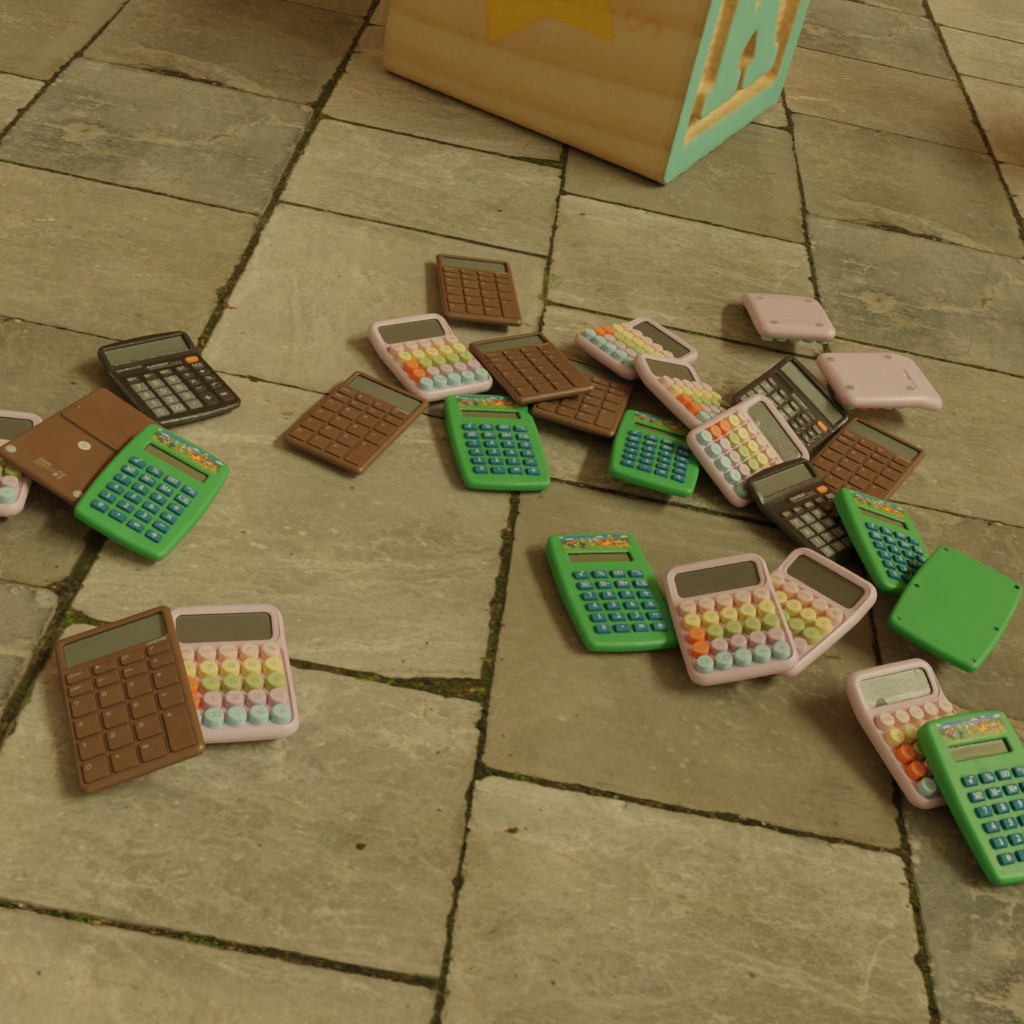}\hfill %
        \includegraphics[width=0.197\textwidth]{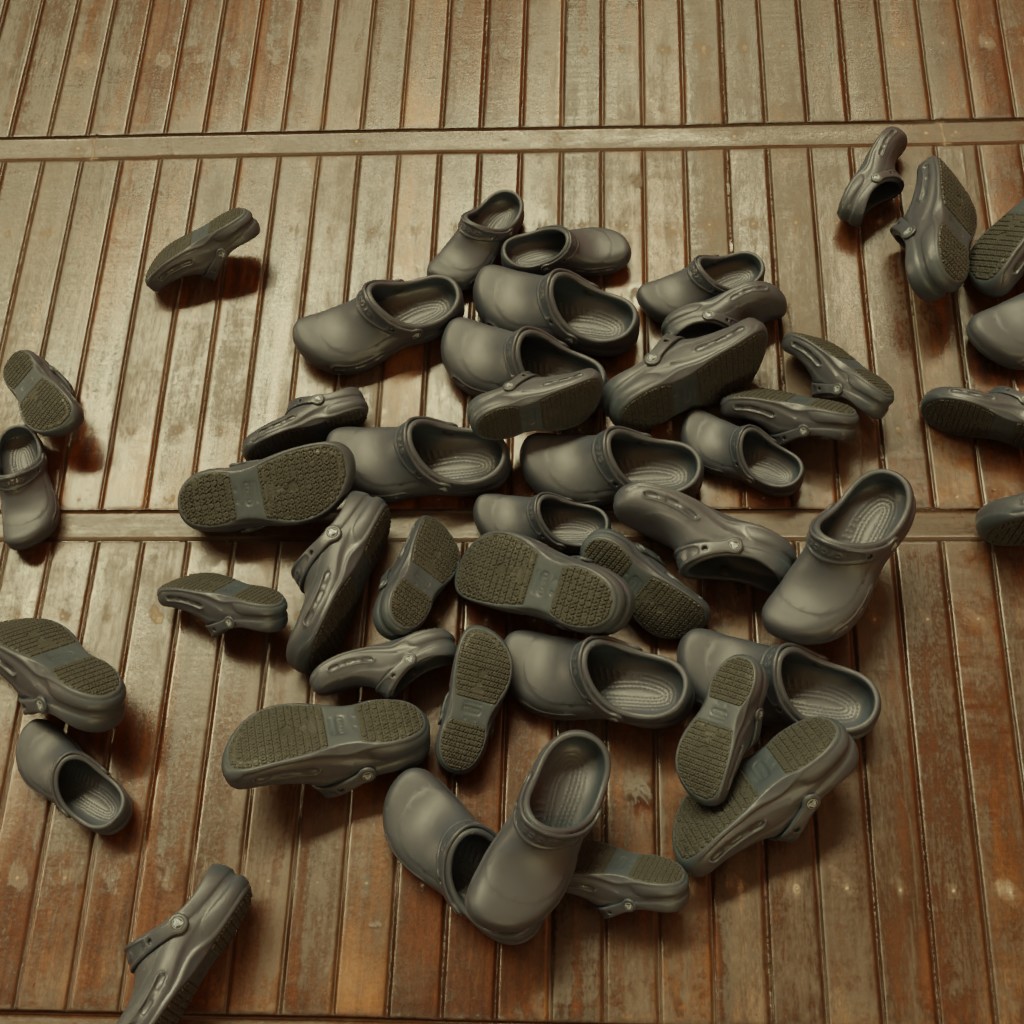}\hfill %
        \includegraphics[width=0.197\textwidth]{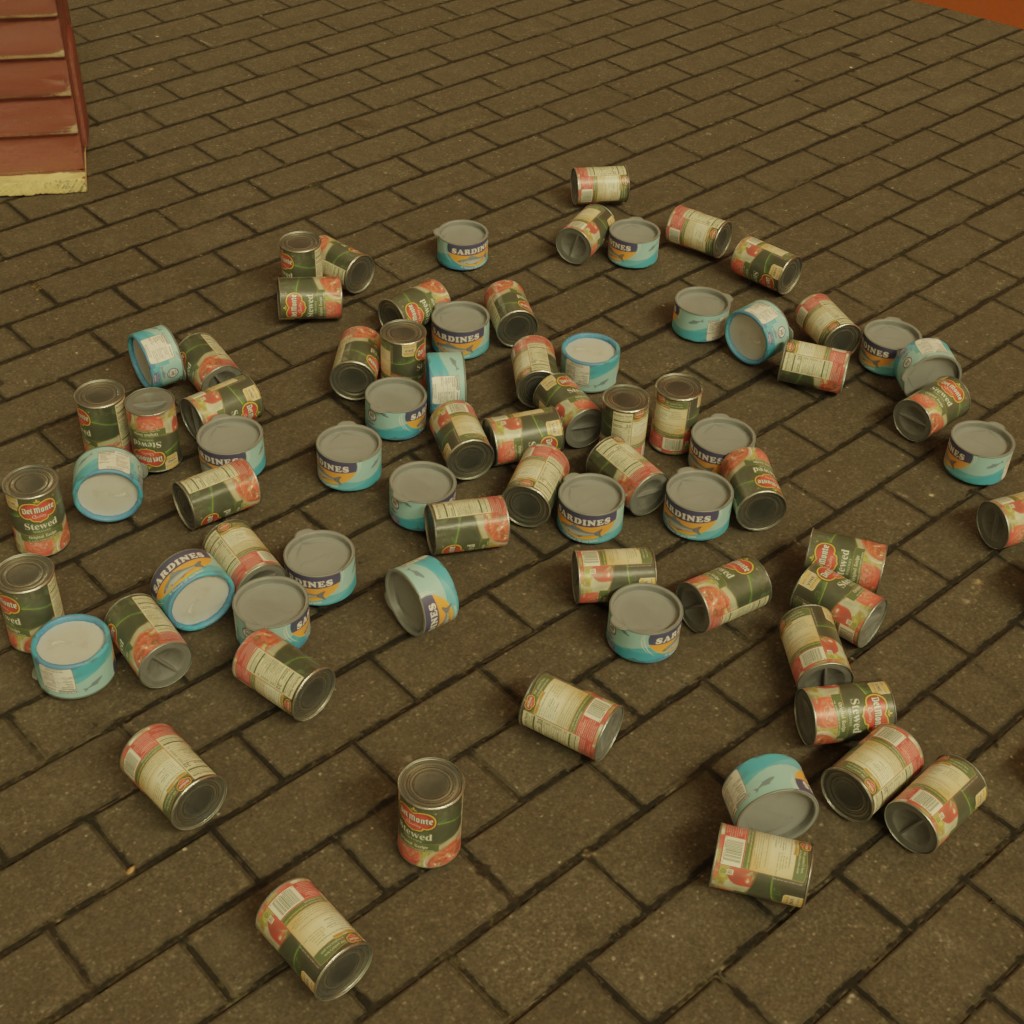}
    \end{subfigure}
    
    \vspace{0.06em} 
    
    \begin{subfigure}{\textwidth}
        \centering
        \includegraphics[width=0.197\textwidth]{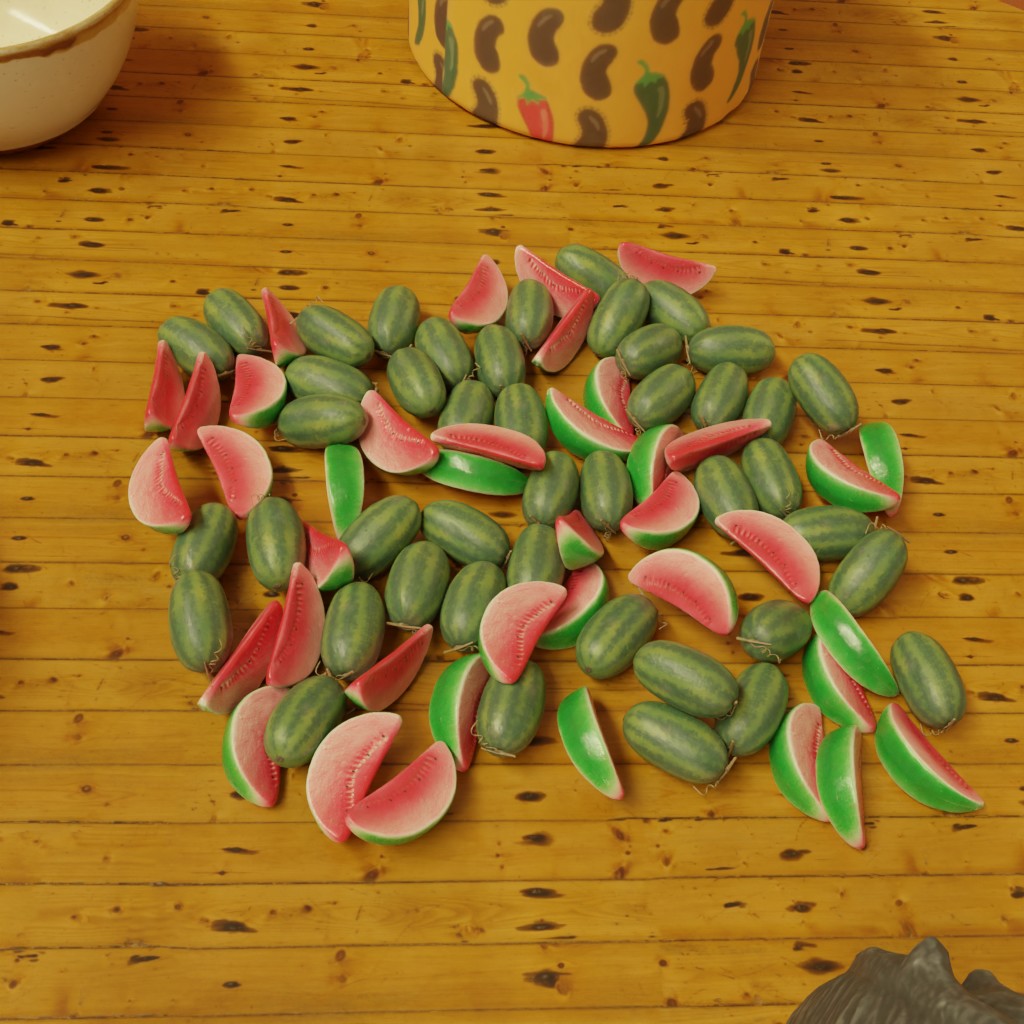}\hfill %
        \includegraphics[width=0.197\textwidth]{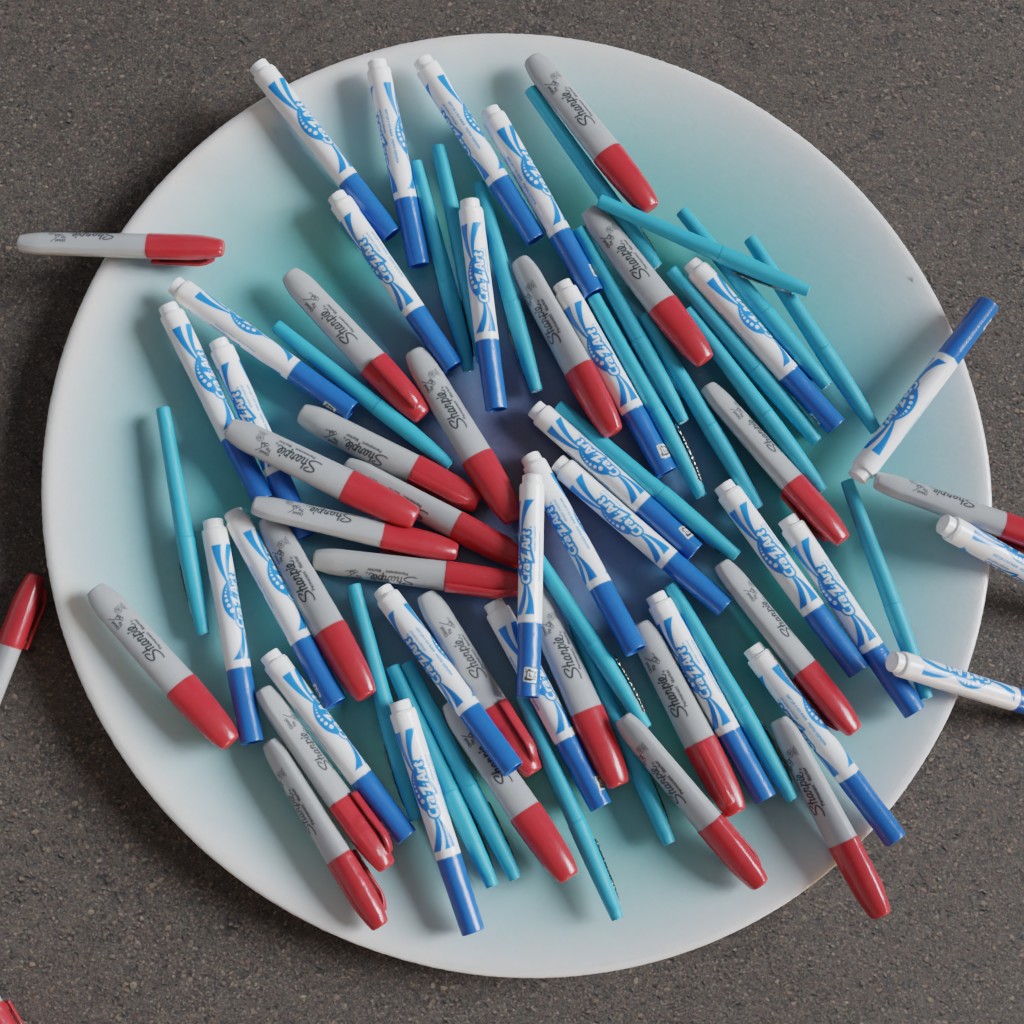}\hfill %
        \includegraphics[width=0.197\textwidth]{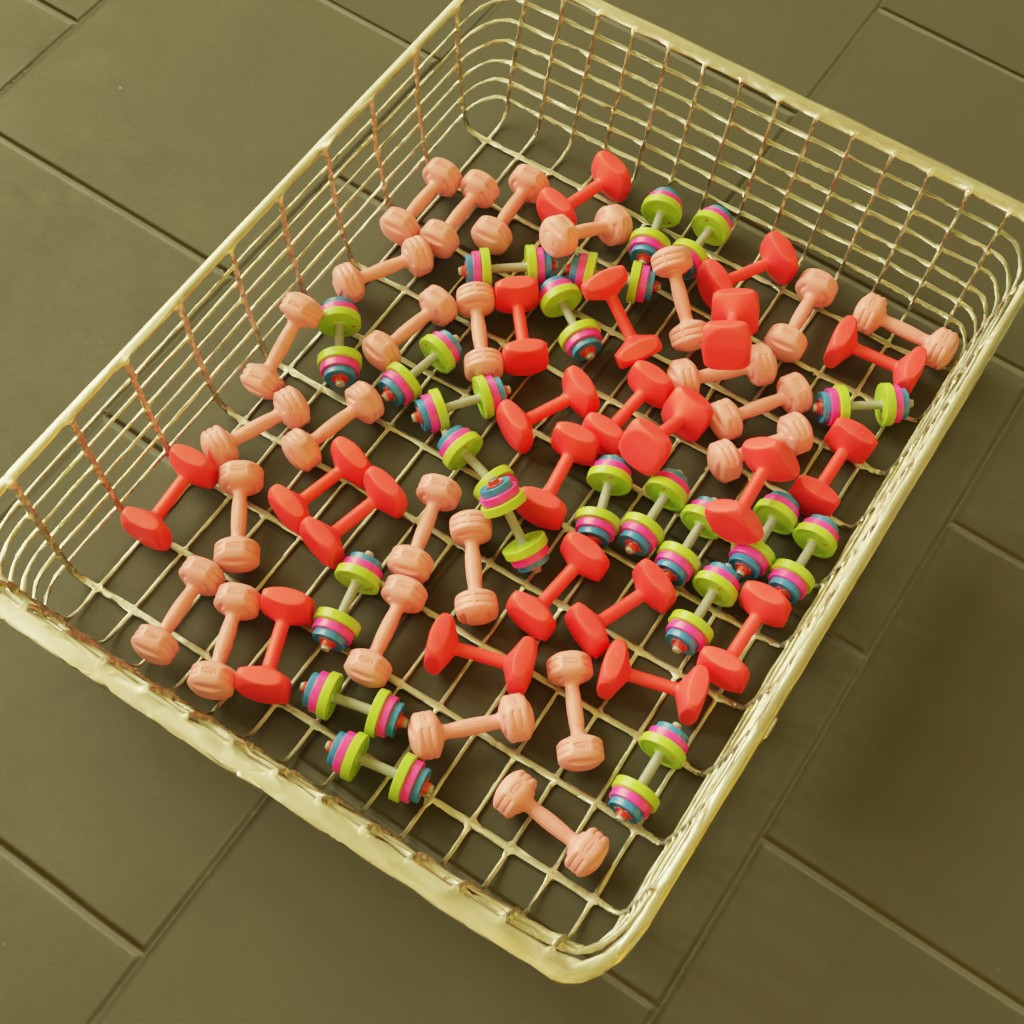}\hfill 
        \includegraphics[width=0.197\textwidth]{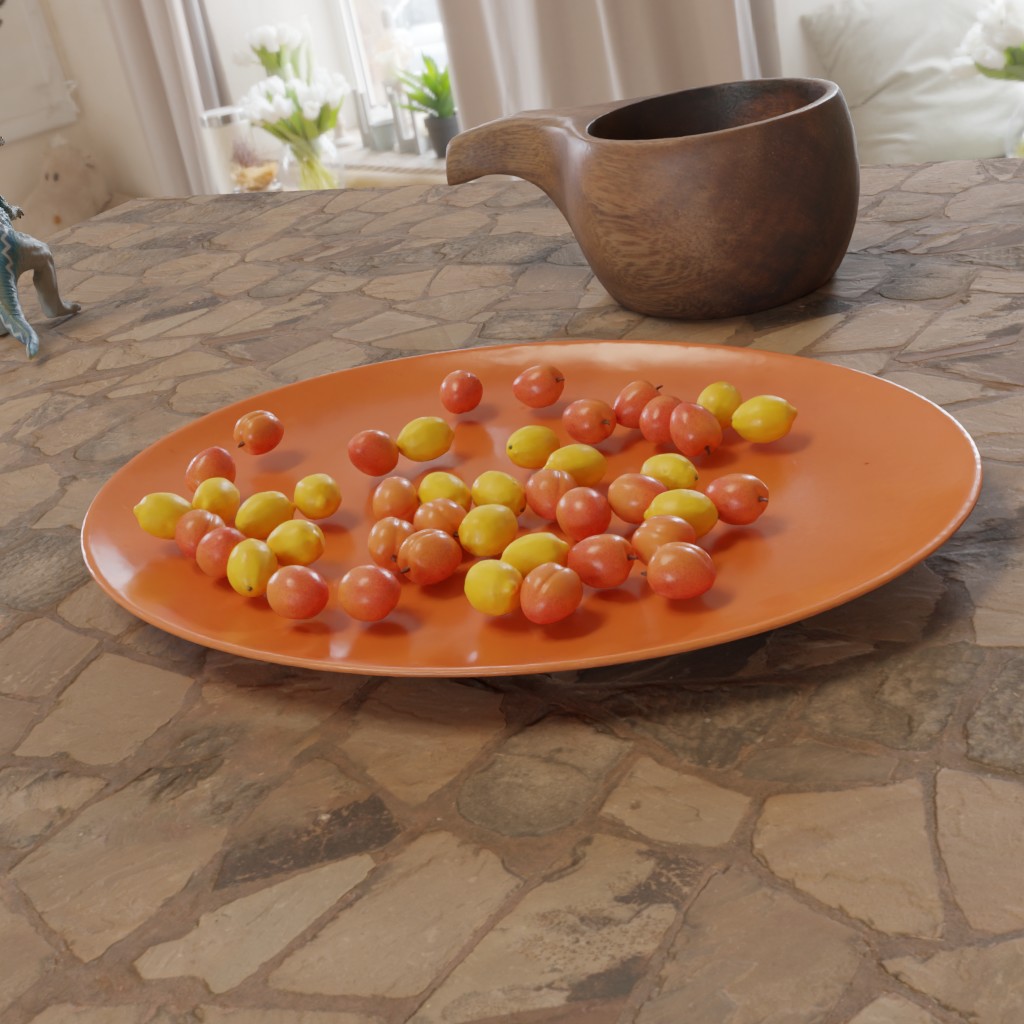}\hfill 
        \includegraphics[width=0.197\textwidth]{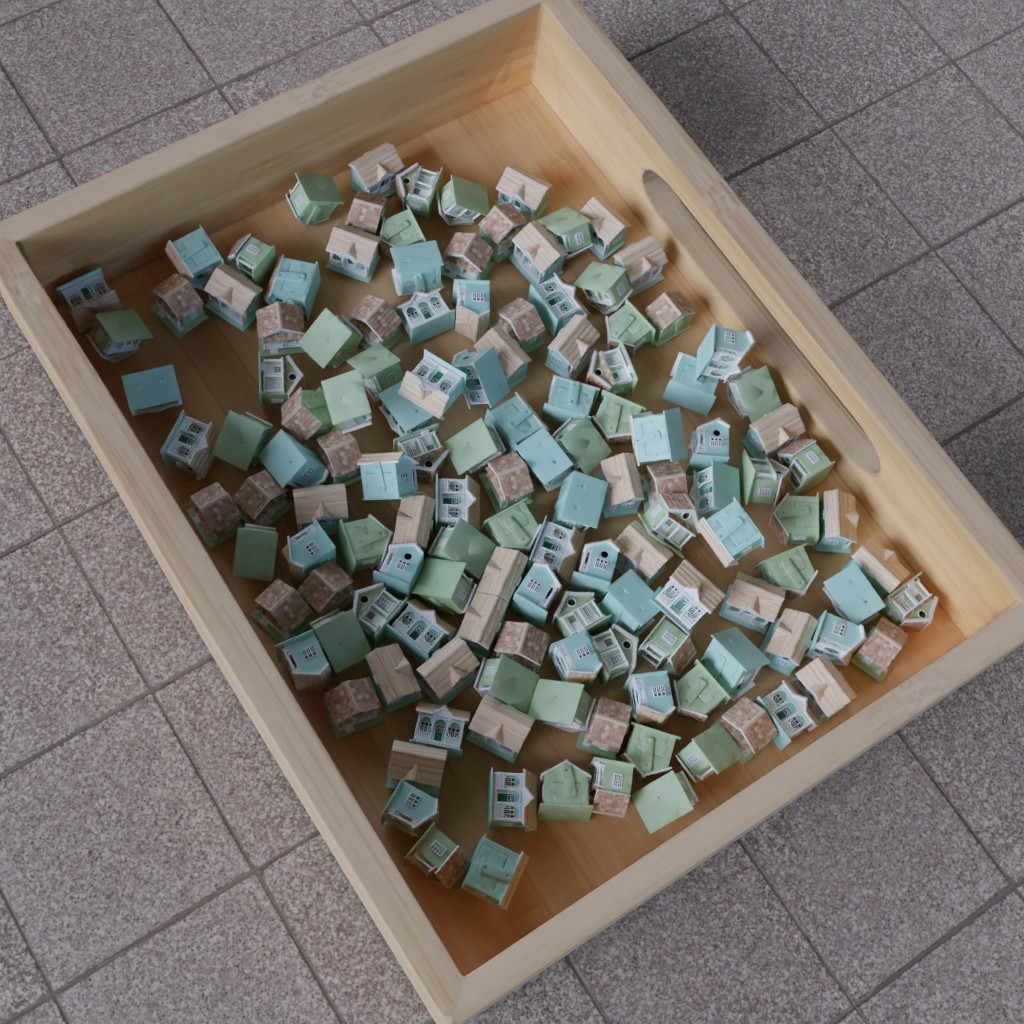}
    \end{subfigure}
    \caption{\textbf{The \acron~Dataset.} Our dataset includes images of mixed objects in settings that are challenging for visual counting models, along with precise and rich ground truth annotations, text descriptions and visual exemplars. }
    \label{fig:teaser}
\end{figure*}

%% file: sec/0_abstract.tex
\begin{abstract}

Object counting is a foundational vision task with over a decade of dedicated research, yet state-of-the-art models still fail systematically in the mixed-object setting that dominates real-world applications such as industrial inspection and product sorting. We show that this gap is strongly driven by limitations in existing training and evaluation data: real counting datasets are prohibitively expensive to annotate and suffer from labeling noise, while existing synthetic alternatives lack 
diversity and realism. We address this with \acron, a dataset and benchmark for mixed-object counting designed to target the failure modes of current counting models. To overcome the high cost of constructing and labeling such data, we develop an automatic generation pipeline that synthesizes images, fine-grained textual descriptions, and pixel-perfect counting annotations at scale, eliminating the labeling ambiguity that plagues prior datasets. Evaluating state-of-the-art counting models on \acron~exposes severe degradation in the mixed-object setting. More importantly, training these models on our synthesized data yields substantial gains on real-world benchmarks, reducing MAE by 20.14\% on FSC-147 
and by 18.3\% on PairTally
. These results establish \acron~as both a benchmark and a training dataset for fine-grained counting, and demonstrate that our pipeline, which produces effectively unlimited labeled data, 
helps address a long-standing bottleneck in 
counting models.
The \acron~dataset is available at \url{https://corentindumery.github.io/projects/mixcount.html}.


\end{abstract}

%% file: sec/1_intro.tex
\section{Introduction}

\input{figs/data_gap}

Visual object counting has been a fundamental task in computer vision for several decades. Many real-world applications ranging from cell-counting~\cite{Flaccavento11,phelan1997cell,xie2018microscopy}, crowd-counting~\cite{zhang2015cross,ranjan2018iterative,liu2019context}, wildlife monitoring~\cite{arteta2016counting}, or industrial inspection~\cite{khule2015automated,dumery2026automated} heavily rely on automated counting techniques to save significant time and enable further data analysis.
In recent years, the capabilities of counting models have greatly expanded.
General models that can count \textit{anything}~\cite{ranjan2021learning} have been proposed. They can be prompted to count a specific class of objects via an \textit{exemplar}, which is simply an image of a single instance of that object. Recently, counting models have become even more accessible with models for which the object to be counted is specified via a text prompt~\cite{amini2023open, jiang2023clip}.
These developments have made automated visual counting even more attractive and greatly broadened its potential.

However, we observed that recent models such as CountSE~\cite{liu2025countse} and CountGD~\cite{amini2024countgd}, that perform very well when objects are easily distinguishable and separated, struggle significantly in the mixed object setting where objects with subtle visual differences need to be distinguished. We hypothesize that the cause lies principally in the training data. The standard FSC-147 dataset~\cite{ranjan2021learning} for training state-of-the-art counting models lacks images with multiple types of objects annotated. 
In addition to being severely limited in both scale and diversity, real datasets also suffer from unreliable supervision as the ground-truth is manually created by human annotators. Furthermore, existing synthetic datasets~\cite{peinl2026situate, hobley2024abc} do not have the same scale limitation, but lack diversity due to their simplistic design. This results in an important domain gap between training samples and real-world images, limiting the generalization of the downstream model. 
Finally, attempting to construct training data by prompting a diffusion model to generate a predefined number of objects is ineffective due to their inability to synthesize images with a controlled count~\cite{kajic2024,d2025afreeca}. 


In parallel, 3D reconstruction has made significant advances in recent years~\cite{mildenhall2021nerf,dong2025digital,kerbl20233d}, leading to increasingly realistic and widespread real-world captures of objects~\cite{dong2025digital}, materials~\cite{zaal2021polyhaven,eppel2024vastextures}, and lighting environments~\cite{gardner2017learning,zaal2021polyhaven}. These advancements make it possible to generate arbitrary numbers of synthetic scenes with the photorealism of real data, since the components of the scenes are issued from real-world captures themselves. This has been successfully exploited in recent datasets such as Kubric~\cite{greff2022kubric} for video tracking, HyperSim~\cite{roberts2021hypersim} for 3D understanding, or OmniObject3D~\cite{wu2023omniobject3d} for 3D generation. In this work, we aim to construct the first equivalent data generator for visual counting.

Therefore, our proposed solution to bridging the data gap is a complete data generation framework that takes as input real-world captures and outputs photorealistic training samples that can be used to train a counting model. We design our generator to produce training data matching the current failure modes highlighted in \cref{fig:data_gap}, including repetitive backgrounds, visually-similar and self-similar objects, or size variations. The produced samples have pixel-perfect ground-truth, as well as unprecedented scale and diversity. We also expand the capabilities of our generator to provide several possible exemplars and text prompts for each object, making the downstream counting models more robust to the input prompt.

Finally, we release \acron, the first dataset generated with this framework. \acron~includes 58,000 counting samples featuring 1522 different objects. In our experiments, training state-of-the-art counting models on \acron~led to a substantial reduction in model prediction error of around $-20\%$ across modern benchmarks. \acron~features three levels of text descriptions for each object in all images, as well as several possible exemplars, enabling future works on prompt sensitivity and flexibility. We then employ the test set of \acron~to perform a rigorous benchmarking of modern counting models, revealing their limitations and highlighting the need for improved training datasets.

In summary, our contributions are:
\begin{itemize}
    \item The \acron~dataset, a large-scale, photorealistic, and diverse dataset to train and evaluate visual counting models, 
    \item A data generation framework that synthesizes virtually infinite counting data for fine-grained object counting and can easily be tailored to specific applications,
    \item We demonstrate that models trained on \acron~achieve superior performance on real-world datasets, with 18.3\% and 20.14\% error reductions in the challenging PairTally~\cite{nguyen2025can} and FSC-147~\cite{ranjan2021learning} benchmarks, respectively,
    \item A rigorous evaluation of current state-of-the-art visual counting methods, exposing significant performance degradation in the mixed-objects setting.
\end{itemize}

%% file: figs/data_gap.tex
\begin{figure}[t]
    \centering
    \begin{minipage}{0.37\textwidth}
        \vspace{2.2em}
        \centering
        \includegraphics[width=\linewidth]{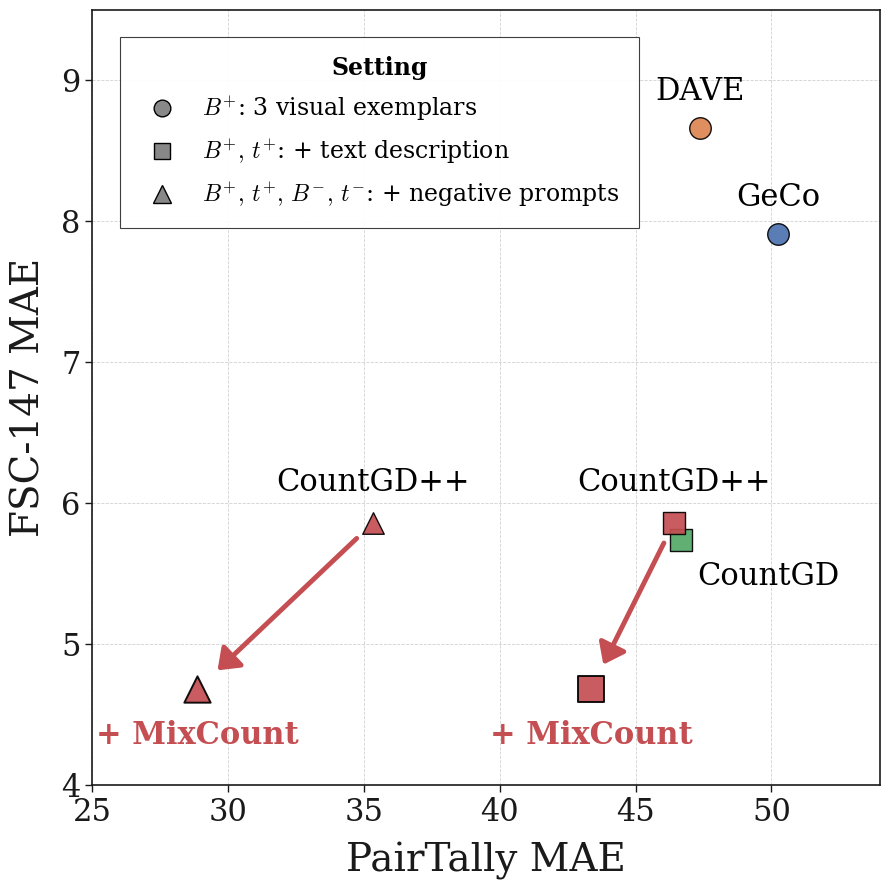}
    \end{minipage}
    \hfill
    \begin{minipage}{0.62\textwidth}
        \centering
        \setkeys{Gin}{width=0.325\linewidth} 

        \begin{minipage}{0.08\linewidth} \hfill \end{minipage}%
        \begin{minipage}{0.91\linewidth}
            \centering
            \begin{minipage}{0.32\linewidth}
                \centering\scriptsize\textbf{(a) PairTally} \cite{nguyen2025can} \\
                ``\textbf{big} marbles, not small''\\
                \raisebox{1.0ex}{``}\scalebox{0.5}{\includegraphics[height=3.0em]{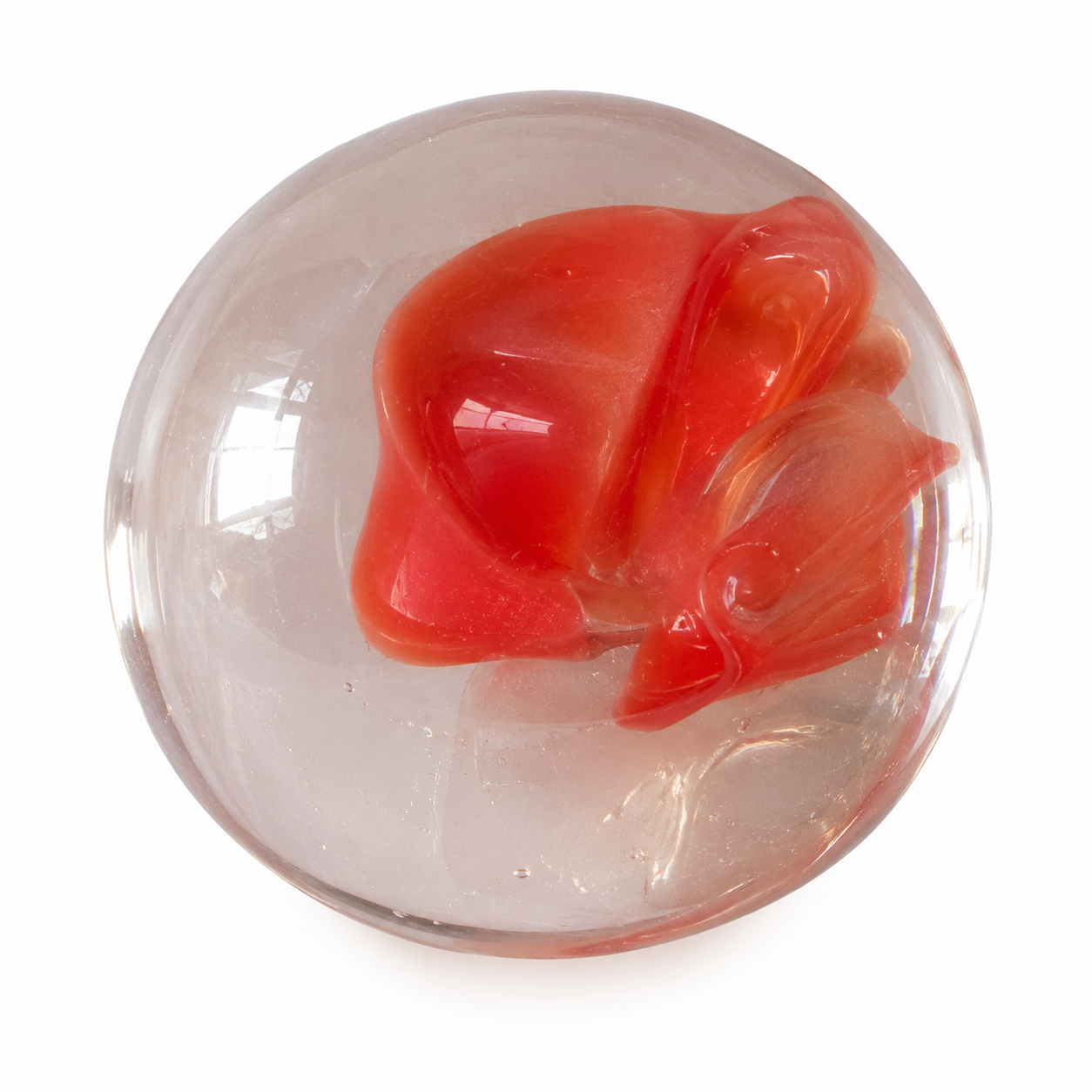}}\raisebox{1.0ex}{\scalebox{1.0}{, not}} \raisebox{0.5ex}{\scalebox{0.35}{\includegraphics[height=3.0em]{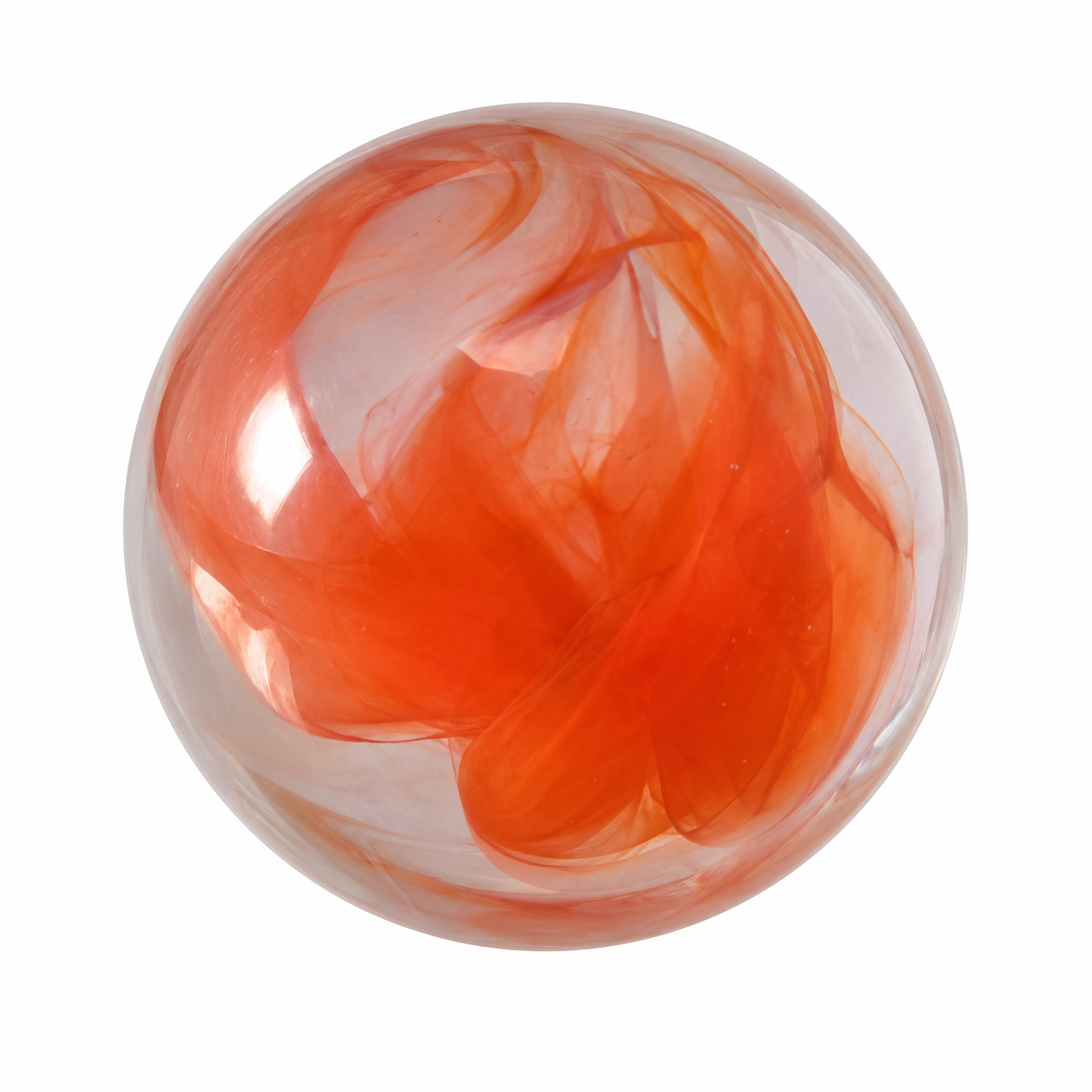}}}\raisebox{1.5ex}{''}
            \end{minipage}\hfill
            \begin{minipage}{0.32\linewidth}
                \centering\scriptsize\textbf{(b) FSC-147} \cite{ranjan2021learning}\\
                ``sunglasses''\\
            \end{minipage}\hfill
            \begin{minipage}{0.32\linewidth}
                \centering\scriptsize\textbf{(c) MixCount} \\
                ``blue cast iron teapot''\\
                \raisebox{1.0ex}{``}\scalebox{0.7}{\includegraphics[height=3.0em]{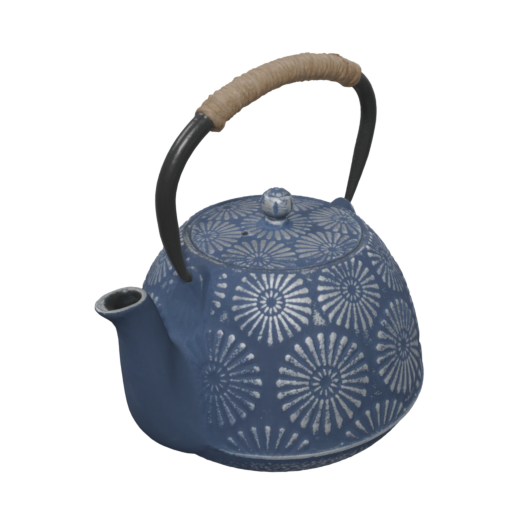}}\raisebox{1.0ex}{''}
            \end{minipage}
        \end{minipage}

        \vspace{0.0em}

        \begin{minipage}[c]{0.08\linewidth} 
            \centering
            \rotatebox[origin=c]{90}{\scriptsize \textbf{CountGD++}}
        \end{minipage}%
        \begin{minipage}[c]{0.91\linewidth}
            \centering
            \includegraphics{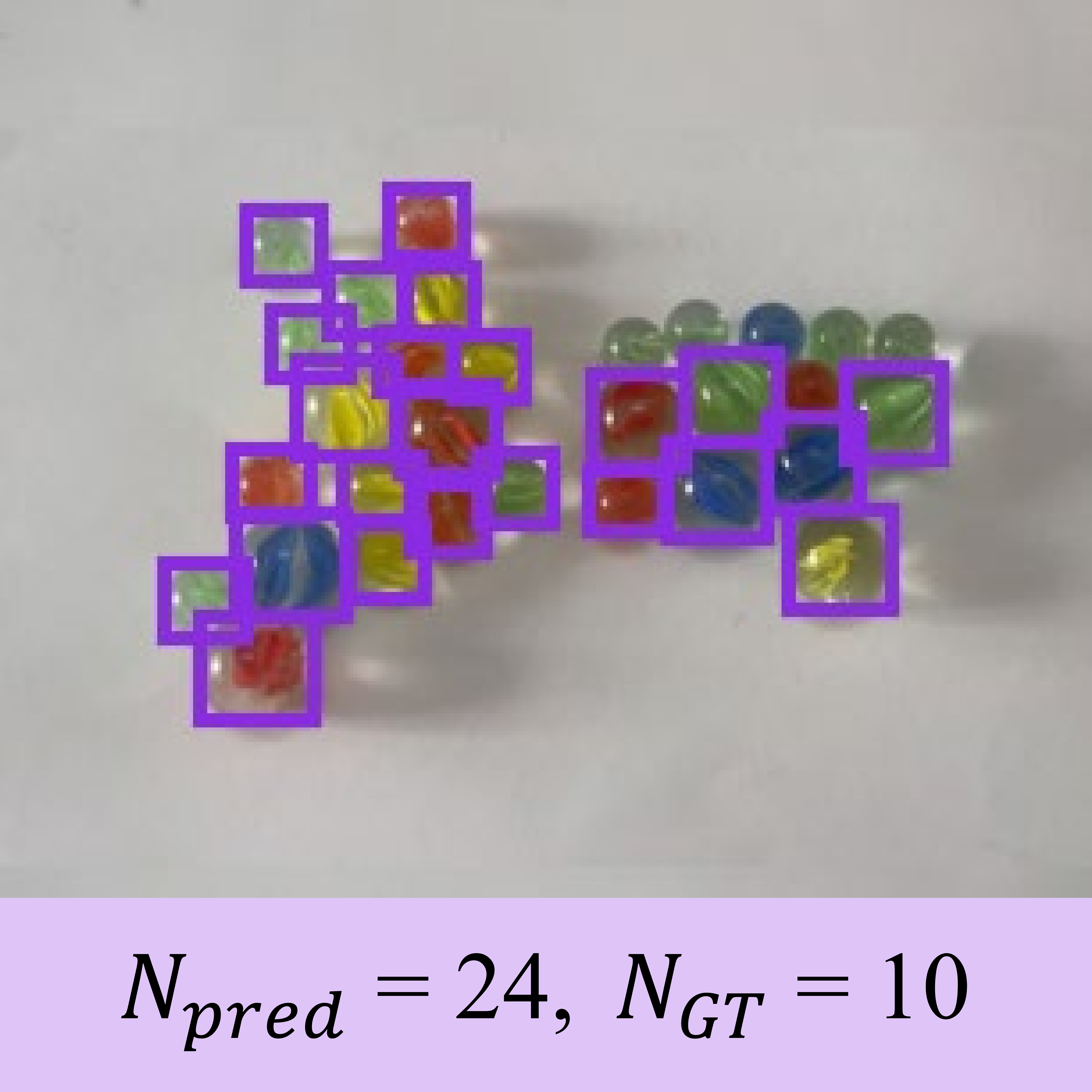}\hfill
            \includegraphics{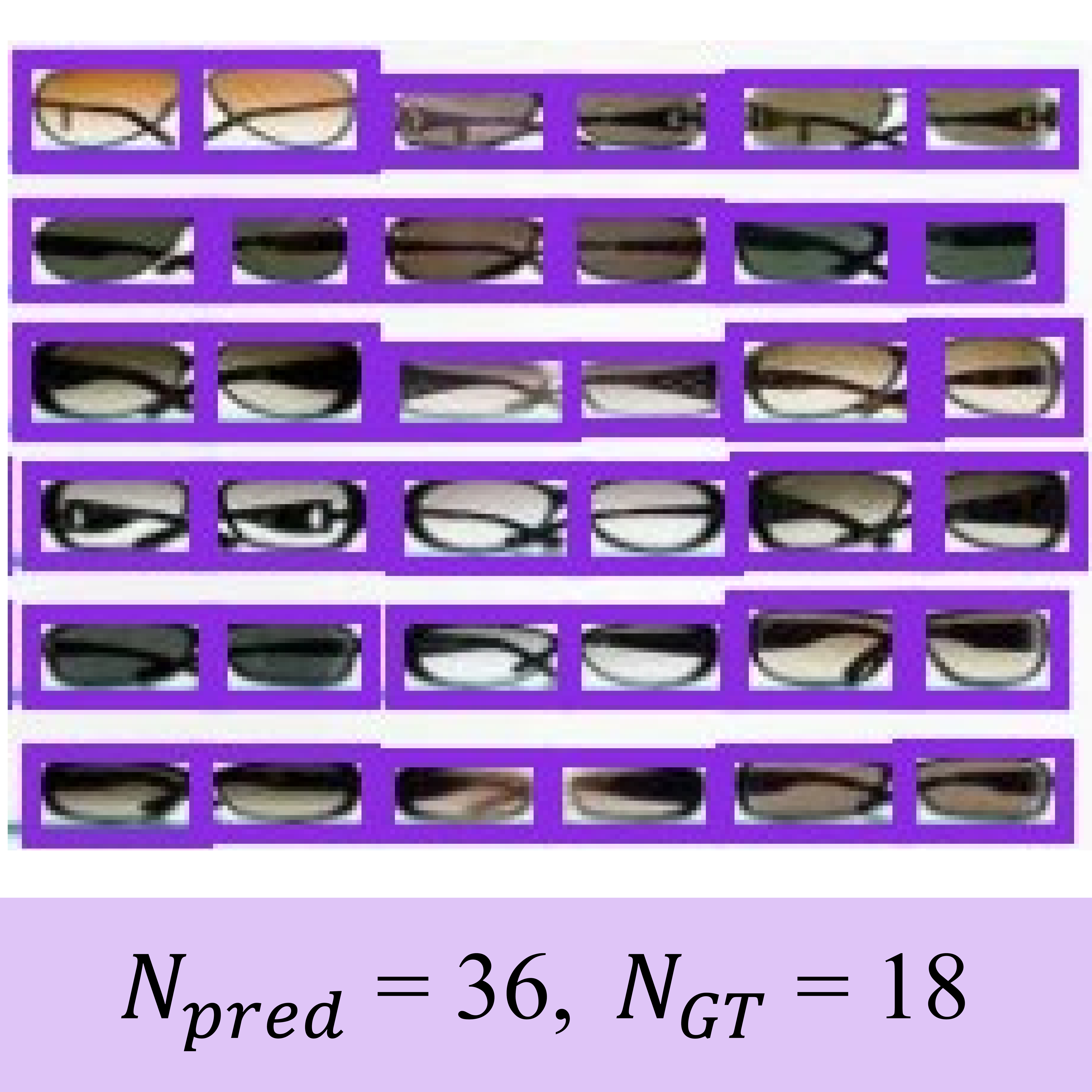}\hfill
            \includegraphics{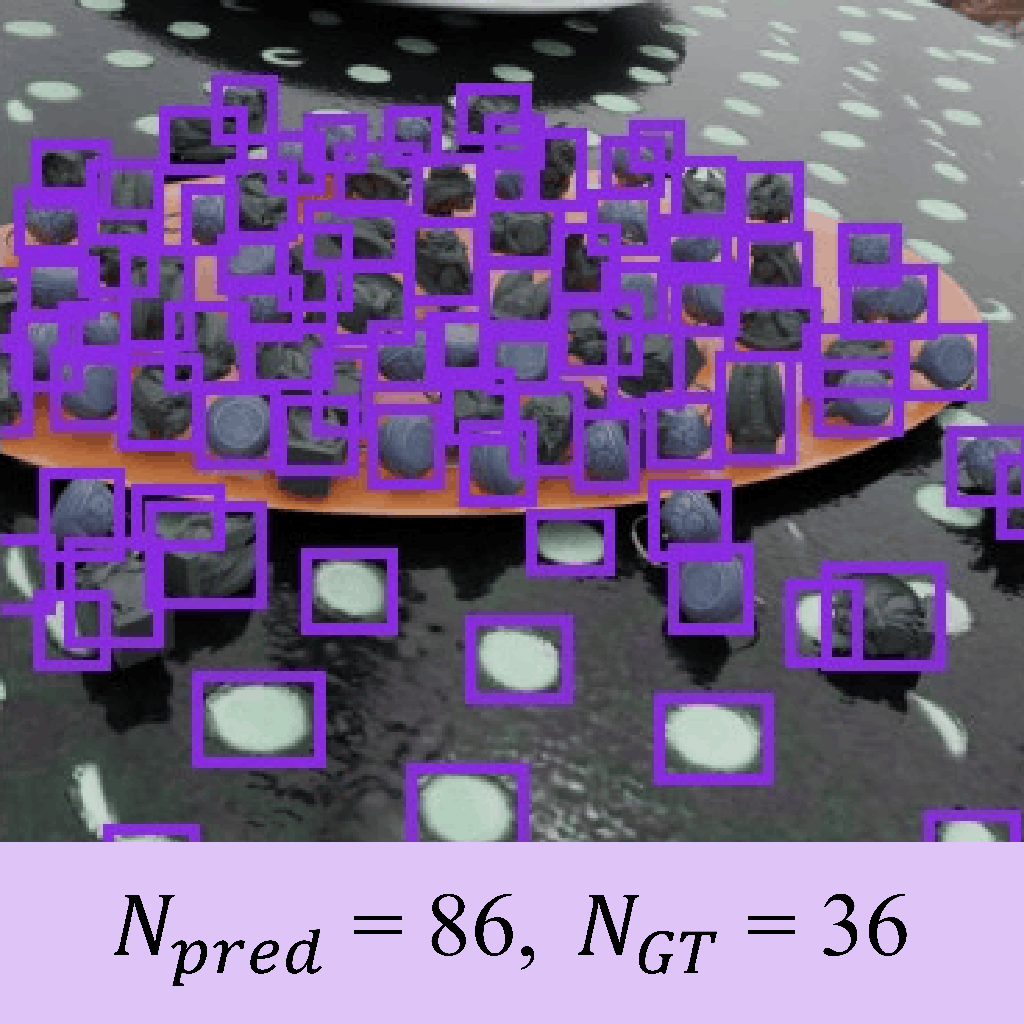}
        \end{minipage}

        \vspace{0.18em} 

        \begin{minipage}[c]{0.08\linewidth}
            \centering
            \rotatebox[origin=c]{90}{\scriptsize \textbf{%
                \setlength{\tabcolsep}{0pt} 
                \begin{tabular}{c} CountGD++ \\ (+ MixCount) \end{tabular}%
            }}
        \end{minipage}%
        \begin{minipage}[c]{0.91\linewidth}
            \centering
            \includegraphics{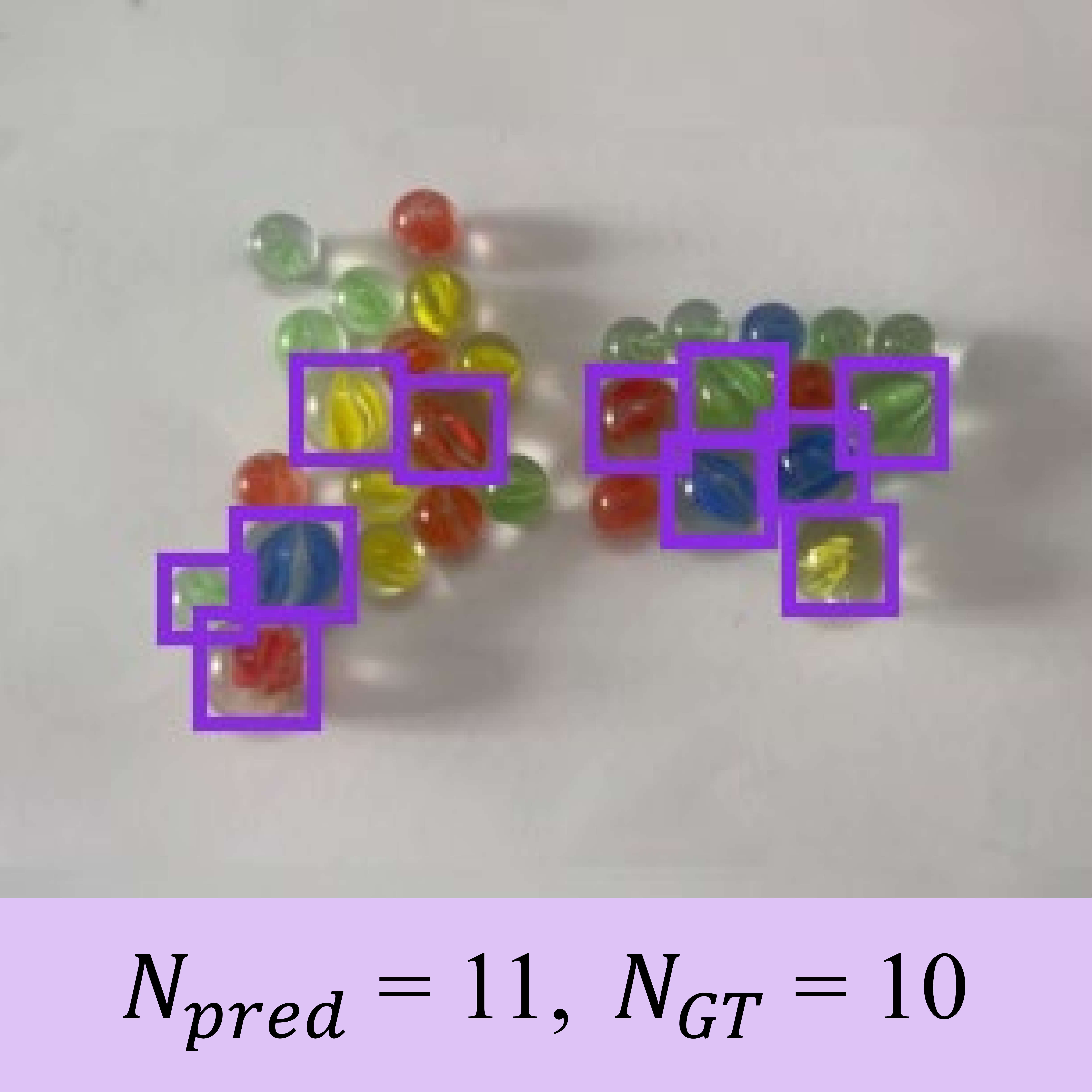}\hfill
            \includegraphics{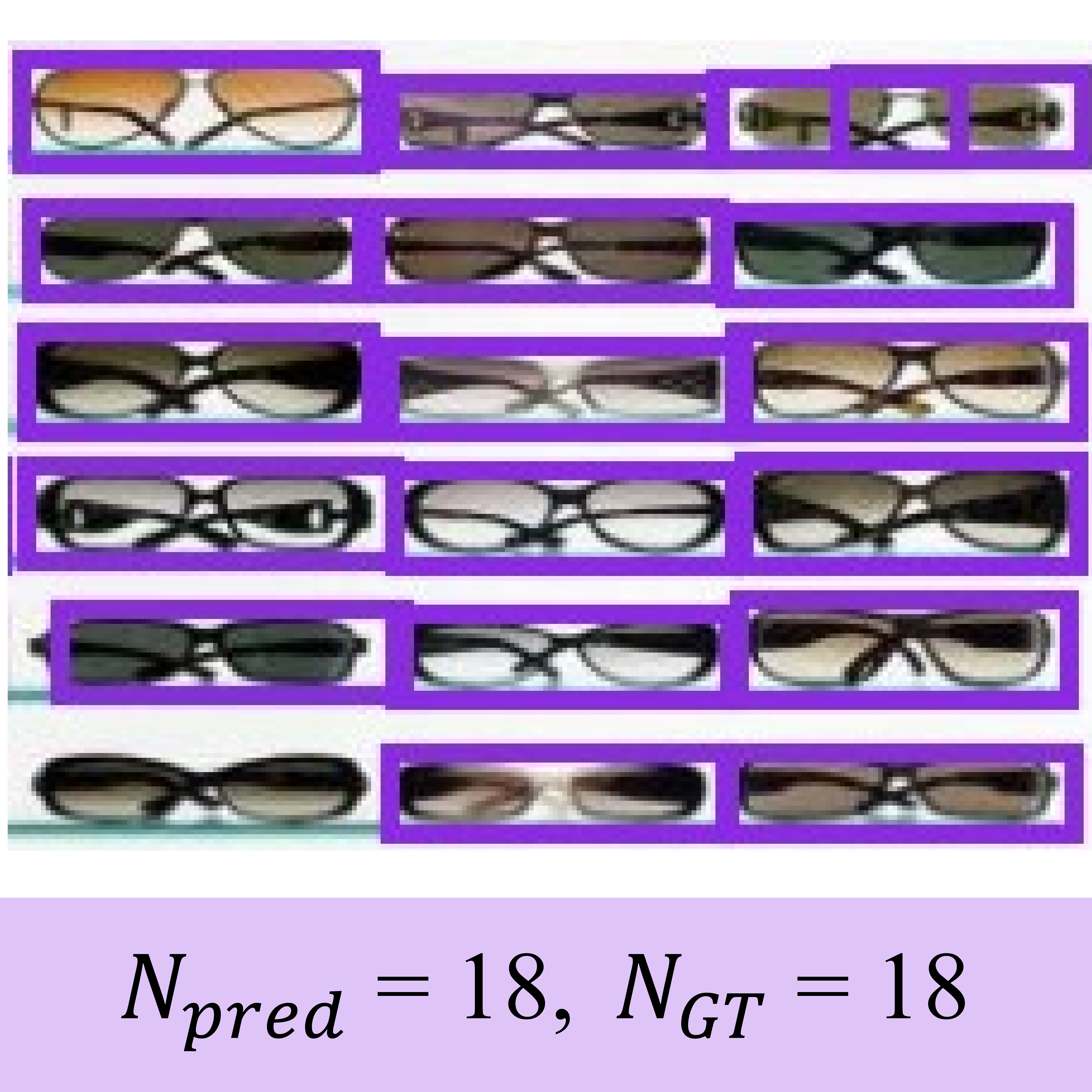}\hfill
            \includegraphics{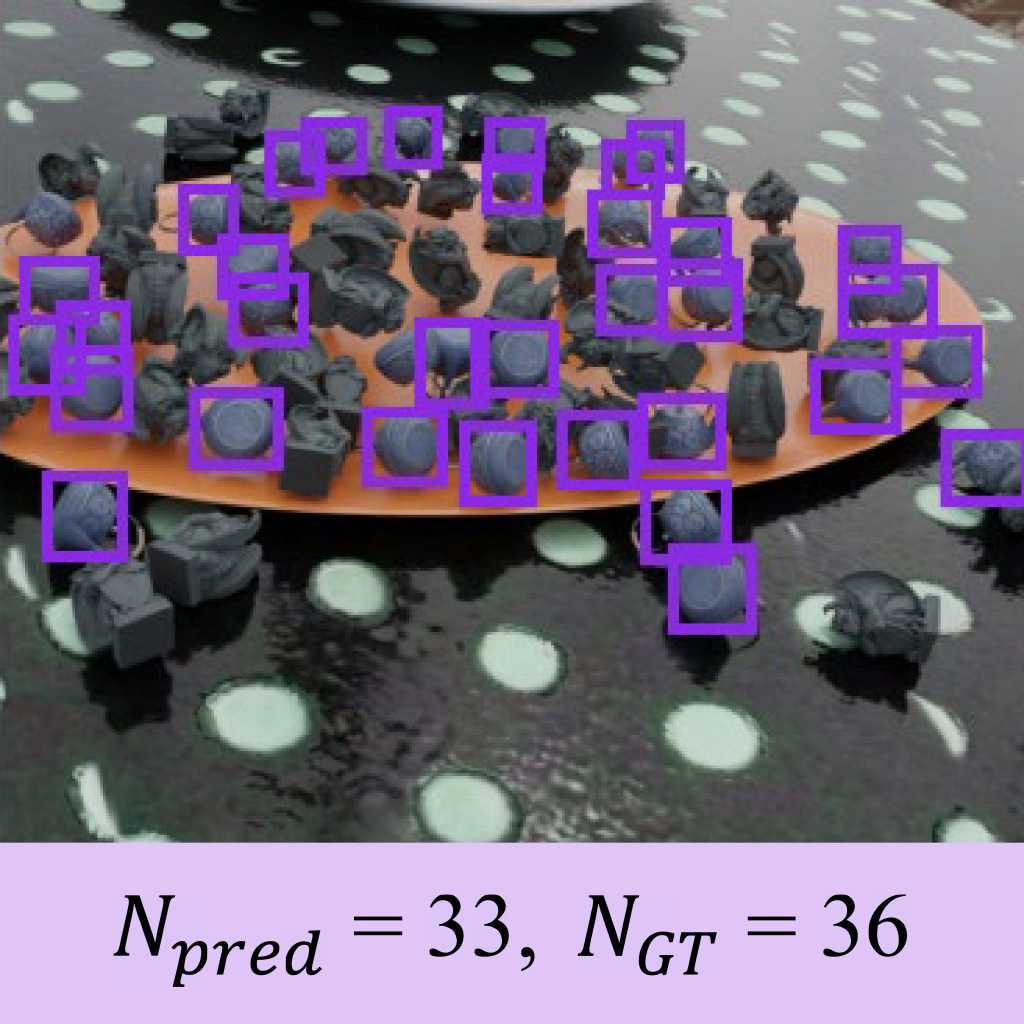}
        \end{minipage}
    \end{minipage}
    \caption{\textbf{Bridging the data gap.} Visual counting models often struggle to distinguish similar objects (a), fail to recognize self-similar components as a single entity (b), and are easily distracted by repetitive background patterns (c). We show that these failure modes can be substantially reduced with targeted training data, and design \acron~to bridge this gap, resulting in -20\% error on recent benchmarks. Prompts are shown above the images. 
    }
    \label{fig:data_gap}
\end{figure}


%% file: sec/2_rw.tex
\section{Related works}

\input{figs/comparison_table}

\paragraph{Visual counting models.}


Methods for automated object counting in images started as class-specific, capable of counting only one type of object. These include techniques for counting humans~\cite{zhang2015cross,ranjan2018iterative,liu2019context,zhang2025semi} in crowds, predefined items in industrial inspection and retail~\cite{jenkins2023countnet3d,dumery2026automated}, animals in the wild~\cite{arteta2016counting}, and cells~\cite{Flaccavento11,guo2021sau}. These models are trained to count only one type of object and thus cannot automatically count new objects at inference. Later developments allowed the user to tell the counting model what to count with \emph{prompts}, enabling the model to adapt to different objects without retraining.

More recent \emph{open-world} counting models can in theory count \emph{any} specified object given both the input image and multi-modal object prompts. CounTX~\cite{amini2023open}, CLIP-Count~\cite{jiang2023clip}, and CountSE~\cite{liu2025countse} accept text prompts that describe the object. Visual question answering~\cite{antol2015vqa,zhang2018learning,kim2025visual} and Referring Expression Counting \cite{dai2024referring,wang2025exploring,zou2026decoupling} models rely on open-ended and contextual text prompts. These developments were enabled by the availability of text prompts for the FSC-147~\cite{ranjan2021learning} and REC-8k~\cite{dai2024referring} datasets.
CounTR~\cite{liu2022countr}, GeCo~\cite{pelhan2024novel,pelhan2026generalized}, and LOCA~\cite{djukic2023low} accept visual exemplars, in the form of bounding boxes over example instances. State-of-the-art models CountGD~\cite{amini2024countgd} and CountGD++~\cite{amini2026open} accept both visual exemplars and text as prompts. By accepting prompts as extra inputs, these models adapt to counting new objects at inference. While this progress is promising, these techniques still struggle to distinguish between visually similar objects, even when given detailed prompts. 

\input{figs/sample_comparison}

\paragraph{Counting datasets.}

The most commonly used datasets for training and evaluating counting models only have one category of object labeled per image, including cars~\cite{mundhenk2016large}, humans~\cite{sindagi2019pushing,sindagi2020jhu}, cells~\cite{huang2020bcdata}, fish~\cite{kay2022caltech}, or surgery instruments~\cite{bhyri2026chain}.

In open-world counting, the standard FSC-147~\cite{ranjan2021learning} dataset contains 6135 images with objects from 147 different categories, but only one type of object is labeled per image. Similarly FSCD-LVIS~\cite{nguyen2022few} has images with different categories of objects, but only one type is labeled per image. This pattern is common in video~\cite{amini2026open,huang2026countex} and stacked object counting~\cite{dumery2025counting,dumery2025stackcounting} too, where datasets label one class of object per sample as well. This has embedded a bias in SoTA counting models trained and evaluated on these datasets, which struggle to distinguish between visually similar instances.

Recent work introduces new but limited real-world datasets with multiple categories of objects labeled. The PairTally~\cite{nguyen2025can} test set includes 681 images of paired objects that are mixed together with ground truth counts, visual exemplars, and text descriptions. REC-8k~\cite{dai2024referring} includes about 8000 images split into training, validation, and test sets with point annotations and fine-grained text descriptions. The OmniCount-191~\cite{mondal2025omnicount} dataset consists of 30,230 images split into training and test sets. 
The Pixmo-Count~\cite{deitke2025molmo} dataset contains images, text descriptions, and point annotations for 36,900 training images and 540 validation and test images. The Lookalikes test set~\cite{dalessandro2025justsaytheword} contains images, text descriptions, and ground truth counts. PairTally and the Lookalikes datasets lack training sets, so they are not suitable for training counting models. REC-8k and Pixmo-Count lack bounding boxes or visual exemplars. OmniCount-191~\cite{mondal2025omnicount} contains annotations errors pointed out in \cite{AminiNaieni26} such as missing labels.


To address the limitations of real-world data, recent work has turned to synthetic data. The synthetic MCAC~\cite{hobley2024abc} dataset features high counts of mixed objects and ground truth bounding boxes. The synthetic SITUATE~\cite{peinl2026situate} dataset contains images with only four object classes: cubes, spheres, cones and cylinders. Both MCAC and SITUATE lack photorealistic images. MCAC lacks text descriptions and counts hidden objects, making it unsuitable for the 2D multi-modal counting task, and the images in SITUATE are highly simplistic. Thus, while synthetic data pipelines offer the potential to generate effectively unlimited labeled counting data, existing approaches lack the realism, diversity, accurate annotations, and prompts required for training and evaluating modern multi-modal counting models. We compare our \acron~dataset to these synthetic alternatives in \cref{fig:qualitative_comparison}.

Overall, existing datasets lack a combination of (i) multiple visually similar object categories labeled per image, (ii) high-quality multi-modal prompts and annotations, and (iii) scalable, realistic, and diverse training data, restricting progress in mixed-object counting. Some of these limitations are highlighted in \cref{tab:method_comparison}.

%% file: figs/comparison_table.tex
\begin{table}[t]
    \centering
    \footnotesize
    \setlength{\tabcolsep}{4pt}
    \begin{tabular}{lccccc}
        \toprule
        \textbf{Feature} & 
        \textbf{\makecell{MCAC \cite{hobley2024abc}\\(ECCV24)} } & 
        \textbf{\makecell{SITUATE \cite{peinl2026situate}\\(VISAPP26)}} & 
        \textbf{\makecell{FSC-147 \cite{ranjan2021learning}\\(CVPR21)}} & 
        \textbf{\makecell{PairTally\cite{nguyen2025can}\\(DICTA25)}} &
        \textbf{\makecell{\acron\\(Ours)}} \\
        \midrule
        \textbf{Multiple object types labeled per image} & \cmark & \cmark & \xmark & \cmark & \cmark \\
        \textbf{Fine-grained text prompts} & \xmark & \xmark & \xmark & \cmark & \cmark \\
        \textbf{External exemplars} & \xmark & \xmark & \xmark & \xmark & \cmark \\
        \textbf{Exemplar score} & \xmark & \xmark & \xmark & \xmark & \cmark \\
        \textbf{Segmentation masks} & \cmark & \cmark & \xmark & \xmark & \cmark \\
        \textbf{Bounding boxes} & \cmark & \cmark & \xmark & \xmark & \cmark \\
        \midrule
        \textbf{\# Object classes} & 343 & 4 & 147 & 98 & \textbf{1522} \\
        \textbf{\# Images} & 20,483 & 6875 & 6135 & 681 & \textbf{58,000} \\
        \bottomrule
    \end{tabular}
    \vspace{0.5em}
    \caption{\textbf{Comparison of \acron~against existing counting benchmarks.} \acron~remedies key limitations of prior counting datasets while directly addressing the failure modes of existing counting models. \acron~includes complex environments (e.g., distractors, repetitive backgrounds, perspective distortion, self-similar objects), provides richer prompts (e.g., external exemplars, exemplar scores, multiple detailed text descriptions per object), and richer annotations (e.g., bounding boxes, segmentation masks, depth maps) than prior work.}
    \label{tab:method_comparison}
\end{table}

%% file: figs/sample_comparison.tex
\begin{figure*}[t]
    \centering
    \begin{subfigure}[b]{0.325\textwidth}
        \centering
        \includegraphics[width=0.49\linewidth]{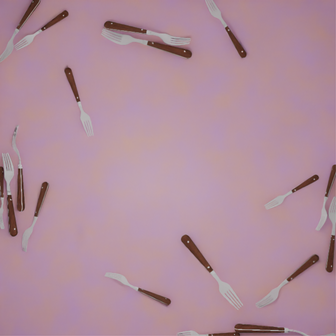} \hfill
        \includegraphics[width=0.49\linewidth]{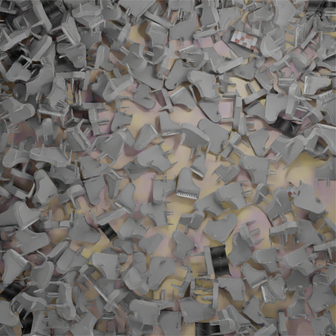} \\
        \vspace{0.5mm}
        \includegraphics[width=0.49\linewidth]{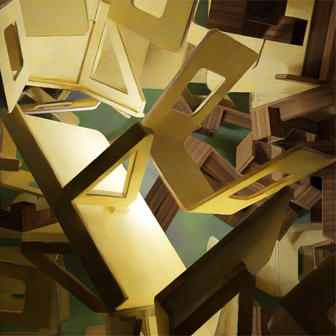} \hfill
        \includegraphics[width=0.49\linewidth]{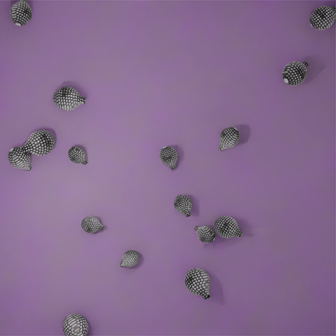}
        \caption{MCAC \cite{hobley2024abc}}
        \label{fig:mcac}
    \end{subfigure}
    \hfill
    \begin{subfigure}[b]{0.325\textwidth}
        \centering
        \includegraphics[width=0.49\linewidth]{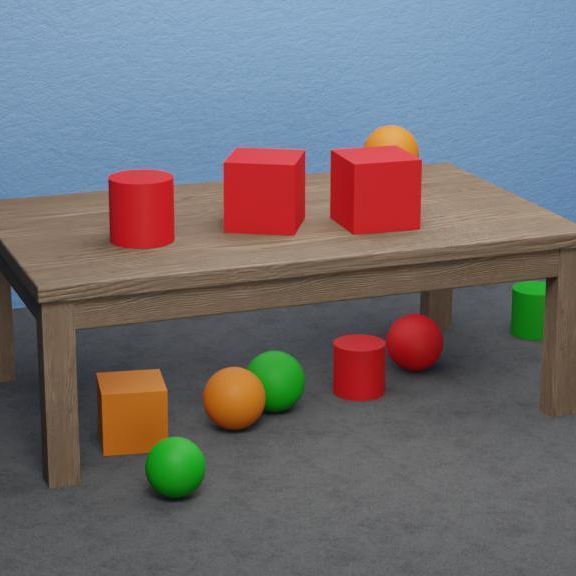} \hfill
        \includegraphics[width=0.49\linewidth]{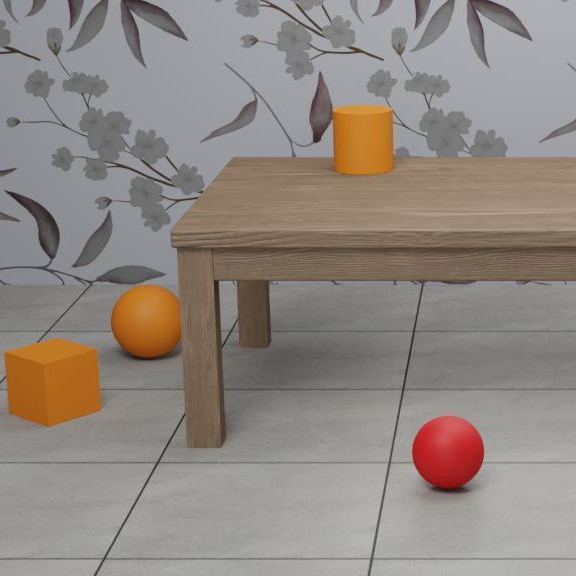} \\
        \vspace{0.5mm}
        \includegraphics[width=0.49\linewidth]{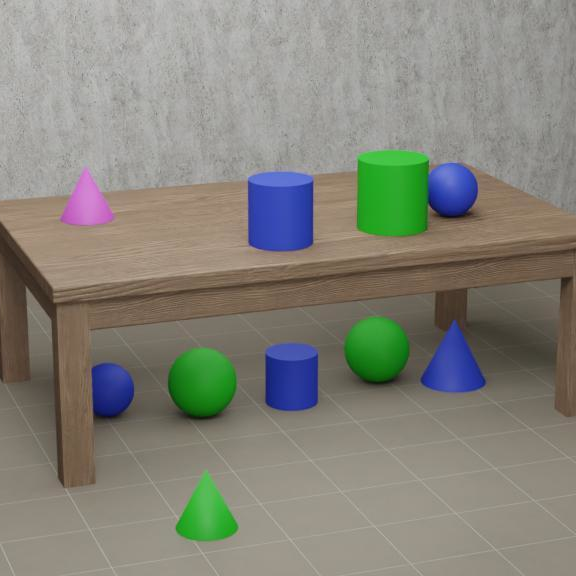} \hfill
        \includegraphics[width=0.49\linewidth]{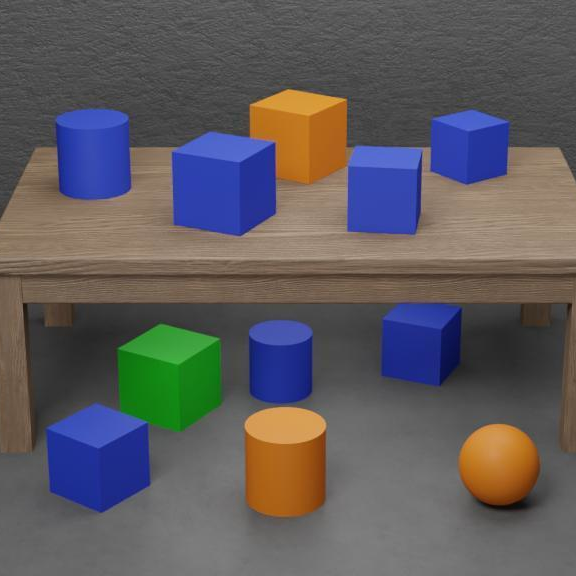}
        \caption{SITUATE \cite{peinl2026situate}}
        \label{fig:situate}
    \end{subfigure}
    \hfill
    \begin{subfigure}[b]{0.325\textwidth}
        \centering
        \includegraphics[width=0.49\linewidth]{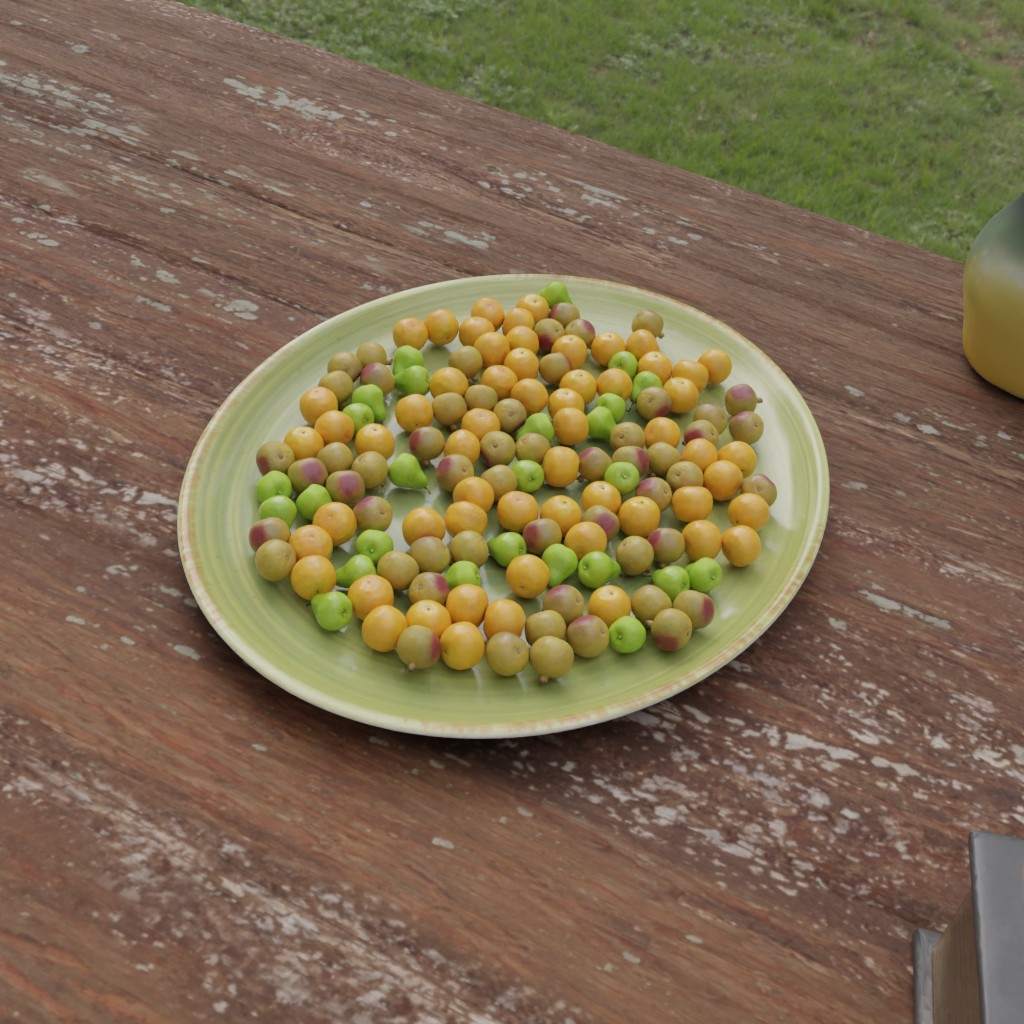} \hfill
        \includegraphics[width=0.49\linewidth]{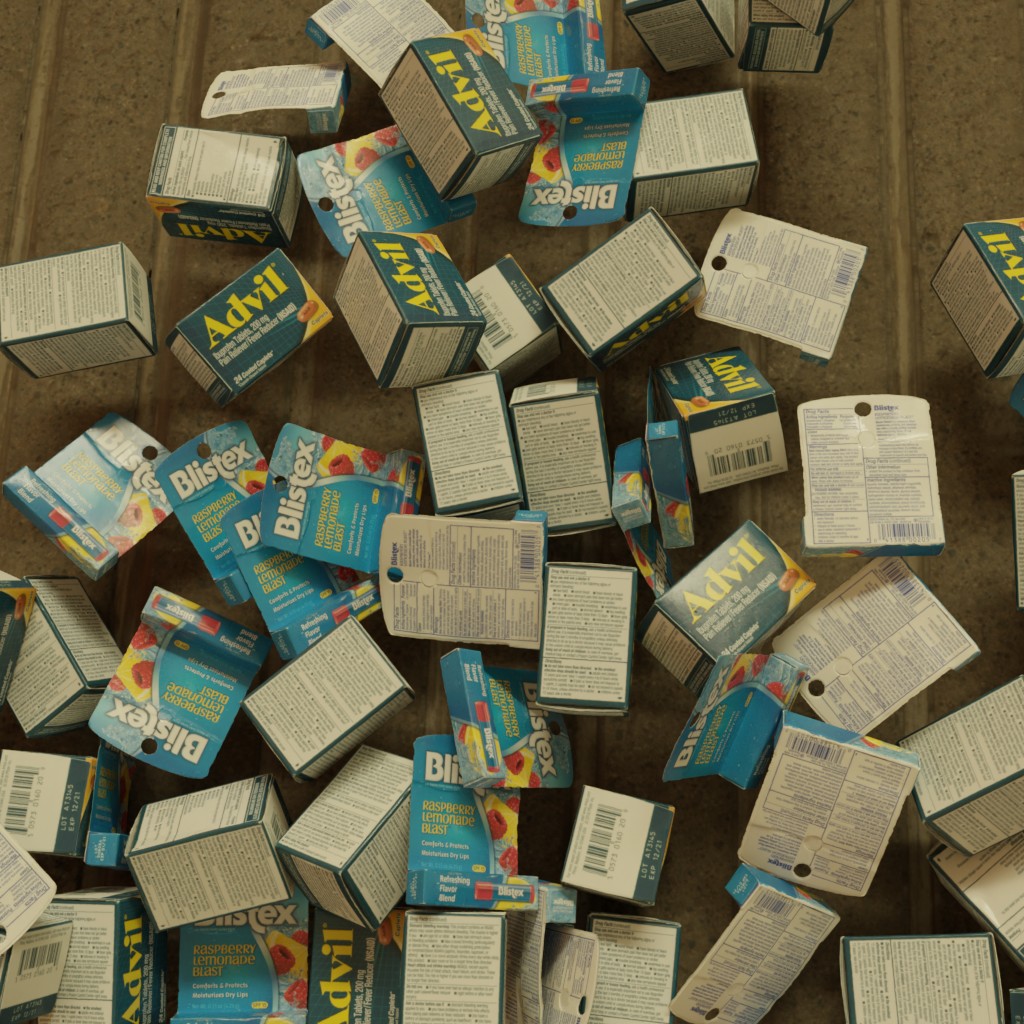} \\
        \vspace{0.5mm}
        \includegraphics[width=0.49\linewidth]{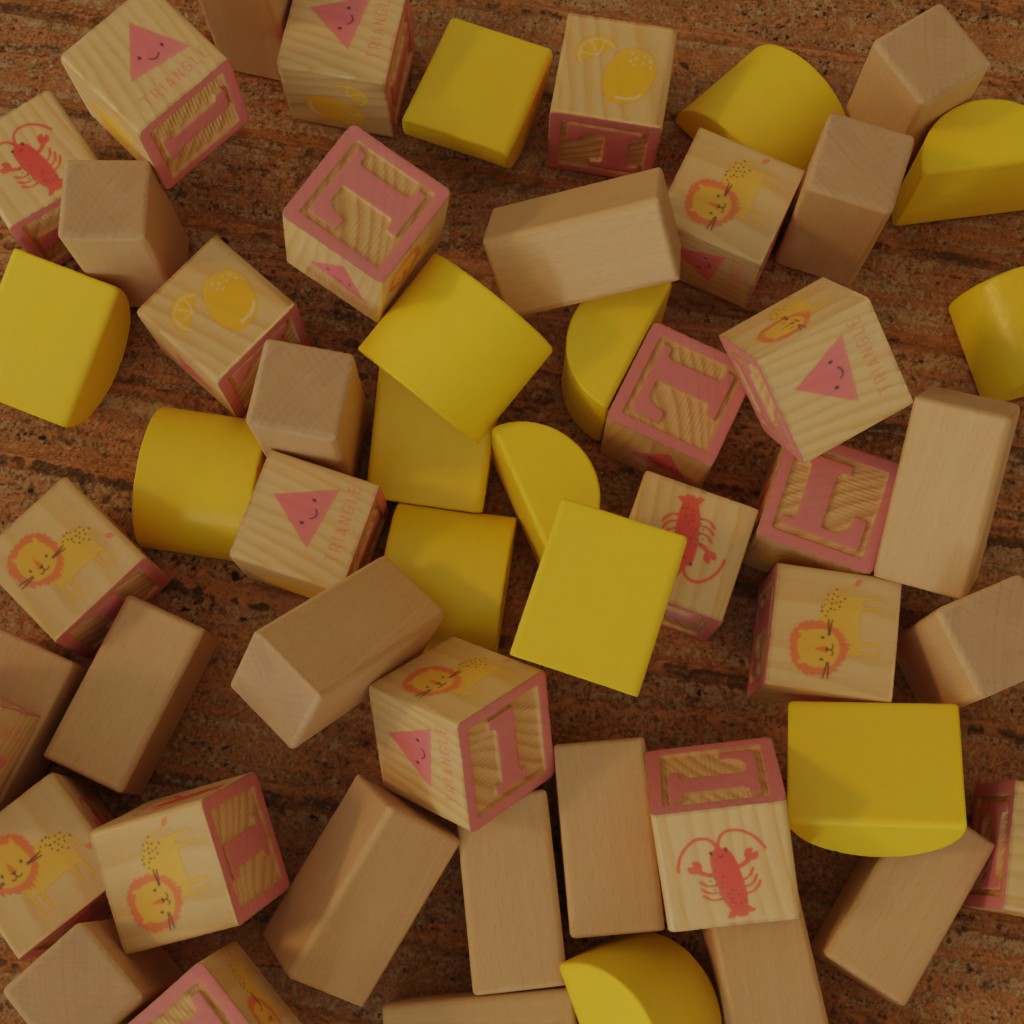} \hfill
        \includegraphics[width=0.49\linewidth]{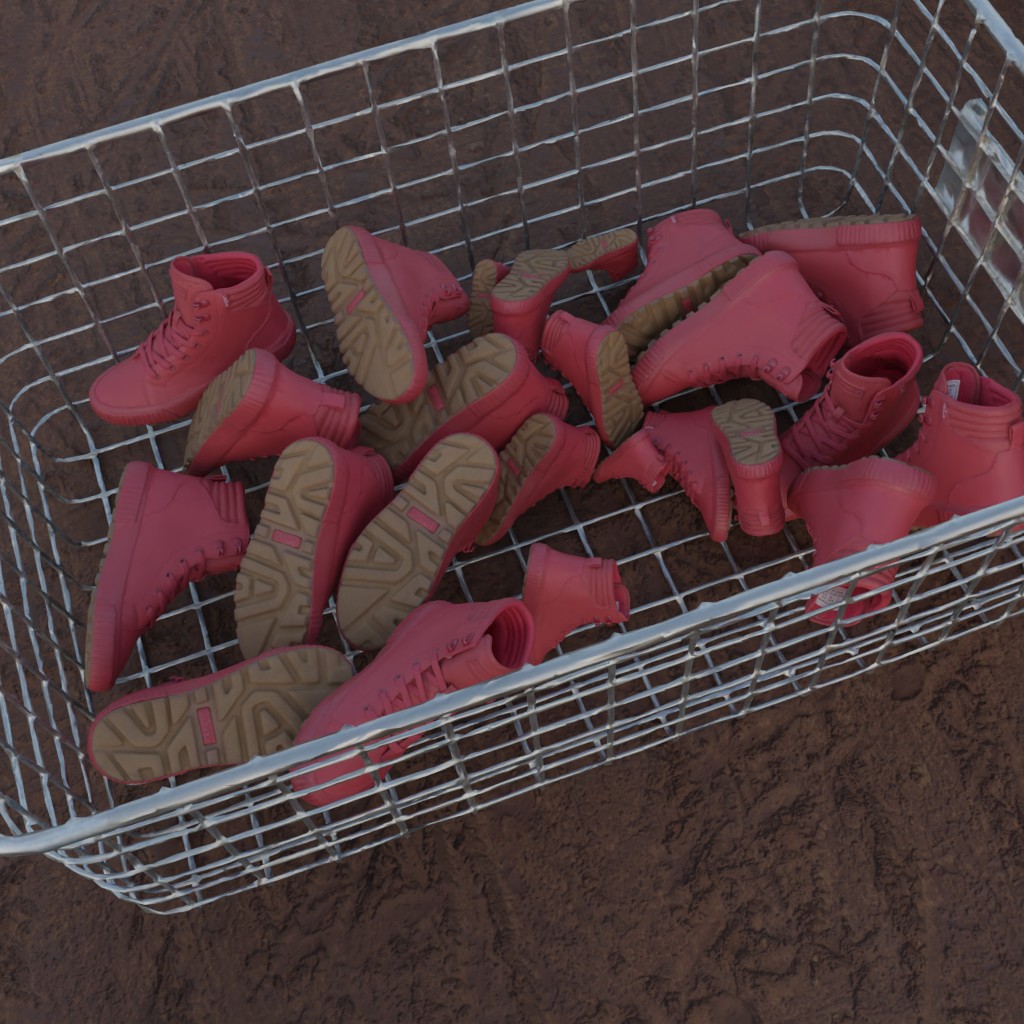}
        \caption{MixCount (Ours)}
        \label{fig:mixcount}
    \end{subfigure}
    
    \caption{\textbf{Large-scale synthetic data comparison.} (a) MCAC~\cite{hobley2024abc} and (b) SITUATE~\cite{peinl2026situate} samples lack the complexity of real-world images. In comparison, (c) \acron~samples are more diverse, realistic, and challenging.}
    \label{fig:qualitative_comparison}
\end{figure*}

%% file: sec/3_dataset.tex
\section{The \acron~dataset and data generator}

\input{figs/data_features}

Our work is built upon the observation that existing visual counting datasets are either limited in diversity and scale or lack realism. Large-scale datasets such as the ones in \cref{fig:qualitative_comparison} typically include a small number of objects, feature very simplistic environments, and have limited diversity. Real-world training sets such as FSC-147~\cite{ranjan2021learning} lack images with labeled mixed objects.
As shown in the experiments, this has become a major bottleneck for the performance and generalization of the counting models trained on these datasets. In \cref{sec:dataset_card}, we describe a large-scale training dataset that addresses this issue with photorealistic, diverse, and accurate samples. 

In \cref{sec:datasgen_pipeline}, we present the system we developed to acquire this dataset and explain how it can be tailored for specific applications to synthesize more large-scale data in the future. As will be demonstrated in \cref{sec:experiments}, the data generated through this process unlocks new capabilities for state-of-the-art counting models. By simply training existing architectures on the \acron~dataset, we significantly outperform previous baselines on all counting metrics, highlighting the current need for high-quality and large-scale training data.

\subsection{\acron: a modern large-scale counting dataset}\label{sec:dataset_card}

\paragraph{Diversity and scale.}
With this work, we release the initial version of the \acron~dataset, the first large-scale photorealistic synthetic dataset for training and benchmarking visual counting models. \acron~represents a significant leap in the diversity of richly annotated counting datasets, as can be seen in \cref{fig:qualitative_comparison}. This is achieved through procedural generation and the complete randomization of the objects, their surrounding environment, camera placement and light conditions. 
Overall, \acron~includes 58,000 images 
containing a total of 4 million objects to count. There are 68 objects per-scene on average sampled from two to four different classes. This encompasses 1522 different commonly found real-world objects spanning a large number of categories. We provide additional statistics on the dataset and objects in our supplementary.

\paragraph{Object prompts.} 

To allow the training and evaluation of recent multimodal models~\cite{amini2024countgd, liu2025countse, AminiNaieni26}, each object in \acron~is associated with both text descriptions and visual exemplars. These \textit{object prompts} can be used to specify which object is to be counted and we display some examples in \cref{fig:data_features}.
Regarding text descriptions, we observed that previous datasets \cite{hobley2024abc, ranjan2021learning} often suffer from simplistic or ambiguous descriptions, such as \textit{bird}, \textit{seagull}, \textit{finger food} in FSC-147~\cite{ranjan2021learning}. As a consequence, the counting models trained on these datasets generalize poorly to real-world applications, in which several qualifiers are often necessary to designate a specific object.
To remedy this, each object in \acron~ is annotated at three different levels: short, concise and detailed descriptions. These serve as unique identifiers and we ensure that no two objects have an identical text prompt.

We also introduce two different kinds of visual exemplars. An \textit{internal} exemplar, represented as a bounding box in the image, and an \textit{external} exemplar, which is an image of the same object but in a different setting. \acron~includes both internal and external exemplars. Internal exemplars are assigned a score $s$ that relates to the underlying object's visibility and size in the image. \acron~also includes external exemplars of each object in canonical pose in front of a white background, as well as crops from different scenes of our dataset.

\input{figs/annotations}

\paragraph{Annotations.}

Every object is annotated with a bounding box. These annotations are saved in COCO format \cite{lin2014microsoft} and can directly be integrated in the majority of existing pipelines to train visual counting models. 
By design, all objects in the ground-truth annotations are guaranteed to be visible in that image, which was an issue in previous datasets such as MCAC \cite{hobley2024abc} where objects were often completely hidden yet still included in the ground truth annotation, leading to inconsistent supervision in training.
Finally, we also save instance and class segmentations of the objects, as illustrated in \cref{fig:GT_maps}, as well as ground-truth depth and normal maps. 

\paragraph{Splits.}
We divide the \acron~dataset into train-val-test split, including 56000, 1000 and 1000 images, respectively. To guarantee that no object found in the test set can be seen in the train set, we split the objects to be counted directly, such that each object can only be found in a single split. 
To ensure compatibility with models trained and evaluated on FSC-147~\cite{ranjan2021learning}, we match each object in our dataset with a category in FSC-147 when applicable and use the same train-test split. 
For example, since \textit{ball} is a train class in FSC-147, all types of balls such as \textit{Light orange football} or \textit{Camo patterned basketball} are part of the train split. Finally, the objects which do not broadly match any of FSC-147's limited classes are then assigned a random split to match our target distribution.


\subsection{Data generator}\label{sec:datasgen_pipeline}


Our central motivation is to provide counting data that is as diverse as possible, photorealistic, and challenging.
Thus, we design the \acron~data generator to address the failure modes of current counting models in a fully automatic fashion and at scale. We also designed it to be versatile and easily adaptable to generate new data for specific applications.
We opt for a synthetic data generator depicted in~\Cref{fig:data_generator} that incorporates captures of real-world objects, environments, and materials, and physically simulates them to compose realistic scenes. In combination with physically-based rendering, this provides our synthesized samples the photorealism and complexity of real-world images while maintaining the annotation quality and scale of synthetic data. Due to space constraints, further details on our data generator, visualization of external exemplars, and dataset statistics can be found in our supplementary.

\paragraph{Objects.}
For each scene, we start by sampling two to four different object classes. Objects are randomly picked from the Digital Twin Catalog (DTC)~\cite{dong2025digital}, a dataset of high-quality captures of real-world objects. We manually inspect the dataset to filter out objects that are too similar to be distinguished by text or visual exemplar.
After the first object class is created, the subsequent classes are generated either by modifying its size, by sampling another object from the same DTC category, or by sampling from a different category with a reduced probability. 
This favors meaningful combinations while preserving the possibility to generate improbable mixes adding to data diversity.

\paragraph{Simulation.} We sample 3D positions and spawn an object from a random class at each position. A slight scale modifier is applied to each object, ensuring all objects of the same class are different while remaining visually similar. We then employ the physical simulation of Blender~\cite{blender} to let the objects naturally form a realistic scene, with objects touching and partially occluding each other.

\input{figs/data_generator}

\paragraph{Occlusions.} 
Previous dataset generation pipelines provide no guarantee that all objects to be counted are indeed visible. 
\acron~solves this by explicitly measuring the visibility of each object and rejecting occluded objects. Specifically, for each scene a segmentation image is rendered from which masks for all $N$ objects in the scene can be extracted. However, these masks only cover parts of the objects since other objects or elements of the environment may occlude them partially or completely. We will thus refer to these masks as the \textit{occluded} masks $(M_{o,i})_{1\leq i \leq N}$. To remedy this, we also render \textit{unoccluded} masks $(M_{u,i})_{1\leq i \leq N}$, in which all occlusions are ignored, by doing a fast render pass per-object with all other objects disabled.
This allows our data generator to measure, for an object $i$, the visibility ratio $v_i = \frac{|M_{o,i}|}{|M_{u,i}|}$. Given an input visibility threshold $v_t$, the occluded objects with $v_i < v_t$ are thus removed from the scene prior to rendering. Since doing this only makes the remaining objects more visible, this guarantees that all remaining objects are visible by at least $v_t$.

\paragraph{Environment.} 
Models trained on existing datasets~\cite{hobley2024abc,peinl2026situate,dumery2026automated} fail to generalize to the rich and challenging environments found in real-world images, partly due to their simplistic backgrounds. \acron~addresses this issue with procedural generation, real object captures and physically-based rendering. All assets used to produce \acron~are based on real-world captures and are parts of publicly released datasets, ensuring the complete reproducibility of our dataset.

First, we employ real-world captures of environment maps to provide a complex and diverse lighting environment, both in the directionality of light sources and their intensity. For each scene, a random environment map is sampled from the 2233 available in LavalIndoor~\cite{gardner2017learning} dataset or the 963 in Polyhaven~\cite{zaal2021polyhaven}. We also use 568 physically-based rendering materials (PBR) from Polyhaven~\cite{zaal2021polyhaven} and 150 from VasTexture~\cite{eppel2024vastextures} to texture the floor upon which the objects lie. We purposefully favor repetitive textures that are challenging for counting models and are absent in previous datasets.

Finally, we introduce \textit{distractors} to the scene, which are objects from the DTC dataset just like the main objects to count. However, these objects do not repeat and only serve to add complexity to the scene and cast shadows on the main objects. These objects significantly complicate lighting and add another layer of realism to the synthesized scenes. Note that distractors are guaranteed to be different objects from the main objects to count, and their bounding boxes are also reported in the final results. 

\paragraph{Rendering.}
In each simulated scene, we define the camera origin through its azimuthal and polar angles $(\phi, \theta)$ as well as its radius $r$, which are sampled from an acceptable range. The camera rotation is dynamically set to look at the center of the objects to count, then slightly jittered to introduce diversity and mimic an imperfect capture.
Finally, we generate an image through ray-tracing~\cite{blender}, which computationally simulates reflections in the scene and produces photorealistic renders. 

\textbf{Exemplars and text descriptions.} Internal exemplars are extracted from the segmentation maps. To provide external exemplars, we also render each DTC object under a standard lighting and with a white background. We task Gemini 2.5~\cite{comanici2025gemini} with captioning these images into short, concise, and detailed descriptions for each object. Following this, we remove all objects with identical descriptions to ensure each object is uniquely distinguishable.
\paragraph{Customization.}
There are numerous real-world counting tasks which are currently restricted by the limited amounts of domain-specific annotated data. We designed the \acron~data generator to be highly versatile and easily tailored for a specific application. 
For instance, in the case of industrial inspection where specific boxes and items are present in a warehouse, it is now fairly straightforward with image-to-3D or text-to-3D methods~\cite{xiang2025trellis,xiang2025trellis2} to generate 3D assets of the specific items of interest. These assets can then be plugged into the \acron~data generator to synthesize a large-scale dataset tailored for this application. The number of objects generated, multi-view camera placement, and numerous other parameters can then be customized.


%% file: figs/data_features.tex
\renewcommand{\tabularxcolumn}[1]{m{#1}}
\newcolumntype{Z}{>{\centering\arraybackslash}X}
\newcolumntype{M}[1]{>{\centering\arraybackslash}m{#1}}

\begin{figure*}[t]
    \centering
    \setlength{\tabcolsep}{4pt}
    \renewcommand{\arraystretch}{0.9}

    \begin{tabularx}{\linewidth}{@{} c c M{0.10\linewidth} M{0.14\linewidth} Z @{}}
        \toprule
        \multirow{2}{*}{\small\textbf{Image}} & \multirow{2}{*}{\small\textbf{Exemplars}} & \multicolumn{3}{c}{\small\textbf{Descriptions}} \\
        \cmidrule(l){3-5}
         &  & {\scriptsize\textbf{Short}} & {\scriptsize\textbf{Concise}} & {\scriptsize\textbf{Detailed}} \tabularnewline
        \midrule

        \multirow{2}{*}[2.45ex]{\includegraphics[width=0.17\linewidth, valign=m]{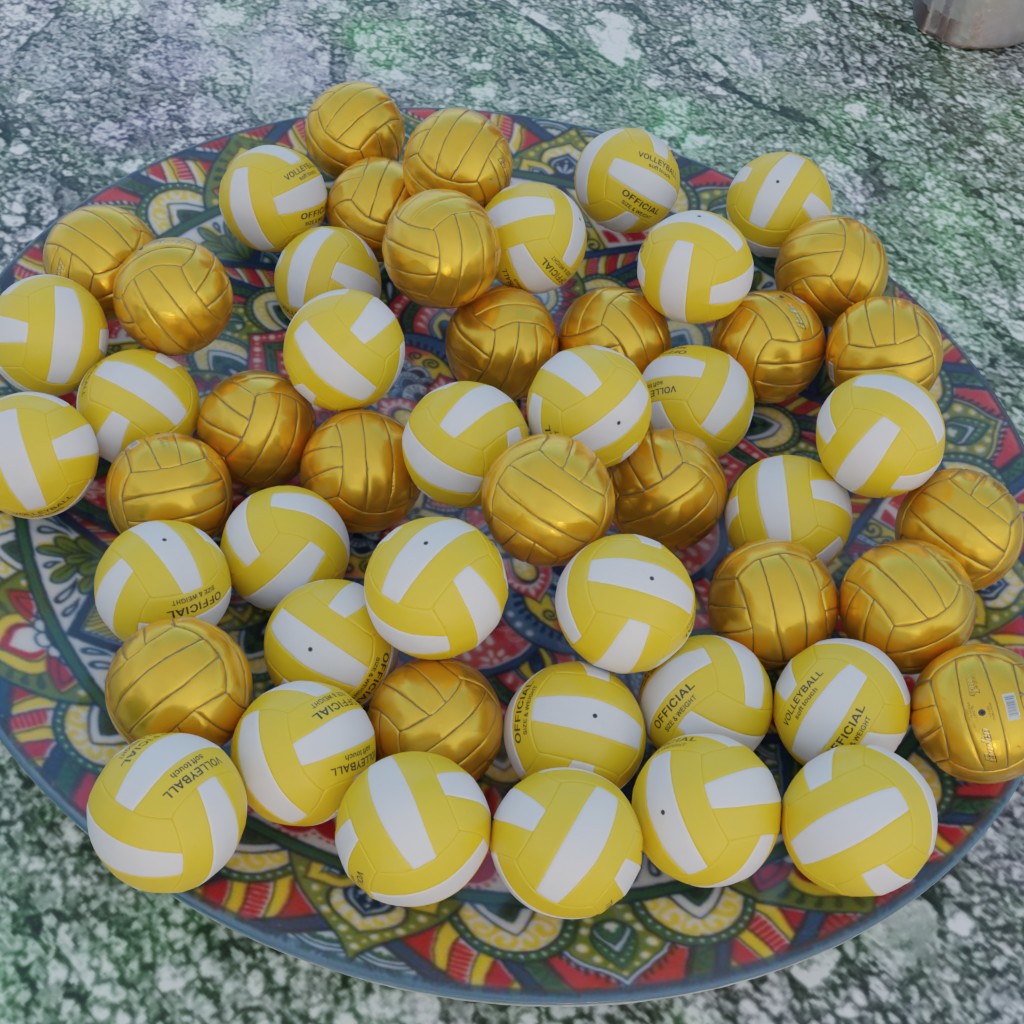}}
        & \begin{tabular}{@{}ccc@{}}
            \includegraphics[width=0.05\linewidth, valign=m]{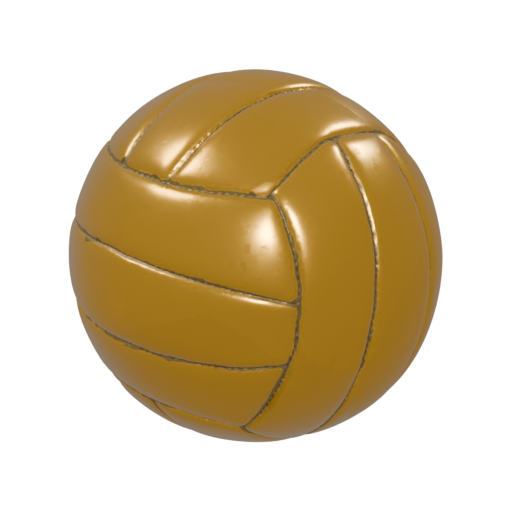} &
            \includegraphics[width=0.05\linewidth, valign=m]{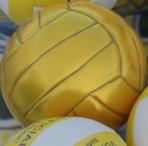} &
            \includegraphics[width=0.04\linewidth, valign=m]{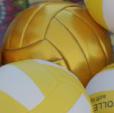} \\[-0.2em]
            {\tiny ext.} & {\tiny s=1.0} & {\tiny s=0.60}
          \end{tabular}
        & {\scriptsize gold paneled volleyball}
        & {\scriptsize Shiny golden volleyball with dark seams}
        & {\scriptsize Shiny golden volleyball features segmented panels joined by dark seams, displaying a smooth and reflective surface.} \\
        \cmidrule(lr){2-5}

        & \begin{tabular}{@{}ccc@{}}
            \includegraphics[width=0.05\linewidth, valign=m]{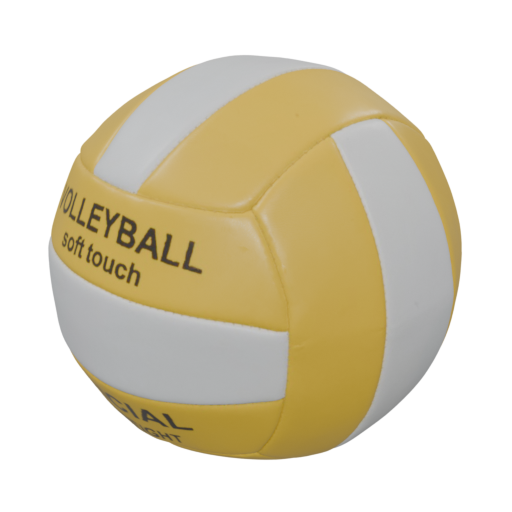} &
            \includegraphics[width=0.05\linewidth, valign=m]{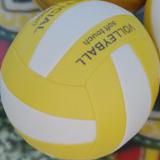} &
            \includegraphics[width=0.04\linewidth, valign=m]{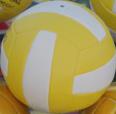} \\[-0.2em]
            {\tiny ext.} & {\tiny s=1.0} & {\tiny s=0.52}
          \end{tabular}
        & {\scriptsize yellow white volleyball}
        & {\scriptsize Yellow and white paneled volleyball with text}
        & {\scriptsize Volleyball composed of alternating yellow and white panels, with `VOLLEYBALL soft touch' printed in black lettering on a yellow section.} \\
        \midrule

        \multirow{2}{*}[2.45ex]{\includegraphics[width=0.17\linewidth, valign=m]{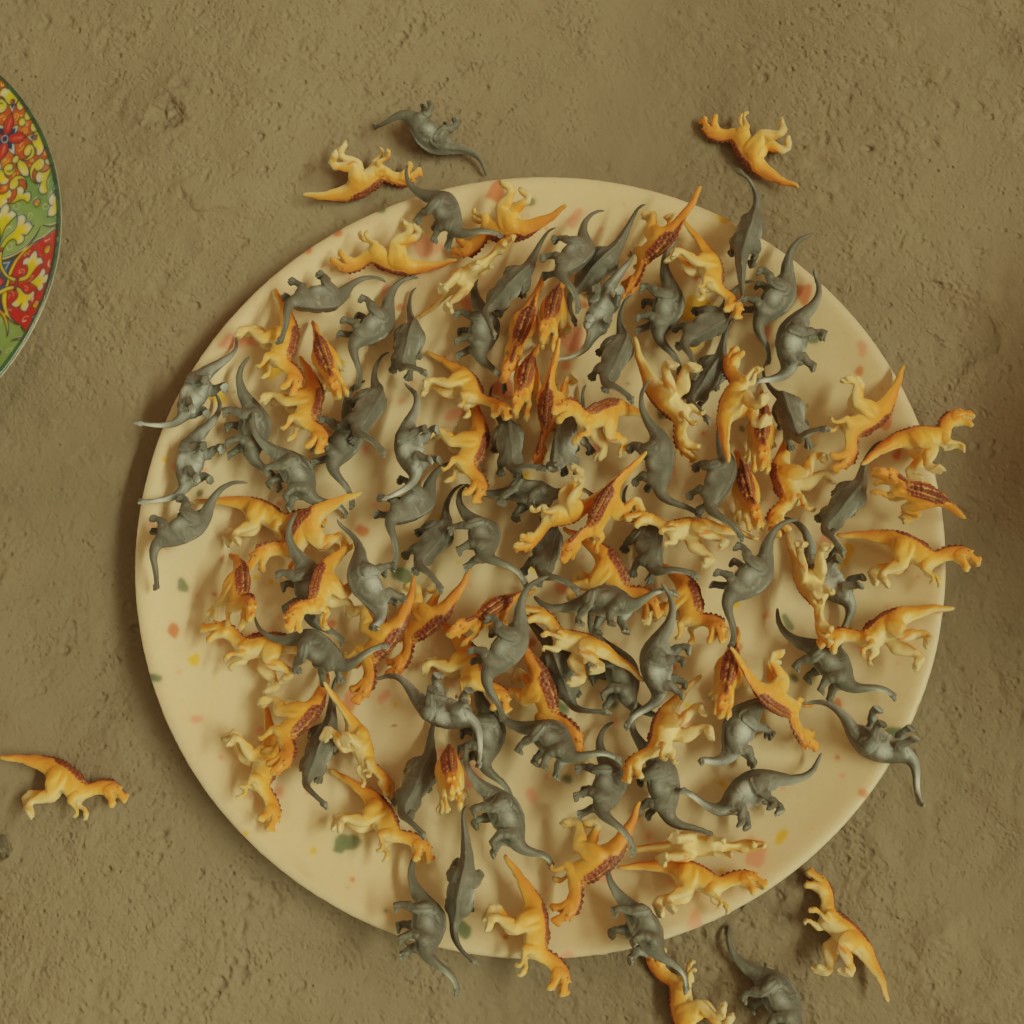}}
        & \begin{tabular}{@{}ccc@{}}
            \includegraphics[width=0.05\linewidth, valign=m]{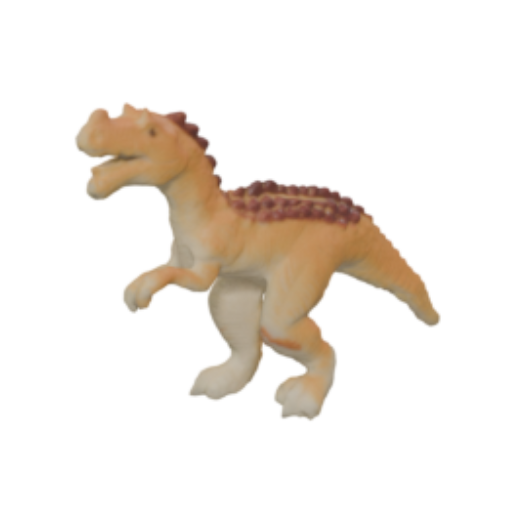} &
            \includegraphics[width=0.040\linewidth, valign=m]{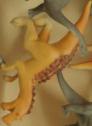} &
            \includegraphics[width=0.016\linewidth, valign=m]{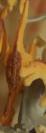} \\[-0.2em]
            {\tiny ext.} & {\tiny s=1.0} & {\tiny s=0.50}
          \end{tabular}
        & {\scriptsize tan bumpy dinosaur figure}
        & {\scriptsize Light brown dinosaur toy with ridged back}
        & {\scriptsize Tan-colored dinosaur toy, featuring a textured, ridged back and a bipedal stance. Its head has distinctive horn-like bumps above its eyes.} \\
        \cmidrule(lr){2-5}

        & \begin{tabular}{@{}ccc@{}}
            \includegraphics[width=0.05\linewidth, valign=m]{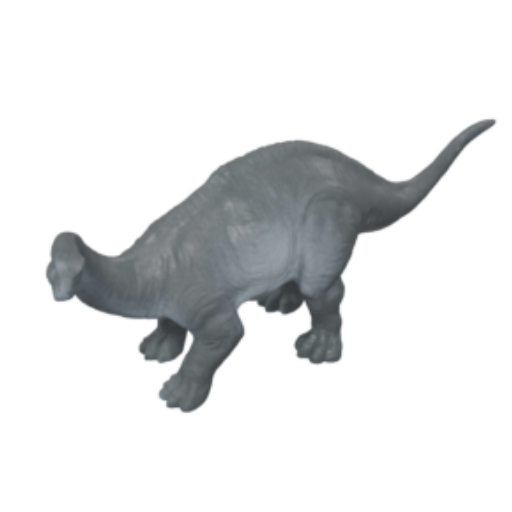} &
            \includegraphics[width=0.032\linewidth, valign=m]{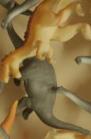} &
            \includegraphics[width=0.025\linewidth, valign=m]{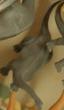} \\[-0.2em]
            {\tiny ext.} & {\tiny s=1.0} & {\tiny s=0.63}
          \end{tabular}
        & {\scriptsize grey apatosaurus figure}
        & {\scriptsize Grey apatosaurus toy with head down}
        & {\scriptsize Matte grey apatosaurus dinosaur toy with a long neck and tail, depicted with its head lowered towards the ground.} \\
        \bottomrule
    \end{tabularx}
    \caption{\textbf{Dataset features.} \acron~includes different exemplars and tiered description granularity.}
    \label{fig:data_features}
\end{figure*}

%% file: figs/annotations.tex
\begin{figure*}[t]
    \centering
    \setlength{\tabcolsep}{1.5pt} 
    
    \begin{tabularx}{\linewidth}{@{} C C C C C C @{}}
        \textbf{Image} & \textbf{Instance Segmentation} & \textbf{Class Segmentation} & \textbf{Bounding Boxes} & \textbf{Depth} & \textbf{Normal Map} \\
        \midrule
        
        \includegraphics[width=\linewidth, valign=m]{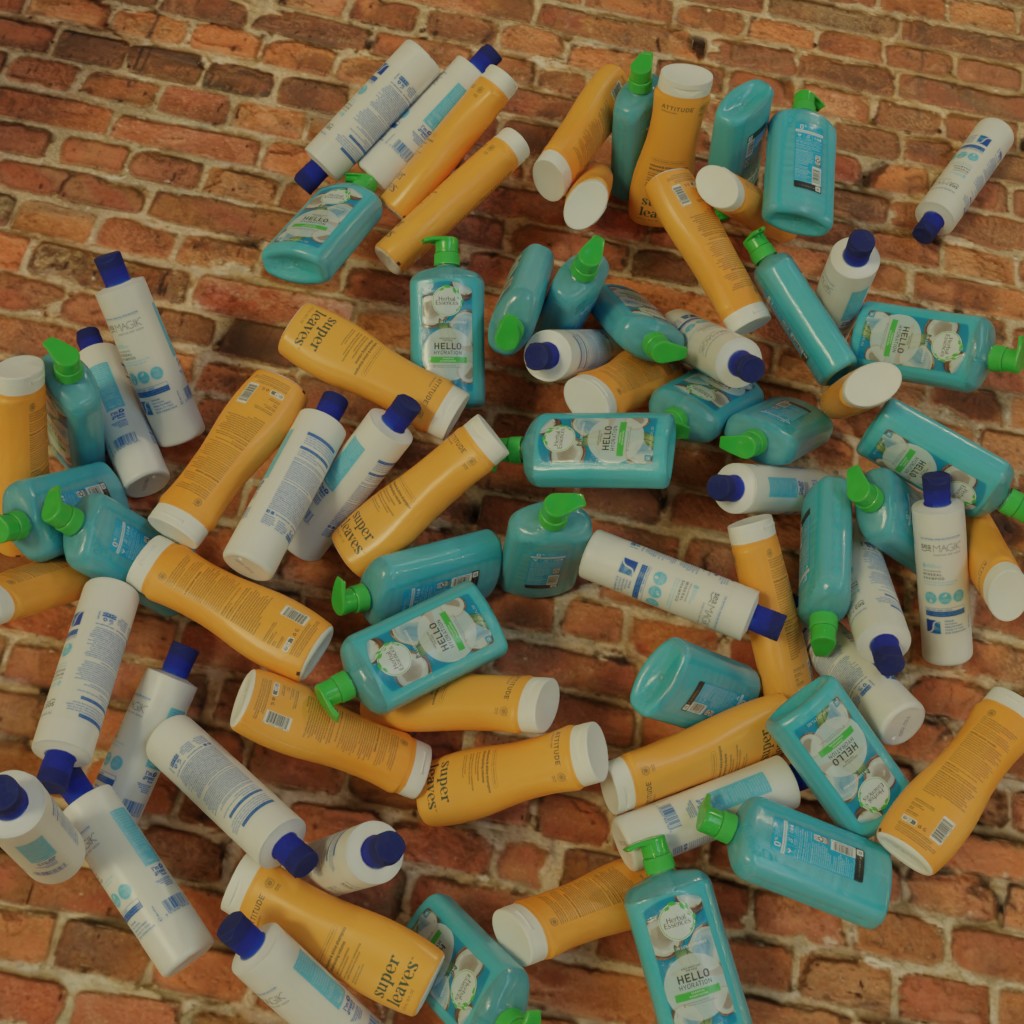} & 
        \includegraphics[width=\linewidth, valign=m]{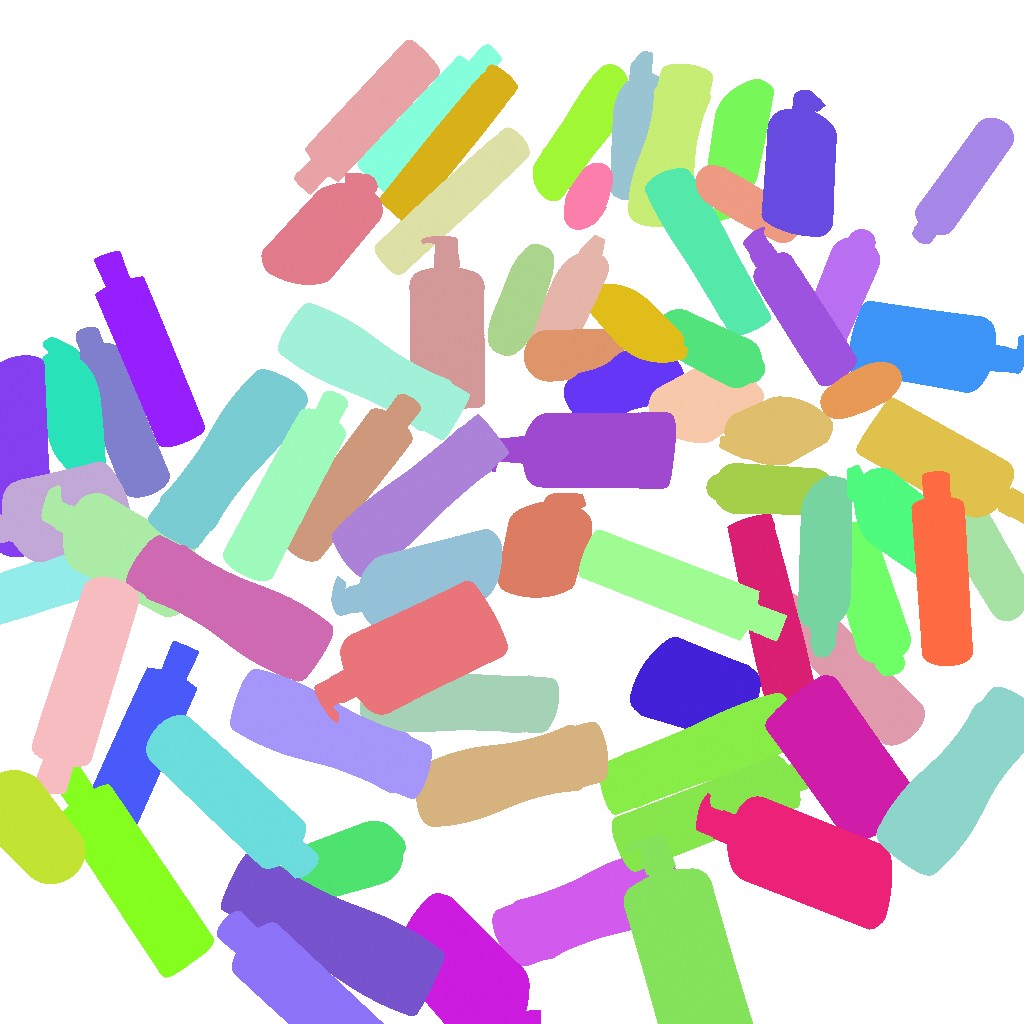} & 
        \includegraphics[width=\linewidth, valign=m]{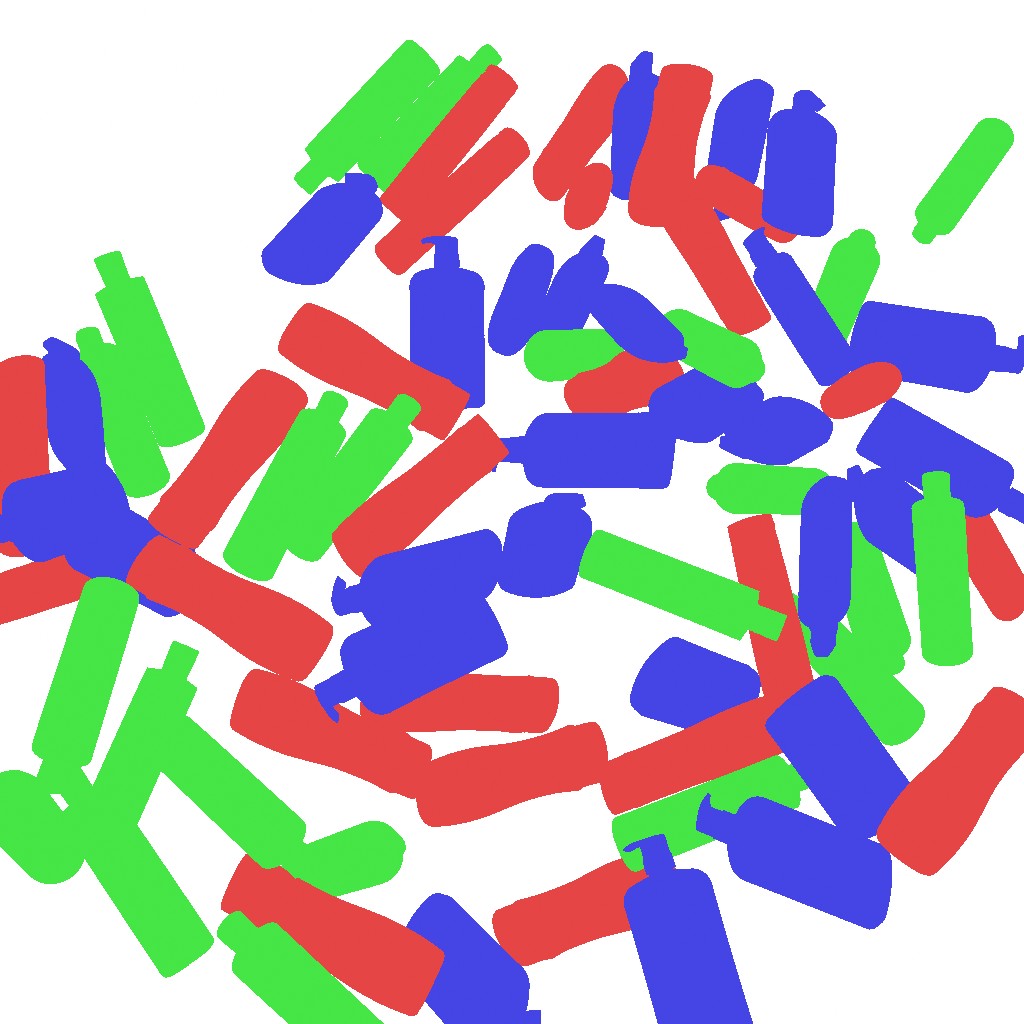} & 
        \includegraphics[width=\linewidth, valign=m]{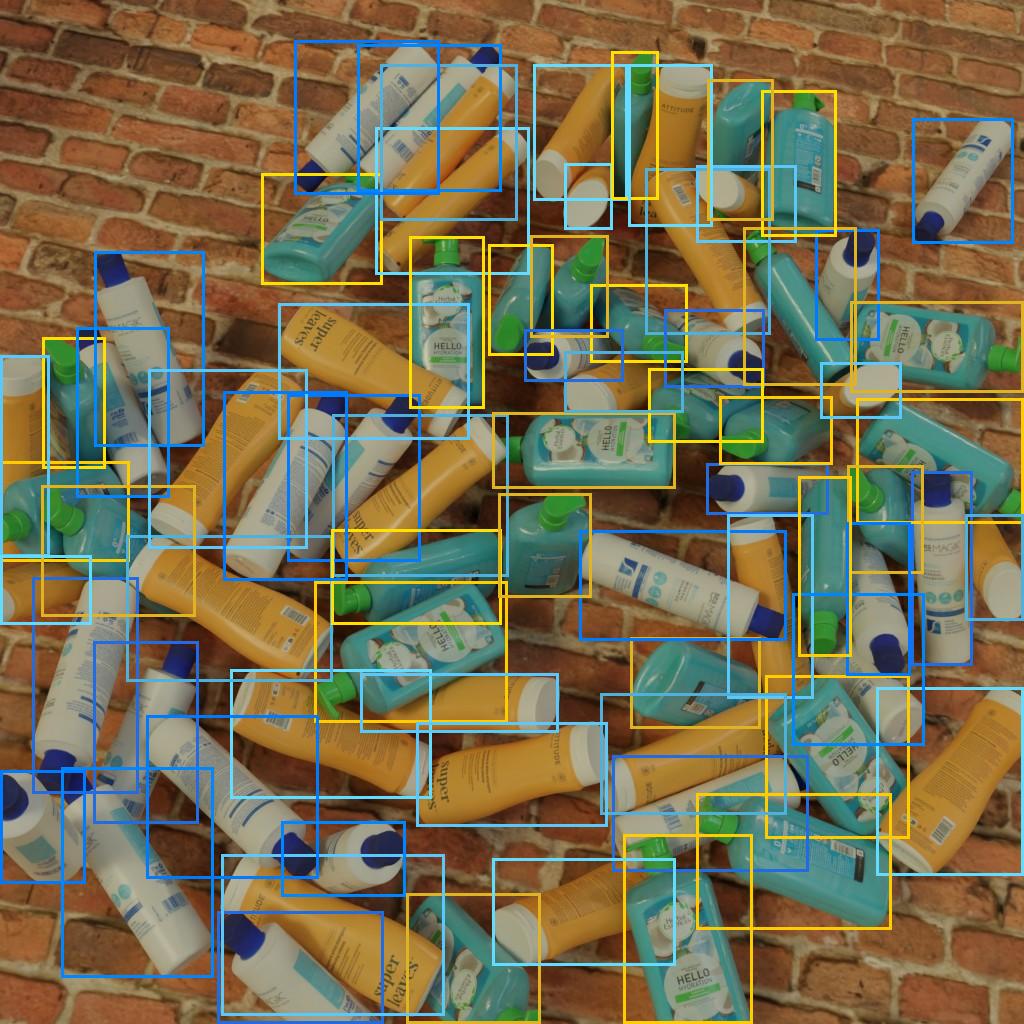} & 
        \includegraphics[width=\linewidth, valign=m]{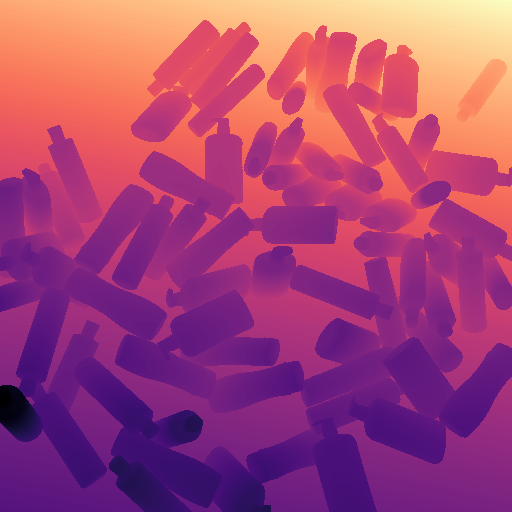} & 
        \includegraphics[width=\linewidth, valign=m]{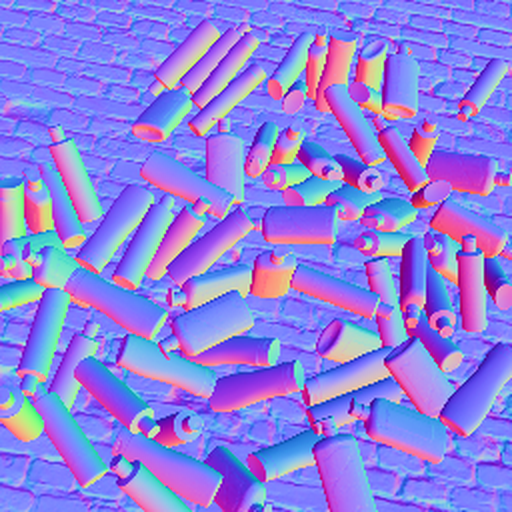} \\
        \addlinespace[2pt]

        \midrule
        
        \includegraphics[width=\linewidth, valign=m]{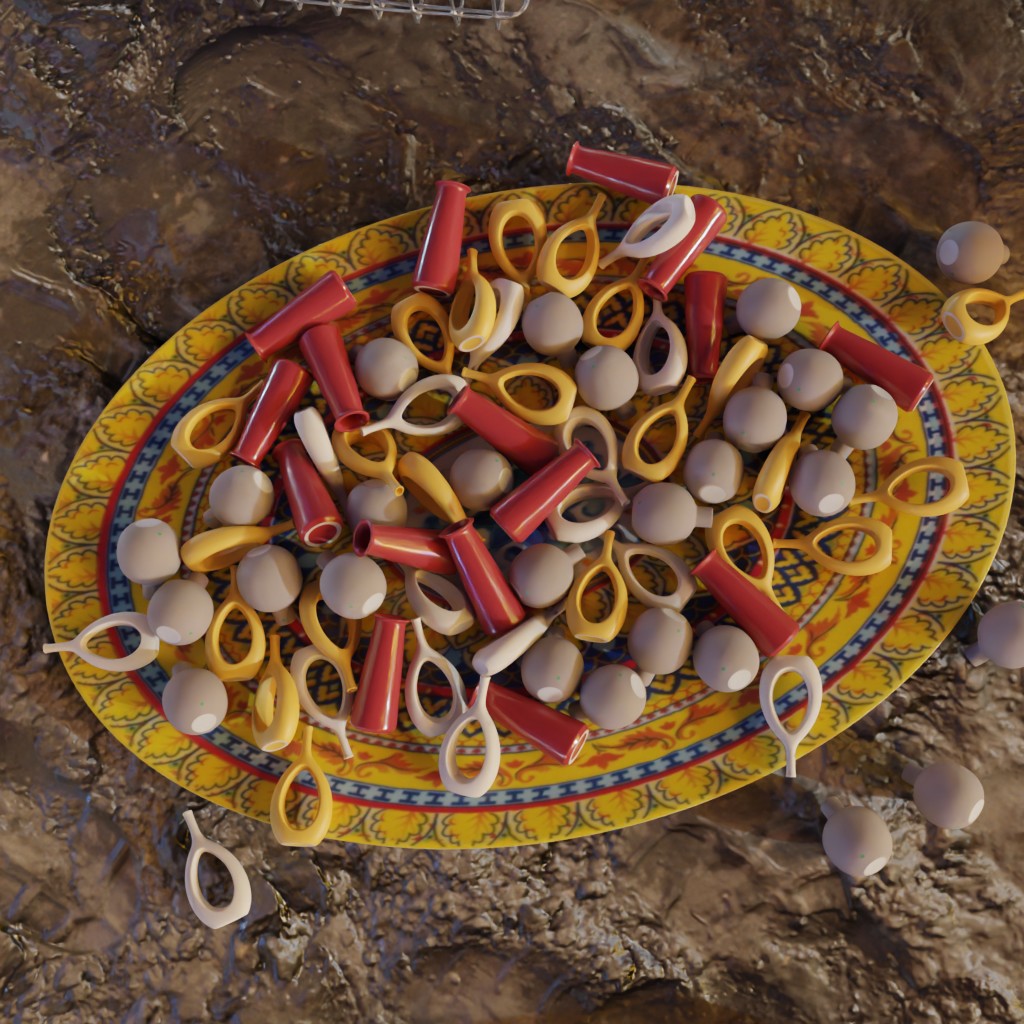} & 
        \includegraphics[width=\linewidth, valign=m]{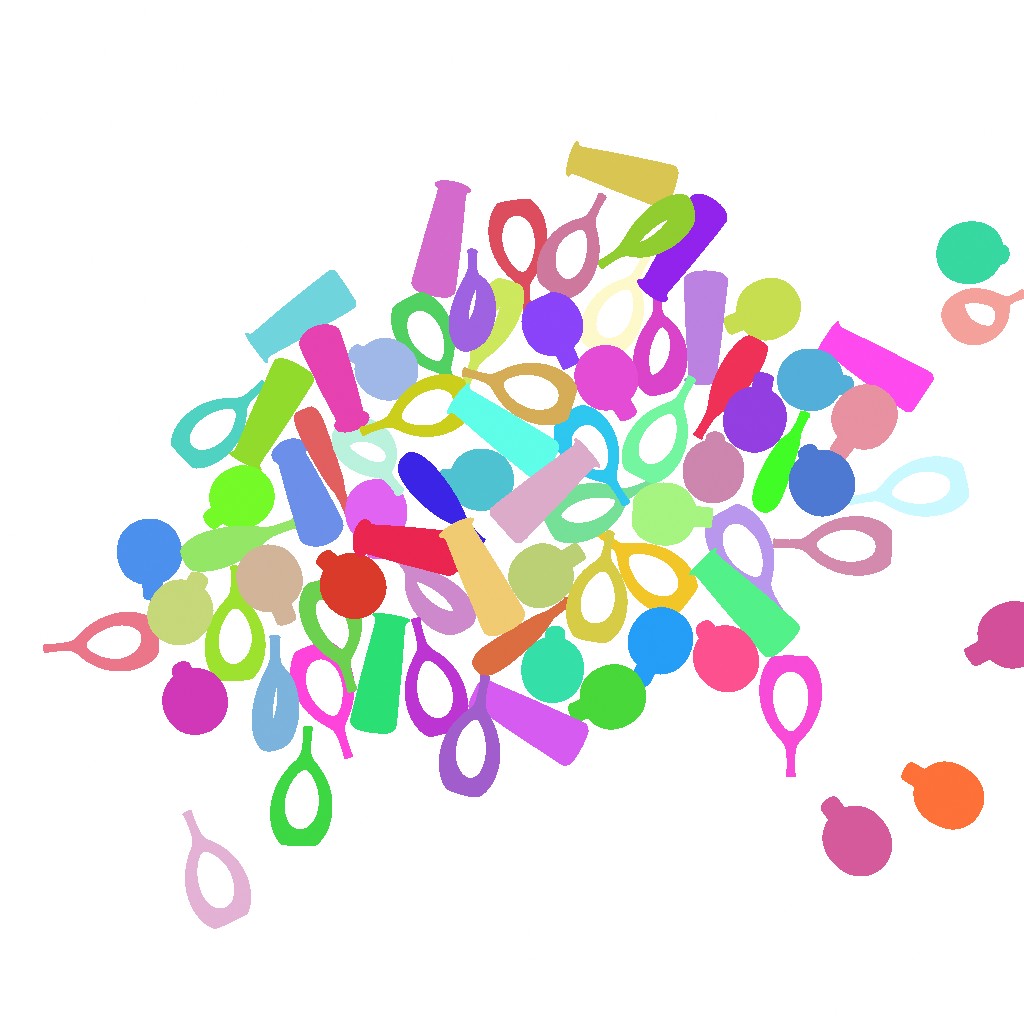} & 
        \includegraphics[width=\linewidth, valign=m]{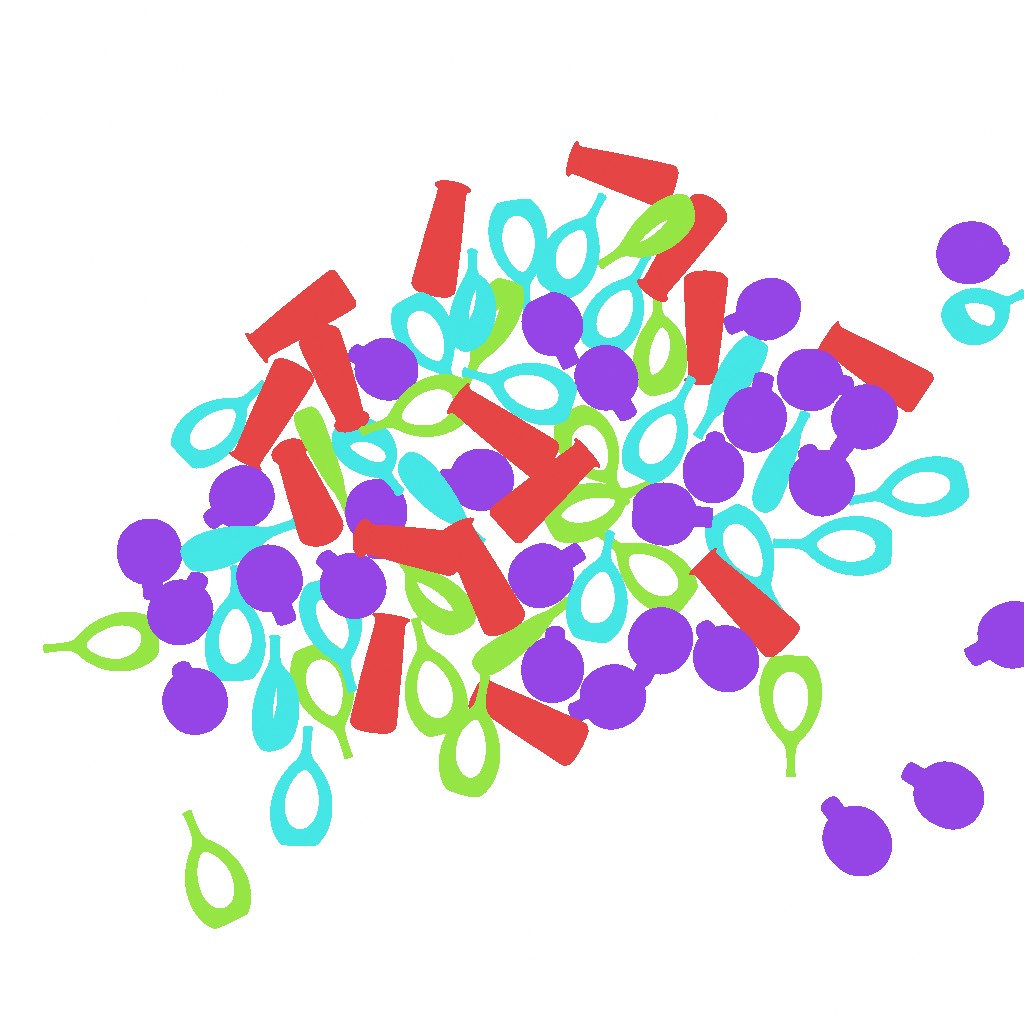} & 
        \includegraphics[width=\linewidth, valign=m]{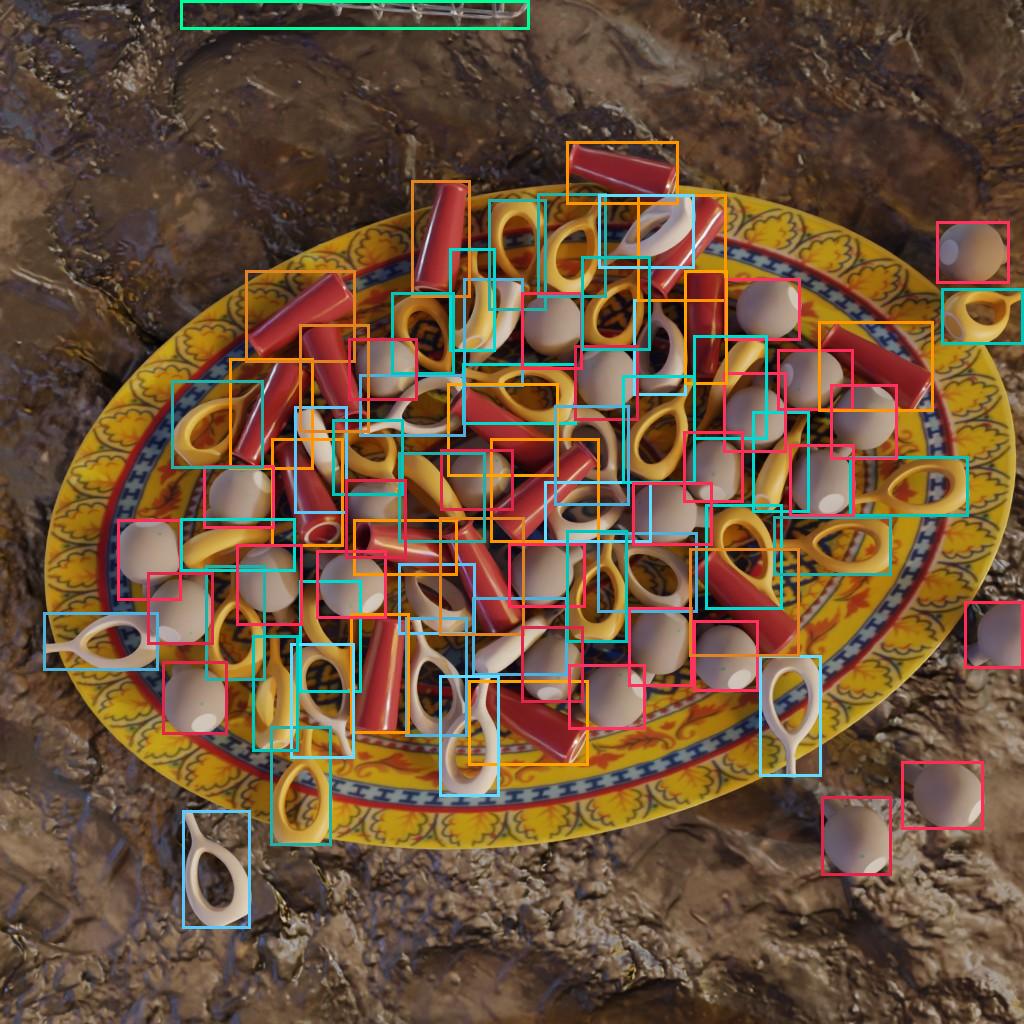} & 
        \includegraphics[width=\linewidth, valign=m]{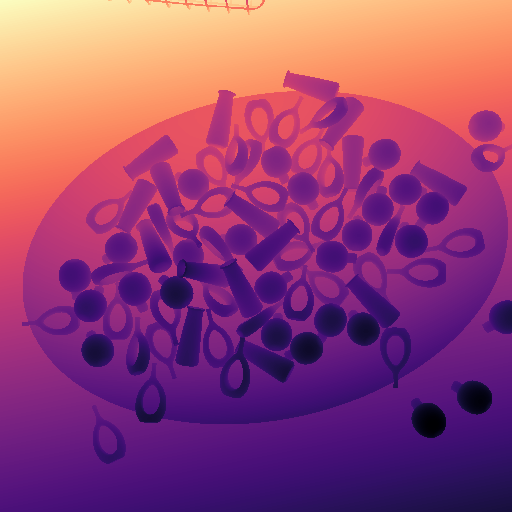} & 
        \includegraphics[width=\linewidth, valign=m]{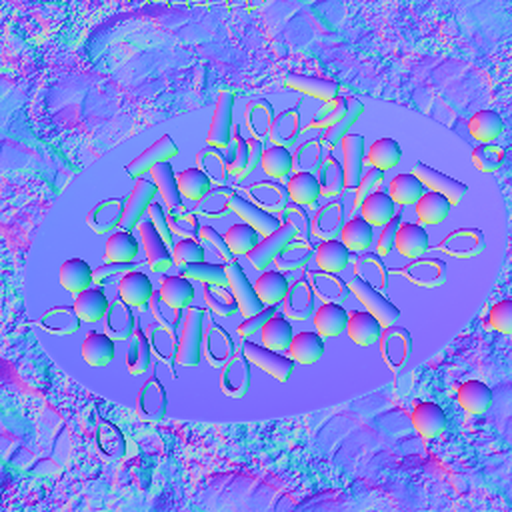} \\
        \addlinespace[2pt]

        \bottomrule
    \end{tabularx}
    \caption{\textbf{Dense ground-truth annotations.} Each sample is annotated with object localization labels, as well as depth and normal maps.} \label{fig:GT_maps}
\end{figure*}

%% file: figs/data_generator.tex
\begin{figure*}[t]
  \centering
  \includegraphics[width=\textwidth]{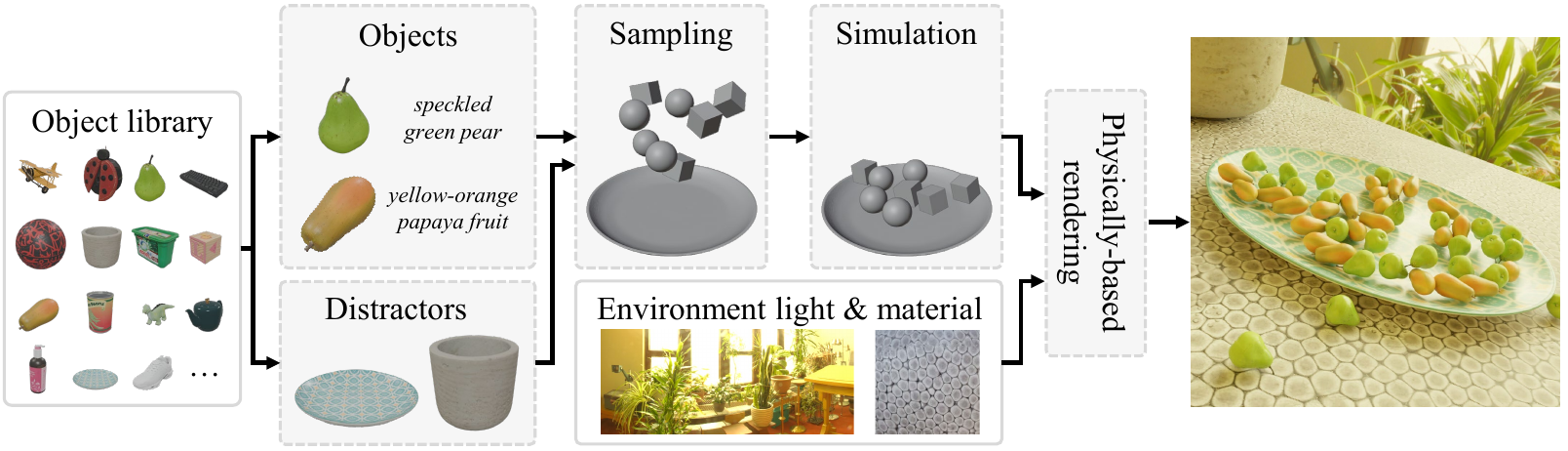}
  \caption{\textbf{Data generator.} Our generator samples objects, distractors, environment and camera placement to procedurally generate photorealistic training samples. All assets are issued from high-quality captures of real-world objects.}
  \label{fig:data_generator}
\end{figure*}

%% file: sec/4_experiments.tex
\section{Experiments}\label{sec:experiments}
We demonstrate that \acron~can be used to improve the counting accuracy of state-of-the-art counting models on real-world data in~\cref{sec:training}, and evaluate these models against the challenging \acron~benchmark in~\cref{sec:benchmarking}. Additional results, including sensitivity studies to the choice of exemplar or text prompt, are provided in our supplementary, as well as detailed evaluation protocols.

\subsection{Experimental setup}

\paragraph{Datasets.} In addition to \acron, two real-world counting datasets are used. FSC-147~\cite{ranjan2021learning} is the standard open-world counting dataset covering 147 classes and 6135 images with 7-3731 objects per image. In each image, only one type of object is annotated. PairTally~\cite{nguyen2025can} is a test set for fine-grained object counting with 681 dense images 
with two object types labeled per image. 

\paragraph{Metrics.} Following \cite{amini2024countgd, liu2022countr, AminiNaieni26, ranjan2021learning}, we report the Mean Absolute Error (MAE) and Root Mean Squared Error (RMSE) on FSC-147~\cite{ranjan2021learning} and PairTally~\cite{nguyen2025can}. On MixCount, we additionally report the Normalized Absolute Error (NAE) and the Accuracy $10\%$ (Acc$10\%$) defined as $NAE = \frac{1}{N}\sum_{i=1}^{N}\frac{|Y_{i} - \hat{Y_{i}}|}{Y_{i}} = \frac{1}{N}\sum_{i=1}^{N}NAE_{i}$ and $Acc10\%=\frac{\sum_{i=1}^{N}\mathbb{I}_{\{NAE_{i} \leq 0.1\}}}{N}$, where $N$ is the number of image-text pairs, $Y_{i}$ is the ground truth count, and $\hat{Y}_{i}$ is the predicted count. 

\paragraph{Implementation.} \CD{half of this could go to supplementary if needed}

For training on \acron, CountGD++~\cite{AminiNaieni26} is trained on a set composed of the FSC-147~\cite{ranjan2021learning} training images, the 1000 FSC-147 mosaicked images from \cite{AminiNaieni26}, and 1000 of the training images from \acron~with all the training hyperparameters borrowed from \cite{AminiNaieni26}. We use 1000 of the \acron~training images to ensure a balance between real and synthetic data and improve the training efficiency. We find that adding this small number of \acron~training images already yields substantial benefits. This model is denoted as CountGD++ (+~\acron). 

The original CountGD++ model is trained on just FSC-147 and the 1000 FSC-147 mosaicked images. At inference, the same adaptive cropping and test-time normalization used in \cite{amini2024countgd, AminiNaieni26} are applied for both CountGD++ and CountGD++ (+~\acron). For CountGD++ (+~\acron), the confidence threshold for FSC-147 is set to 0.31 using the FSC-147 validation set. For PairTally and \acron, the confidence threshold is set using the \acron~validation set for each prompt setting (0.75 when only positive prompts are provided and 0.4 when both positive and negative prompts are provided). For evaluating CountSE~\cite{liu2025countse}, CountGD~\cite{amini2024countgd}, and CountGD++~\cite{AminiNaieni26} on \acron, we use the published pretrained checkpoints from their official repositories.

\subsection{Results - \acron~value for training}\label{sec:training}
In \cref{fig:data_gap}, we show that training on \acron~remedies common failure modes of state-of-the-art counting models. In (a), \acron~enables CountGD++~\cite{AminiNaieni26} to distinguish between differences in object size. In (b), \acron~helps CountGD++ identify each pair of sunglasses as a single unit, instead of counting each self-similar lens, a well-known issue with counting models~\cite{amini2023open}. In (c), \acron~enables CountGD++ to better ignore repetitive backgrounds and focus on the specified object. In \cref{tab:mixcount_train}, we show that these improvements translate to significant quantitative gains on real-world data, reducing the MAE by 20.14\% on FSC-147~\cite{ranjan2021learning} and by 18.3\% on PairTally~\cite{nguyen2025can}.

\begin{table*}[h!]
\begin{center}
{\fontsize{8}{10}\selectfont\begin{NiceTabular}{l|c|c|c|c|c|c|c|c} 
   \hline
    Method & \multicolumn{4}{c}{Prompt} & \multicolumn{2}{c}{FSC-147 Test} & \multicolumn{2}{c}{PairTally Test}\\
    & $t^{+}$ & $B^{+}_{int}$ &$t^{-}$ & $B^{-}_{int}$ & MAE $\downarrow$ & RMSE $\downarrow$ & MAE $\downarrow$ & RMSE $\downarrow$\\
   \hline
    GeCo~\cite{pelhan2024novel} & \xmark & \cmark & \xmark & \xmark & 7.91 & 54.28 & 50.24 & -\\
    DAVE~\cite{Pelhan_2024_CVPR} & \xmark & \cmark & \xmark & \xmark & 8.66 & 32.36 & 47.37 & - \\
    CountGD~\cite{amini2024countgd} & \cmark & \cmark & \xmark & \xmark & 5.74 & 24.09 & 46.67 & 70.85 \\
    CountGD++~\cite{AminiNaieni26} & \cmark & \cmark & \xmark & \xmark &  5.86 & 18.79 &  46.41 & 69.52\\
    CountGD++ (+~\acron) & \cmark & \cmark & \xmark & \xmark & \textbf{4.68} & \textbf{15.66} & 43.36 & 67.02\\
    CountGD++~\cite{AminiNaieni26} &  \cmark & \cmark & \cmark & \cmark & - & - & 35.33 & 60.83\\
    CountGD++ (+~\acron) &  \cmark & \cmark & \cmark & \cmark & - & - & \textbf{28.87} & \textbf{54.82}\\
    
    \hline
\end{NiceTabular}}
\vspace{-1mm}
\caption{\label{tab:mixcount_train} Results on the {\bf FSC-147~\cite{ranjan2021learning}} and {\bf PairTally~\cite{nguyen2025can}} real-world test sets, with positive text ($t^{+}$), 3 positive visual exemplars from inside each image ($B^{+}_{int}$), negative text ($t^{-}$), 3 negative visual exemplars from inside each image ($B^{-}_{int}$). Note that negative prompts are not available for FSC-147, and PairTally did not report RMSE for all models. 
}
\end{center}
\end{table*}

\subsection{Results - \acron~value for benchmarking}\label{sec:benchmarking}

In \cref{tab:mixcount_eval}, we show that existing state-of-the-art counting models CountSE~\cite{liu2025countse}, CountGD~\cite{amini2024countgd}, and CountGD++~\cite{AminiNaieni26} trained on FSC-147~\cite{ranjan2021learning} perform poorly in the mixed-object setting. The highest Acc10\% 0.33 is achieved by CountGD++ given both positive and negative prompts. 
By training on the \acron~training set, the Acc10\% is improved to 0.69, and the NAE is reduced to 0.14. Without both positive and negative prompts, \acron~remains challenging for existing counting models.
We provide additional details on our benchmarking protocol and many qualitative examples of applying CountGD++ (+~\acron) to the \acron~test set in our supplementary.

\begin{table*}[h!]
\begin{center}
{\fontsize{8}{10}\selectfont\begin{NiceTabular}{l|c|c|c|c|c|c|c|c} 
   \hline
    Method & \multicolumn{4}{c}{Prompt} & \multirow{2}{*}{MAE $\downarrow$} & \multirow{2}{*}{RMSE $\downarrow$} & \multirow{2}{*}{NAE $\downarrow$} & \multirow{2}{*}{Acc10\% $\uparrow$}\\
    & $t^{+}$ & $B^{+}_{int}$ &$t^{-}$ & $B^{-}_{int}$ & &\\
   \hline
    CountSE~\cite{liu2025countse} & \cmark & \xmark & \xmark & \xmark & 32.49 & 41.49 & 7.23 & 0.06\\
    CountGD~\cite{amini2024countgd} & \cmark & \xmark & \xmark & \xmark & 33.00 & 42.25 & 5.64 & 0.05\\
    CountGD++~\cite{AminiNaieni26} & \cmark & \xmark & \xmark & \xmark & 26.77 & 36.20 & 2.73 & 0.09 \\
    CountGD~\cite{amini2024countgd} & \cmark & \cmark & \xmark & \xmark & 41.83 & 54.42 & 11.09 & 0.01 \\
    CountGD++~\cite{AminiNaieni26} & \cmark & \cmark & \xmark & \xmark & 34.55 & 46.57 & 4.60 & 0.06\\
    CountGD++~\cite{AminiNaieni26} &  \cmark & \cmark & \cmark & \cmark & 8.12 & 22.83 & 1.21 & 0.33\\
    \midrule
    CountGD++ (+ \acron) & \cmark & \xmark & \xmark & \xmark & 14.63 & 22.14 & 1.02 & 0.09\\
    CountGD++ (+ \acron) & \cmark & \cmark & \xmark & \xmark & 12.16 & 20.08 & 1.02 & 0.23\\
    CountGD++ (+ \acron) &  \cmark & \cmark & \cmark & \cmark & \textbf{2.11} & \textbf{4.73} &\textbf{ 0.14} & \textbf{0.69}\\
    \hline
\end{NiceTabular}}
\vspace{-1mm}
\caption{\label{tab:mixcount_eval} Results on our {\bf \acron} test set. The symbols for provided prompts are: positive text ($t^{+}$), up to 3 positive visual exemplars from inside each image ($B^{+}_{int}$), negative text ($t^{-}$), up to 3 negative visual exemplars from inside each image for each negative category ($B^{-}_{int}$).}
\end{center}
\end{table*}

%% file: sec/5_conclusion.tex
\section{Discussion and future works}

We have released the \acron~dataset, the first synthetic, large-scale, photorealistic and diverse training and evaluation dataset for open-vocabulary counting constructed from captures of real-world objects. As demonstrated in \cref{sec:experiments}, \acron~will directly improve future counting models and the systems that rely on them. Its new features, including external exemplars and textual descriptions at varying levels of detail, make future work on prompt sensitivity and versatility possible, which will in turn lead to more robust, flexible, and practical visual counting systems. Finally, our versatile and fully automated data generator synthesizes perfectly labeled and diverse counting data at scale. It can easily be adapted to specific application areas to solve new computer vision problems in different domains.

%% file: sec/6_supplementary.tex
\appendix

\section{Additional experiments}
In this section we present and discuss important additional results which were not added to our main paper due to space limitations.

\subsection{Internal vs. external exemplars}

In real-world scenarios, it is often much more practical to generate a single exemplar once for a specific concept, and apply it as a prompt on many images. These images may be captured in very different lighting conditions and background environments.
In \Cref{tab:int_v_ext_exemp}, we analyze the sensitivity of CountGD++ (+~\acron), the best performing model, to internal versus external exemplars, such as the ones displayed in \cref{fig:data_features}. \Cref{tab:int_v_ext_exemp} shows that the counting accuracy of CountGD++ (+~\acron) significantly decreases when switching from internal to external exemplars. Internal exemplars come from inside the image and, thus, factor in lighting, relative size, and other image-specific effects. On the other hand, external exemplars are independent of a particular image. As a result, they often provide less information than internal exemplars do. 
We hope our work will enable future research that will quantify and address this gap, which has important consequences in practical applications of visual counting models.

\begin{table*}[h!]
\begin{center}
{\fontsize{8}{10}\selectfont\begin{NiceTabular}{l|c|c|c|c} 
   \hline
    Exemplar Type & MAE $\downarrow$ & RMSE $\downarrow$ & NAE $\downarrow$ & Acc10\% $\uparrow$\\
    \hline
   Internal & 2.81 & 6.48 & 0.19 & 0.66\\
   External & 5.53 & 10.90 & 0.35 & 0.50 \\
   \hline
\end{NiceTabular}}
\vspace{-1mm}
\caption{\label{tab:int_v_ext_exemp} Results on our {\bf \acron} test set using internal versus external exemplars. Here, CountGD++ (+~\acron) is prompted with positive text, negative text, 1 positive exemplar, and 1 negative exemplar.}
\end{center}
\end{table*}

\subsection{Exemplar score sensitivity}

Selecting an exemplar manually for each object and each image can be a tedious task in a number of real-world applications. Recent methods~\cite{liu2025countse,AminiNaieni26} have proposed techniques to automatically detect visual exemplars from only a text prompt. However, the exemplar quality is typically reduced when compared with manually picked exemplars.

In \cref{fig:exemplar_sensitivity}, we employ the exemplar scores from \acron~to evaluate the sensitivity of our best counting model with regards to the choice of exemplar. The results demonstrate that picking a non-optimal exemplar increases the MAE by $0.13$ or a relative error increase of around $4.7\%$. This represents a significant degradation that can lead to critical failures in many real-world settings. By providing a wide variety of internal exemplars with varying scores, \acron~enables future work to mitigate this issue and make counting models more robust to the exact choice of visual exemplar.

\input{figs/exemplar_sensitivity}

\subsection{Varying levels of detail in text prompts}

In particularly challenging scenarios, end users may be tempted to provide overly detailed descriptions in an attempt to help the model distinguish between two similar classes of objects. Conversely, very short descriptions can be used in other settings, and the downstream counting models should ideally be robust to both cases.

In \Cref{tab:text_prompt_detail}, we analyze the sensitivity of CountGD++ (+~\acron), the best performing model, to different levels of detail in the text prompt. Text prompts in \acron~are tagged as `short,' `concise,' and `detailed' corresponding to increasing levels of descriptiveness. Examples are shown in \cref{fig:data_features}. \Cref{tab:text_prompt_detail} shows that CountGD++ (+~\acron) performs the best when given `concise' prompts. `concise' prompts are more detailed than `short' prompts, and, thus, provide more information to the model about the object to count. Because CountGD++ (+~\acron) was trained on the `short' prompts, the `detailed' prompts may deviate too much from its training distribution causing the error to increase. Future work could focus on training on more descriptive prompts to enable the model to take advantage of additional information about the object.

\begin{table*}[h!]
\begin{center}
{\fontsize{8}{10}\selectfont\begin{NiceTabular}{l|c|c|c|c} 
   \hline
    Level of Text Prompt Detail & MAE $\downarrow$ & RMSE $\downarrow$ & NAE $\downarrow$ & Acc10\% $\uparrow$\\
    \hline
   Short & 6.14 & 11.91 & 0.43 & 0.45\\
   Concise & 4.81 & 9.23 & 0.40 & 0.50\\
   Detailed & 6.61 & 12.32 & 0.47 & 0.43\\
   \hline
\end{NiceTabular}}
\vspace{-1mm}
\caption{\label{tab:text_prompt_detail} Results on our {\bf \acron} test set using different levels of detail in the text prompts. CountGD++ (+~\acron) given only positive and negative text is used.}
\end{center}
\end{table*}

\section{Additional dataset statistics}

\begin{figure}[htbp]
    \centering
    \begin{subfigure}[b]{0.48\textwidth}
        \centering
        \includegraphics[width=\textwidth]{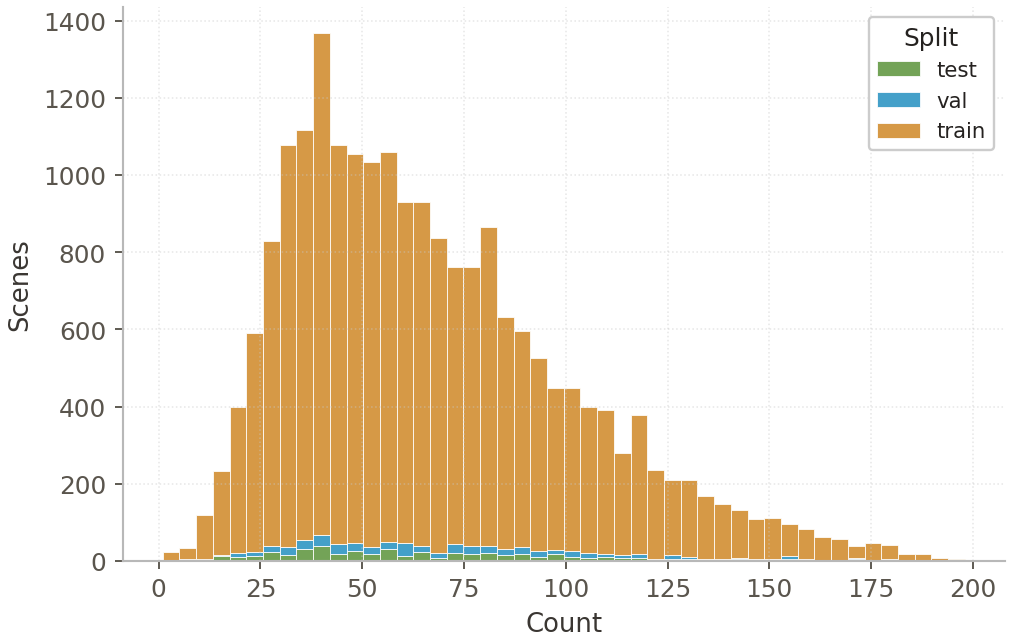}
    \end{subfigure}
    \hfill 
    \begin{subfigure}[b]{0.48\textwidth}
        \centering
        \includegraphics[width=\textwidth]{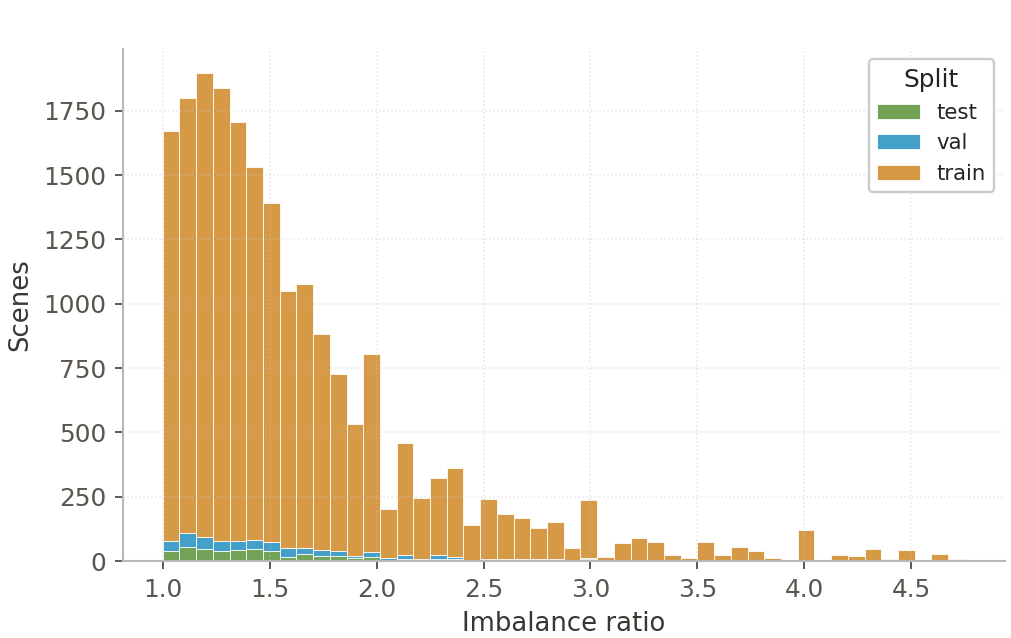}
    \end{subfigure}
    \caption{\textbf{Dataset statistics.}}
    \label{fig:statistics}
\end{figure}

Along with this paper, we release the first version of the \acron~dataset. This release includes 58,000 scenes with  over 4 million objects to count at an average of 67 per scene. 1522 different objects can be found in the dataset, all of which were manually filtered from DTC~\cite{dong2025digital} to remove duplicates and indistinguishable objects.

We display additional histogram statistics in \cref{fig:statistics}. These numbers show the range of objects present per scene, between 1 and 200, as well as the distribution of class imbalance in our dataset. The latter is computed for each scene by dividing the number of instances of the most common object by the number of instances of the least common object class.

\paragraph{Compute resources.}
Our dataset is entirely synthesized on 12-core CPUs, specifically \textit{Intel(R) Xeon(R) Gold 6240 18-Core 2.60GH Processor}. This has the considerable benefit of cutting down costs by a factor of 100 when compared with GPUs, and significantly reduces energy consumption. While it is possible to leverage a GPU for the final rendering of each scene, we found that this makes little difference as the time cost is dominated by the simulation and filtering of the objects.

Furthermore, each scene is generated in 2 minutes on average, and a total of 38 hours of 50 workers of 12 cores each. At a per-core cost of around 0.004\$ per hour, this amounts to less than 70\$. While this is not a negligible cost, it remains greatly inferior to the cost of experimenting with and training visual counting models. This makes our data generator particularly attractive to further generate even larger scale or customized datasets in future research.

\section{MixCount evaluation protocol}

We provide additional details on the benchmarking protocol used to produce the results in~\Cref{tab:mixcount_eval} of our main paper. The ground-truth annotations are provided in COCO~\cite{lin2014microsoft} format, where each object is listed as a bounding box with an object type and a reference image. We thus group objects of the same type within one frame to compute their total count, including the distractor objects present in the background.

The first class of methods, which are above the horizontal bar in~\Cref{tab:mixcount_eval}, have no access to the training or validation sets of \acron. They are evaluated on the \acron~test set without any additional change.
In the second part of~\Cref{tab:mixcount_eval}, we report methods in the \textit{+ MixCount} setting. These methods have access to the training and validation sets of MixCount. They may train on any number of training images and use the validation set for hyperparameter tuning.

\input{figs/additional_samples}

\section{Additional samples}

We display additional samples from our test set in~\cref{fig:pred_samples}. We also display an overlay of the predictions of our best model, CountGD++ (+ MixCount) with positive and negative prompts. The test set features a large diversity of objects, with plastic fruits, markers, toy figurines, wooden blocks, and many more classes. Note that train, val and test objects are completely distinct.

\section{External exemplars}

In~\cref{fig:external_exemplars}, we display several random external exemplars from our dataset. These exemplars are automatically generated in a canonical setting under a white background. Note that our dataset generator makes it straightforward to generate this kind of exemplar for arbitrary objects, as well as variations in different lighting environments and backgrounds. Our dataset also makes it easy to sample external exemplars from other images where the same object is present.

\input{figs/external_exemplars}

\section{\acron~generation parameters}

The majority of simulation parameters used to generate \acron~ can be freely adjusted to customize the data and generate a specific distribution of data. In this section, we go over the most important parameters and explain their role. First, we set global scene scale to $s = 10.0$, which is important for physical simulation to work as expected. Any geometric parameter described as scaled is multiplied by $s$. 

\paragraph{Scene context.}

For each scene, an HDR map is first sampled uniformly from the available group. 
The floor is set as a cube or cylinder and collidable. Floor shape and dimensions are sampled from configured ranges, then scaled by $s$. The floor texture UV scale is also randomized and sampled as $u \sim \mathcal{U}[0.04,0.1]$. A random material is then applied to the floor.

A container is sampled from the Digital Twin Catalog \cite{dong2025digital}, with a random chance of selecting no container at all. Containers are typically plates or trays, which do not greatly limit visibility but add visual complexity and realism. Crucially, we use a decimated mesh for collision detection while keeping the full-resolution mesh in renders. The container is randomly scaled on each axis independently, then scaled by $s$.

\paragraph{Object class construction and sampling.}
The number of counted objects is sampled per scene: $n \sim \text{UniformInteger}[30,200]$, and the number of object classes is sampled as $n_{\text{classes}} \sim \text{UniformInteger}[2,4]$.
For each class, a target size is sampled $d_{\max} \sim \mathcal{U}[0.045,0.12]\cdot s$. Class templates are normalized to this target size. At spawn time, each instance receives multiplicative jitter $\gamma \sim \mathcal{U}[0.98,1.02]$ to introduce variability in the objects.

When a container exists, spawn positions are sampled above an ellipse inscribed in the container axis-aligned bounding box (AABB). With spawn margin $m = 0.04\cdot s$, random points are uniformly sampled in the ellipse, and a random height is sampled from $z\sim\mathcal{U}[0.1,1.5]\cdot s$. If no container is present, XY is sampled from the global range \([-0.3,0.3]\cdot s\).

\paragraph{Physics simulation}

All counted objects, plus floor and container collider when present, are simulated with rigid-body dynamics over $t\in[0,300]$ frames. The solver uses 2 substeps per frame and $4$ solver iterations, which we found to be fast and precise enough. The active object collision shape is the object's convex hull, which is an approximation of the shape but didn't seem to have a significant impact in practice. 

\paragraph{Camera sampling and rendering}

Each scene uses a single camera. Camera pose is sampled on a hemisphere with elevation in \([20^\circ,80^\circ]\), in-plane roll jitter in \([-15^\circ,+15^\circ]\), and distance based on container size: $d = 1.2\cdot 
 \lVert\mathrm{diag}\mathrm{AABB}_{\text{container}})\rVert$, with additional relative jitter on the radius of \(\pm 15\%\).

Images are rendered at $H=W=1024$, and RGB rendering uses Cycles with \(32\) samples and denoising enabled. Mask-oriented passes use Eevee for efficiency. 

\paragraph{Visibility and quality filtering}

Objects are retained only if their origin projects inside every camera frame. Furthermore, for object \(i\), occlusion is estimated from visible vs full masks:
$$
\mathrm{occ}_i = 1-\frac{A_i^{\mathrm{visible}}}{A_i^{\mathrm{full}}}.
$$

An object is discarded if $\max_c \mathrm{occ}_{i,c} \ge v_t = 0.4.$ If all objects are removed, the scene is discarded and a new one is generated.

\paragraph{Distractor objects}

Distractors are enabled to add scene clutter. Their number is sampled as $n_{\text{dist}} \sim \text{UniformInteger}[2,8]$ and the base distractor size is $d_{\text{dist,max}} = 0.7\cdot s$ with a per-instance multiplier $\lambda\sim\mathcal{U}[0.5,1.2]$.
Their placement is constrained to ensure they do not collide with the objects or other distractors with AABB margins.

\paragraph{Exports}

For each rendered scene, the pipeline writes camera outputs and scene metadata. The metadata includes class definitions, per-object records, camera parameters, transform summaries, error diagnostics, and timing fields. 
After all scenes are generated, a dataset-level summary with aggregate statistics and timing is written. COCO annotations are exported with bounding-boxes. A run-level preview grid image is generated.

\section{Licensing}

Our dataset is published under the CC BY-SA licence. Users are free to share and adapt \acron, provided that they give appropriate credit and also distribute their contributions under the same license. We also recommend giving credits to the assets datasets~\cite{eppel2024vastextures,zaal2021polyhaven,dong2025digital,gardner2017learning} when applicable.

Our dataset uses assets issued from real-captures that are publicly available and have permissive licenses. Both Polyhaven~\cite{zaal2021polyhaven} and VasTexture~\cite{eppel2024vastextures} have CC0 license, and their assets are therefore free to download and use for any purpose.
The 3D object models from DTC are, following the \textit{Digital Twin Catalog Dataset License Agreement}, licensed under a variant of CC BY-SA, and cannot be sold or incorporated into a product to be sold. 
Finally, according to the \textit{LavalIndoor End User License Agreement}, the LavalIndoor~\cite{gardner2017learning} database is made available for research purposes and may not be redistributed. The use of data in publications is allowed. 

\section{Limitations}

Synthetic data provides scalable supervision, but its image-label distribution may differ from real data. We use a finite asset pool and certain sampling rules in the number of objects generated, their scale, or camera range. In our pipeline, we mitigate this issue by randomly sampling simulation parameters from large ranges, and employing large numbers of diverse assets. In particular, environment lights vary significantly across the dataset, ranging from dark scenes where the objects are barely visible to very bright and over saturated scenes.

Although annotations are internally consistent, they follow a specific definition of what is countable. Some real benchmarks may use slightly different conventions for partially visible, tiny, or ambiguous instances, which can create label mismatch even when predictions are visually plausible. However, previous datasets are notably inconsistent in this regard, and we provide the first dataset that guarantees visibility of all the objects to count.

\section{Societal impact and risks}

Reliable mixed-object counting has clear positive applications in industrial inspection, agriculture, inventory, and biomedical imaging, and our synthetic pipeline lowers the annotation cost barrier for these uses. The same capabilities can in principle support surveillance and discrimination; we mitigate the most direct version of this risk by restricting \acron~to non-human object categories. Practitioners should also note that synthetic training data can inherit biases from the underlying data generator, and validate on representative real-world distributions before deployment in critical settings.

%% file: figs/exemplar_sensitivity.tex
\begin{figure}[ht!]
    \centering
    \includegraphics[width=0.5\linewidth]{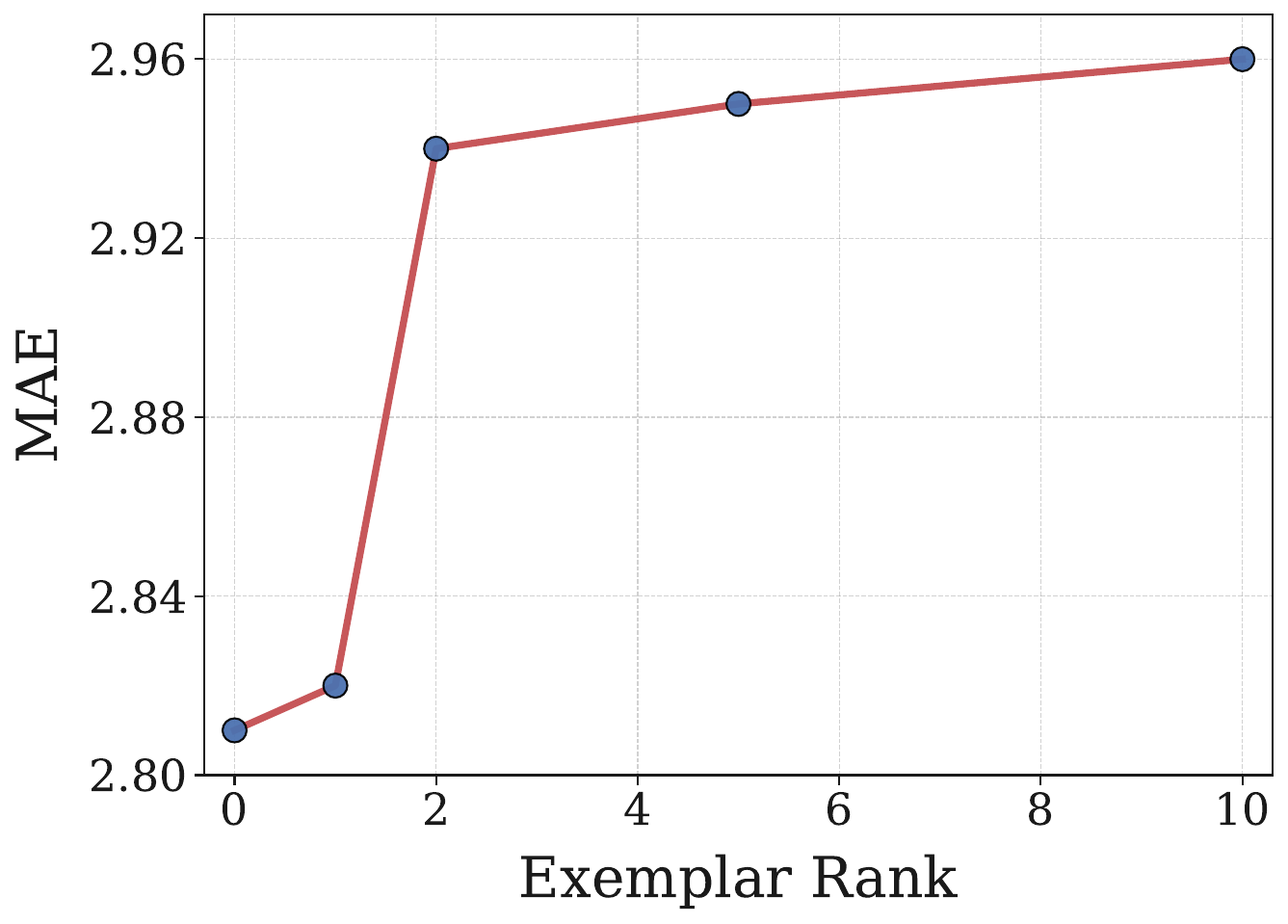}
    \caption{\textbf{Counting MAE vs. exemplar rank.} Counting MAE increases as lower-scoring exemplars are used, indicating that exemplar quality affects counting performance. Lower exemplar rank corresponds to higher exemplar score.}
    \label{fig:exemplar_sensitivity}
\end{figure}

%% file: figs/additional_samples.tex
\begin{figure*}[htbp]
  \centering
  \captionsetup[subfigure]{justification=centering, skip=0pt}
  \begin{subfigure}{0.235\textwidth}
    \centering
    \includegraphics[width=\textwidth]{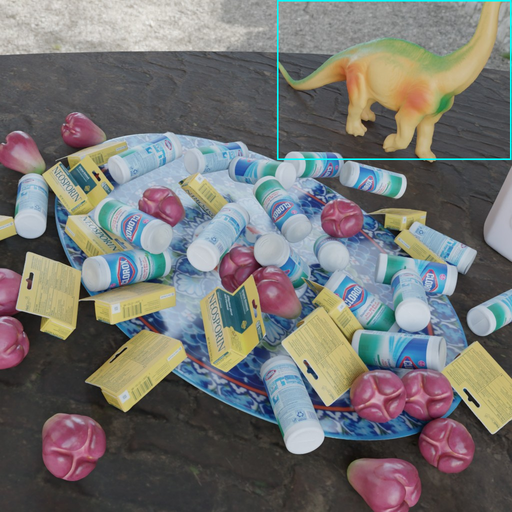}
    \vspace{-2.3ex}
    \caption*{{\scriptsize $N_{pred} = 1, N_{GT} = 1$} \\
               {\tiny ``Cream green dinosaur figure''}}
  \end{subfigure} \hfill
  \begin{subfigure}{0.235\textwidth}
    \centering
    \includegraphics[width=\textwidth]{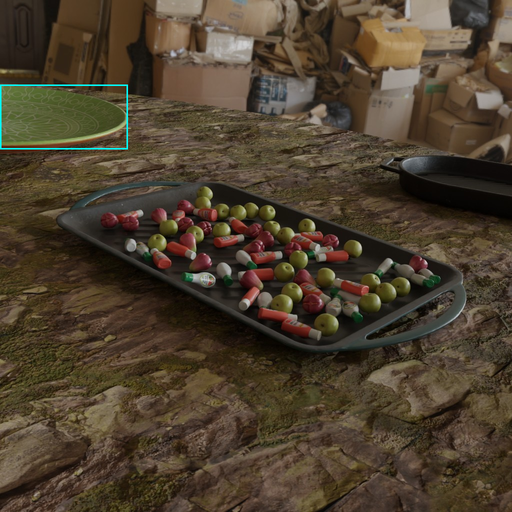}
    \vspace{-2.3ex}
    \caption*{{\scriptsize $N_{pred} = 1, N_{GT} = 1$} \\
               {\tiny ``Green patterned plate''}}
  \end{subfigure} \hfill
  \begin{subfigure}{0.235\textwidth}
    \centering
    \includegraphics[width=\textwidth]{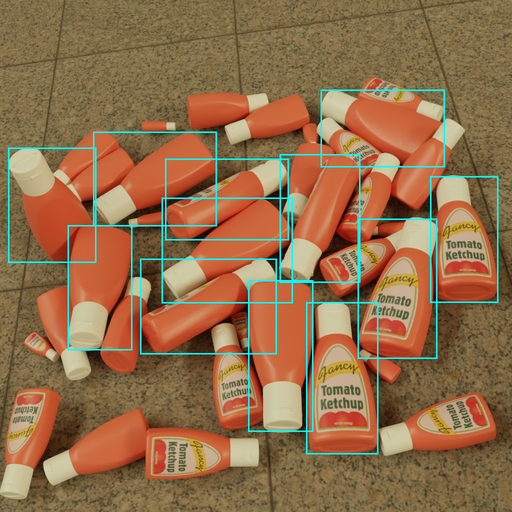}
    \vspace{-2.3ex}
    \caption*{{\scriptsize $N_{pred} = 12, N_{GT} = 12$} \\
               {\tiny ``Larger fancy tomato ketchup toy''}}
  \end{subfigure} \hfill
  \begin{subfigure}{0.235\textwidth}
    \centering
    \includegraphics[width=\textwidth]{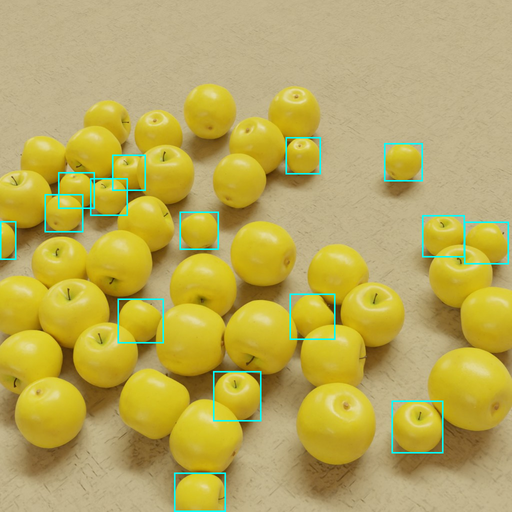}
    \vspace{-2.3ex}
    \caption*{{\scriptsize $N_{pred} = 15, N_{GT} = 15$} \\
               {\tiny ``Smaller golden yellow apple''}}
  \end{subfigure} \\[1.2ex]
  \begin{subfigure}{0.235\textwidth}
    \centering
    \includegraphics[width=\textwidth]{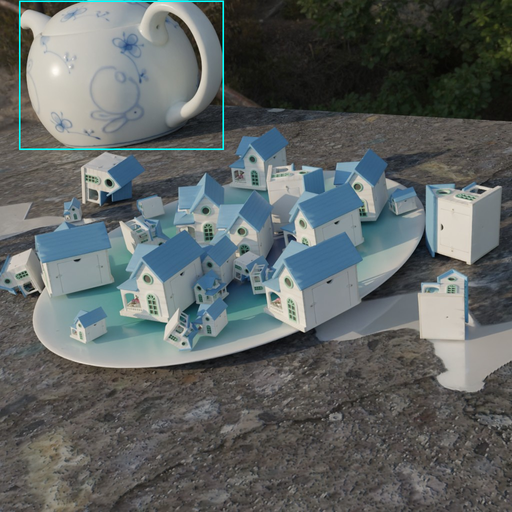}
    \vspace{-2.3ex}
    \caption*{{\scriptsize $N_{pred} = 1, N_{GT} = 1$} \\
               {\tiny ``White rabbit teapot''}}
  \end{subfigure} \hfill
  \begin{subfigure}{0.235\textwidth}
    \centering
    \includegraphics[width=\textwidth]{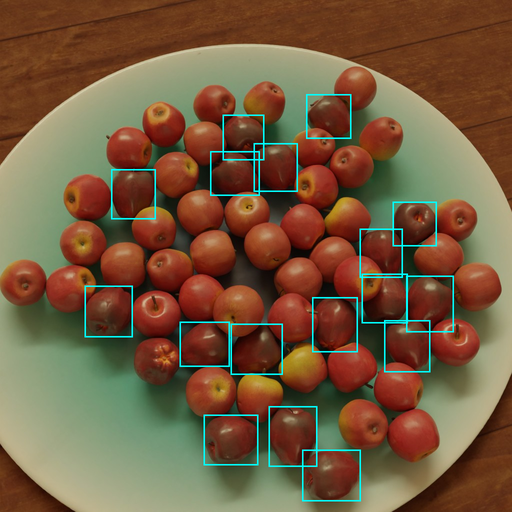}
    \vspace{-2.3ex}
    \caption*{{\scriptsize $N_{pred} = 17, N_{GT} = 18$} \\
               {\tiny ``artificial dark red apple''}}
  \end{subfigure} \hfill
  \begin{subfigure}{0.235\textwidth}
    \centering
    \includegraphics[width=\textwidth]{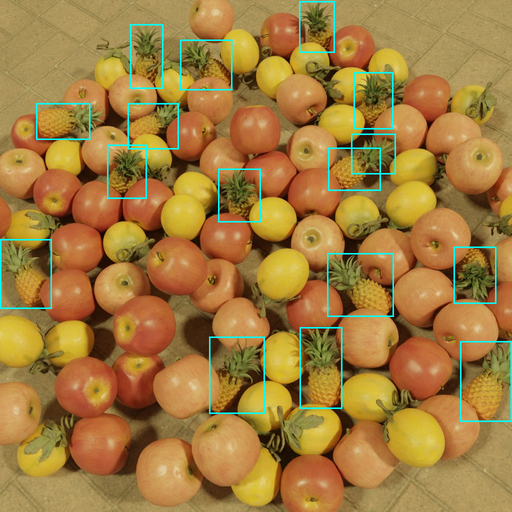}
    \vspace{-2.3ex}
    \caption*{{\scriptsize $N_{pred} = 16, N_{GT} = 15$} \\
               {\tiny ``artificial pineapple decoration''}}
  \end{subfigure} \hfill
  \begin{subfigure}{0.235\textwidth}
    \centering
    \includegraphics[width=\textwidth]{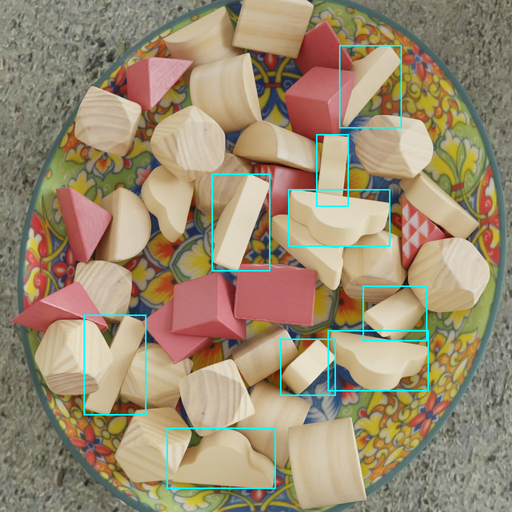}
    \vspace{-2.3ex}
    \caption*{{\scriptsize $N_{pred} = 9, N_{GT} = 12$} \\
               {\tiny ``beige cloud block''}}
  \end{subfigure} \\[1.2ex]
  \begin{subfigure}{0.235\textwidth}
    \centering
    \includegraphics[width=\textwidth]{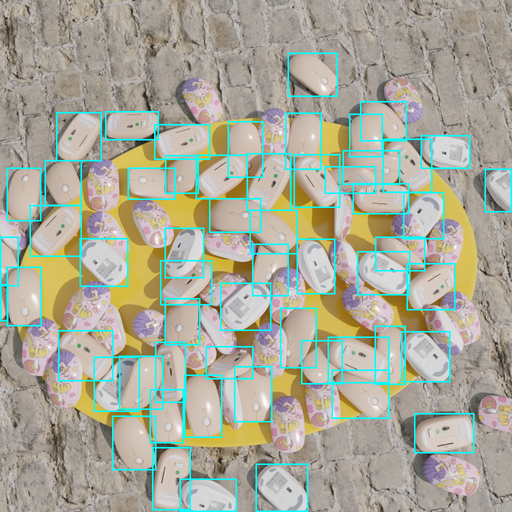}
    \vspace{-2.3ex}
    \caption*{{\scriptsize $N_{pred} = 57, N_{GT} = 45$} \\
               {\tiny ``beige computer mouse''}}
  \end{subfigure} \hfill
  \begin{subfigure}{0.235\textwidth}
    \centering
    \includegraphics[width=\textwidth]{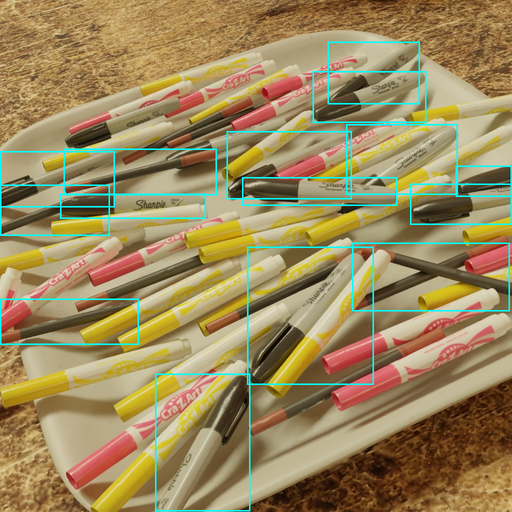}
    \vspace{-2.3ex}
    \caption*{{\scriptsize $N_{pred} = 15, N_{GT} = 13$} \\
               {\tiny ``black Sharpie marker''}}
  \end{subfigure} \hfill
  \begin{subfigure}{0.235\textwidth}
    \centering
    \includegraphics[width=\textwidth]{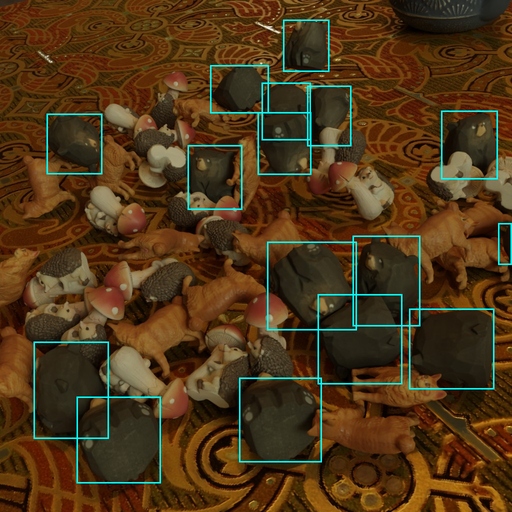}
    \vspace{-2.3ex}
    \caption*{{\scriptsize $N_{pred} = 16, N_{GT} = 16$} \\
               {\tiny ``black bear figurine''}}
  \end{subfigure} \hfill
  \begin{subfigure}{0.235\textwidth}
    \centering
    \includegraphics[width=\textwidth]{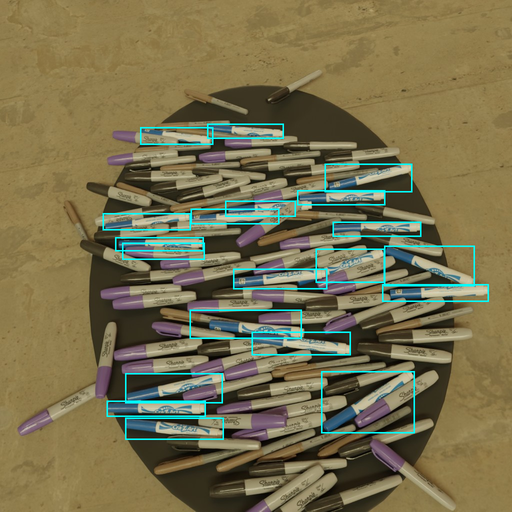}
    \vspace{-2.3ex}
    \caption*{{\scriptsize $N_{pred} = 20, N_{GT} = 19$} \\
               {\tiny ``blue Cra-Z-Art marker''}}
  \end{subfigure} \\[1.2ex]
  \begin{subfigure}{0.235\textwidth}
    \centering
    \includegraphics[width=\textwidth]{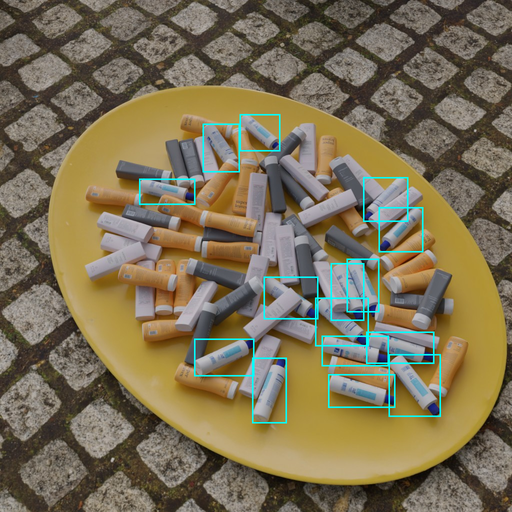}
    \vspace{-2.3ex}
    \caption*{{\scriptsize $N_{pred} = 15, N_{GT} = 15$} \\
               {\tiny ``blue-capped shampoo bottle''}}
  \end{subfigure} \hfill
  \begin{subfigure}{0.235\textwidth}
    \centering
    \includegraphics[width=\textwidth]{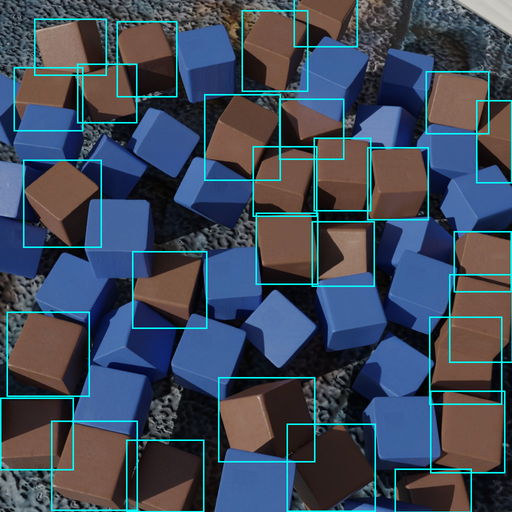}
    \vspace{-2.3ex}
    \caption*{{\scriptsize $N_{pred} = 28, N_{GT} = 28$} \\
               {\tiny ``brown wooden cube''}}
  \end{subfigure} \hfill
  \begin{subfigure}{0.235\textwidth}
    \centering
    \includegraphics[width=\textwidth]{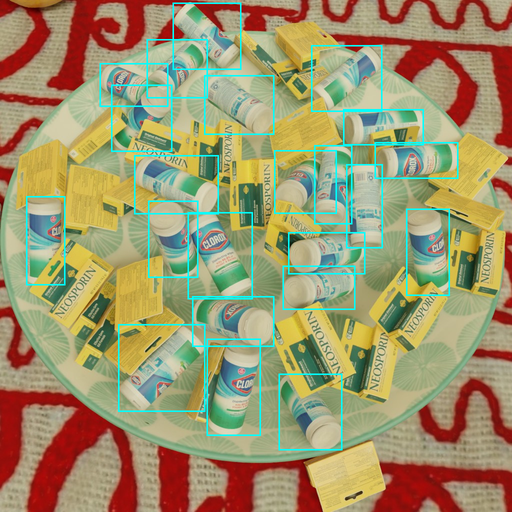}
    \vspace{-2.3ex}
    \caption*{{\scriptsize $N_{pred} = 22, N_{GT} = 23$} \\
               {\tiny ``clorox disinfecting wipes container''}}
  \end{subfigure} \hfill
  \begin{subfigure}{0.235\textwidth}
    \centering
    \includegraphics[width=\textwidth]{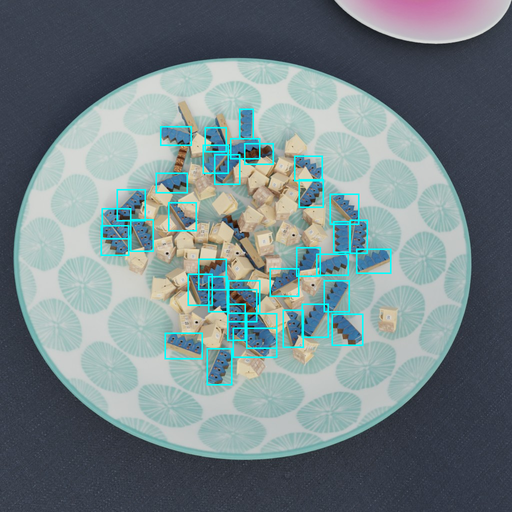}
    \vspace{-2.3ex}
    \caption*{{\scriptsize $N_{pred} = 36, N_{GT} = 41$} \\
               {\tiny ``connected birdhouses, metal roofs''}}
  \end{subfigure} \\[1.2ex]
  \begin{subfigure}{0.235\textwidth}
    \centering
    \includegraphics[width=\textwidth]{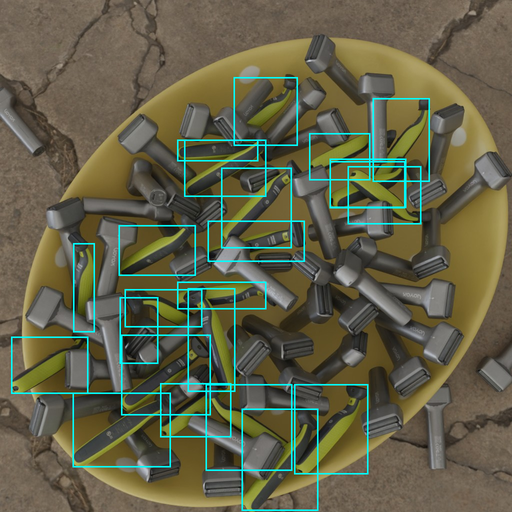}
    \vspace{-2.3ex}
    \caption*{{\scriptsize $N_{pred} = 23, N_{GT} = 21$} \\
               {\tiny ``green and black shaver''}}
  \end{subfigure} \hfill
  \begin{subfigure}{0.235\textwidth}
    \centering
    \includegraphics[width=\textwidth]{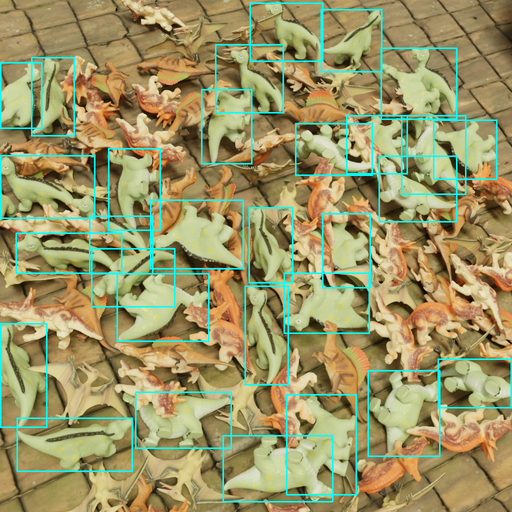}
    \vspace{-2.3ex}
    \caption*{{\scriptsize $N_{pred} = 30, N_{GT} = 32$} \\
               {\tiny ``green dinosaur toy''}}
  \end{subfigure} \hfill
  \begin{subfigure}{0.235\textwidth}
    \centering
    \includegraphics[width=\textwidth]{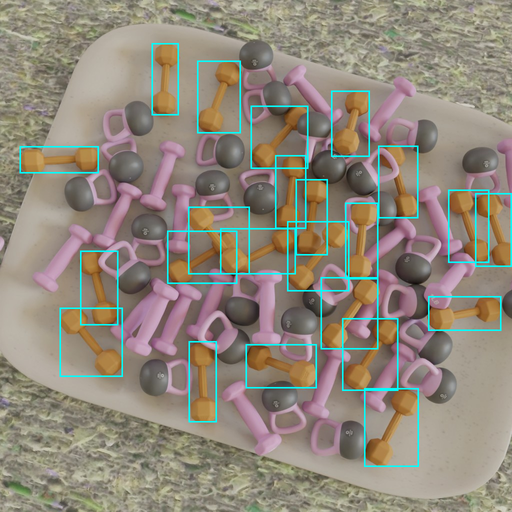}
    \vspace{-2.3ex}
    \caption*{{\scriptsize $N_{pred} = 23, N_{GT} = 23$} \\
               {\tiny ``tangerine hexagonal dumbbell''}}
  \end{subfigure} \hfill
  \begin{subfigure}{0.235\textwidth}
    \centering
    \includegraphics[width=\textwidth]{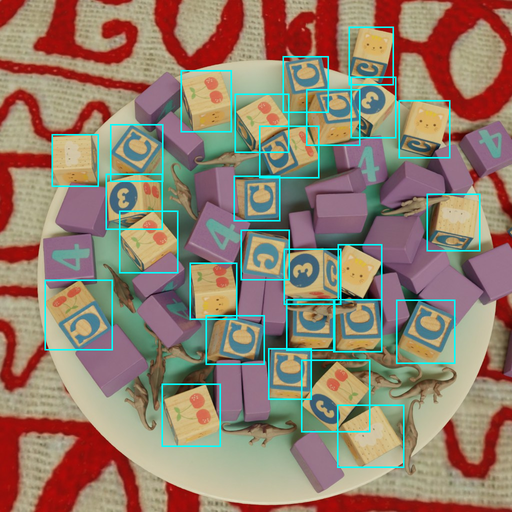}
    \vspace{-2.3ex}
    \caption*{{\scriptsize $N_{pred} = 27, N_{GT} = 28$} \\
               {\tiny ``wood block C 3 cherries''}}
  \end{subfigure} 
  \caption{\textbf{Additional test samples.} We display test samples from the \acron~dataset along with model predictions. We generate predictions with our best model, the CountGD++ model (+MixCount) with positive and negative prompts.}
  \label{fig:pred_samples}
\end{figure*}

%% file: figs/external_exemplars.tex
\begin{figure*}[htbp]
  \centering
  \captionsetup[subfigure]{justification=centering, skip=0pt}
  \begin{subfigure}{0.16\textwidth}
    \centering
    \includegraphics[width=\textwidth]{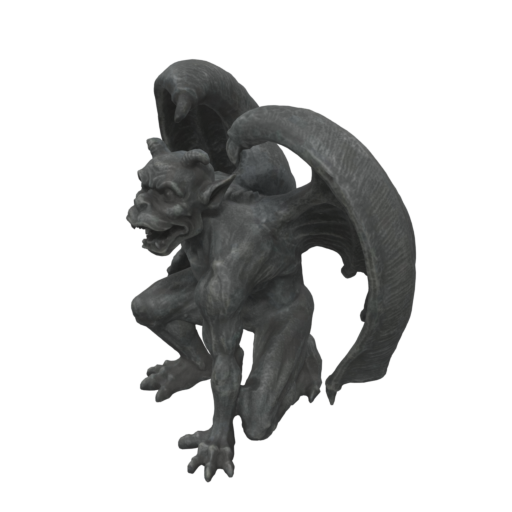}
    \vspace{-1.2ex}
    \caption*{{\tiny ``Kneeling winged gargoyle statue''}}
  \end{subfigure} \hfill
  \begin{subfigure}{0.16\textwidth}
    \centering
    \includegraphics[width=\textwidth]{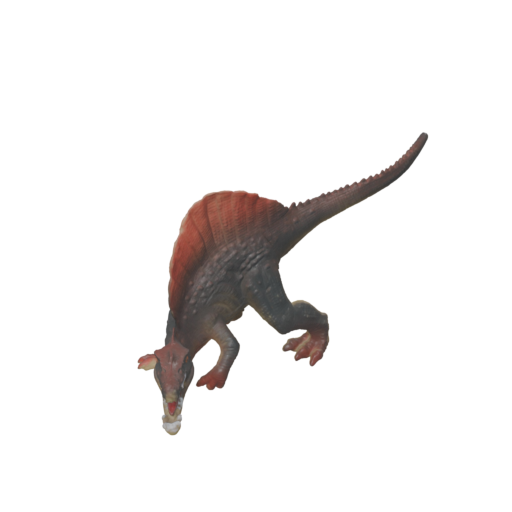}
    \vspace{-1.2ex}
    \caption*{{\tiny ``Orange-grey Spinosaurus toy''}}
  \end{subfigure} \hfill
  \begin{subfigure}{0.16\textwidth}
    \centering
    \includegraphics[width=\textwidth]{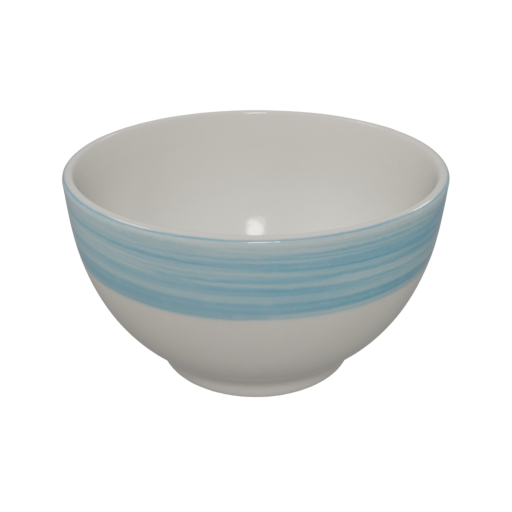}
    \vspace{-1.2ex}
    \caption*{{\tiny ``White light blue striped bowl''}}
  \end{subfigure} \hfill
  \begin{subfigure}{0.16\textwidth}
    \centering
    \includegraphics[width=\textwidth]{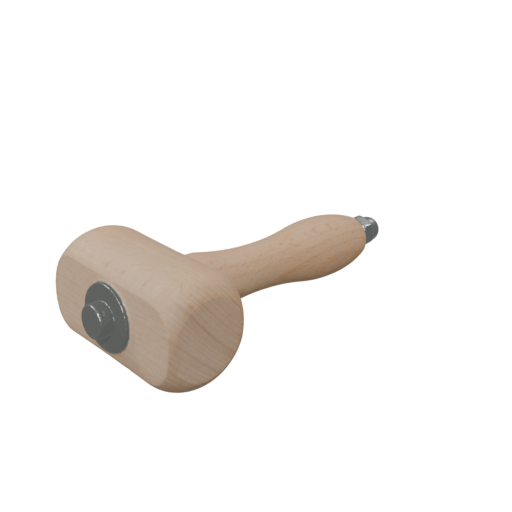}
    \vspace{-1.2ex}
    \caption*{{\tiny ``wooden mallet''}}
  \end{subfigure} \hfill
  \begin{subfigure}{0.16\textwidth}
    \centering
    \includegraphics[width=\textwidth]{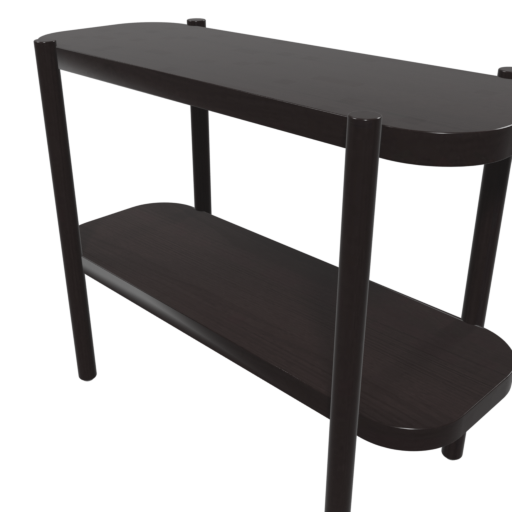}
    \vspace{-1.2ex}
    \caption*{{\tiny ``Dark wood two-tier table''}}
  \end{subfigure} \hfill
  \begin{subfigure}{0.16\textwidth}
    \centering
    \includegraphics[width=\textwidth]{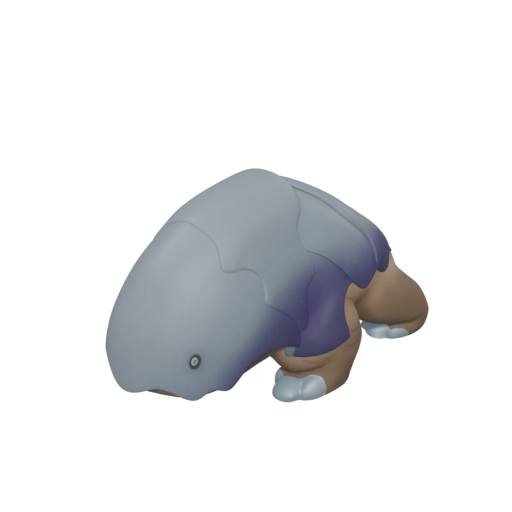}
    \vspace{-1.2ex}
    \caption*{{\tiny ``Armored quadrupedal creature''}}
  \end{subfigure} \\[1.5ex]
  \begin{subfigure}{0.16\textwidth}
    \centering
    \includegraphics[width=\textwidth]{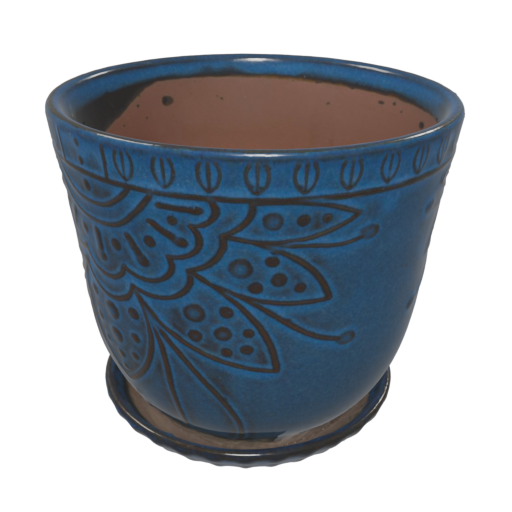}
    \vspace{-1.2ex}
    \caption*{{\tiny ``Patterned blue ceramic planter''}}
  \end{subfigure} \hfill
  \begin{subfigure}{0.16\textwidth}
    \centering
    \includegraphics[width=\textwidth]{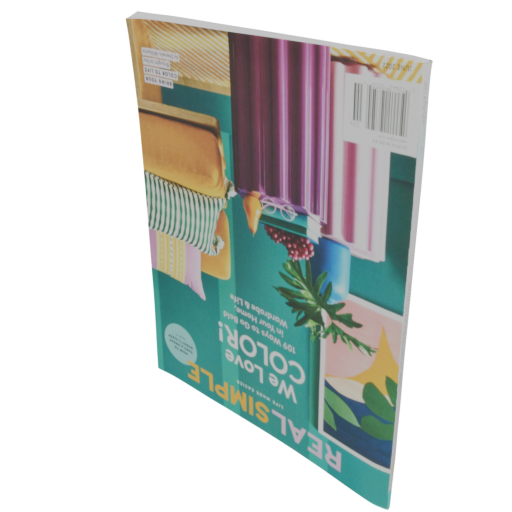}
    \vspace{-1.2ex}
    \caption*{{\tiny ``colorful Real Simple magazine''}}
  \end{subfigure} \hfill
  \begin{subfigure}{0.16\textwidth}
    \centering
    \includegraphics[width=\textwidth]{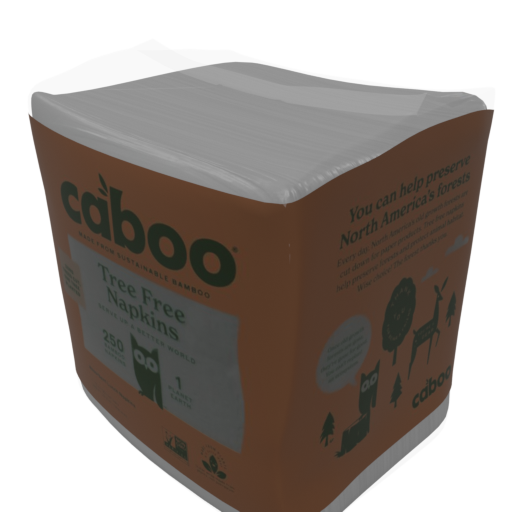}
    \vspace{-1.2ex}
    \caption*{{\tiny ``Caboo bamboo napkins package''}}
  \end{subfigure} \hfill
  \begin{subfigure}{0.16\textwidth}
    \centering
    \includegraphics[width=\textwidth]{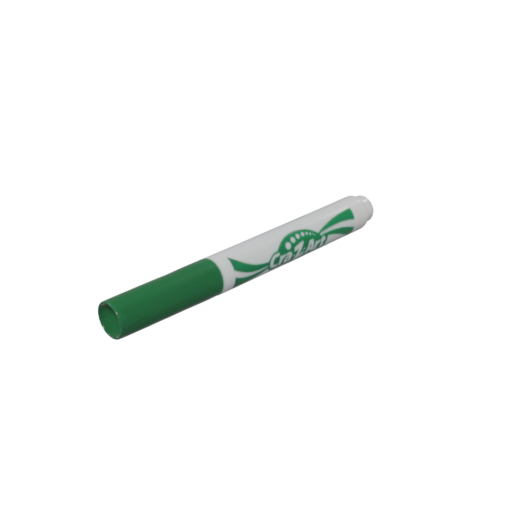}
    \vspace{-1.2ex}
    \caption*{{\tiny ``Green and white marker''}}
  \end{subfigure} \hfill
  \begin{subfigure}{0.16\textwidth}
    \centering
    \includegraphics[width=\textwidth]{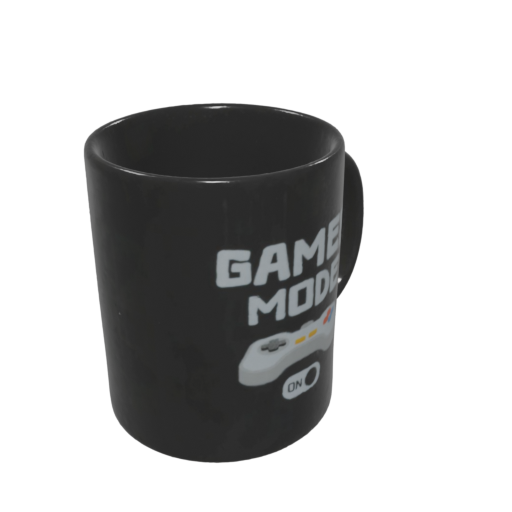}
    \vspace{-1.2ex}
    \caption*{{\tiny ``Black mug game mode''}}
  \end{subfigure} \hfill
  \begin{subfigure}{0.16\textwidth}
    \centering
    \includegraphics[width=\textwidth]{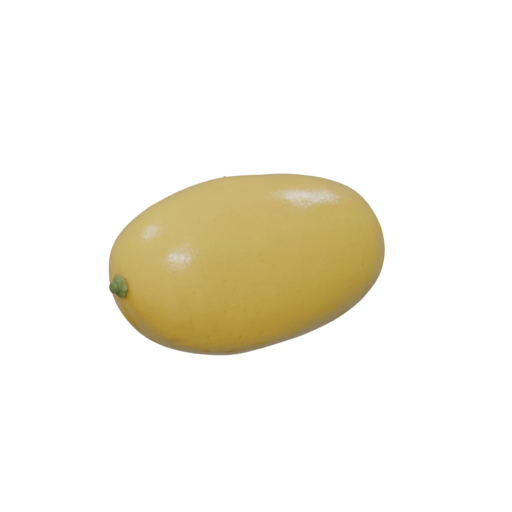}
    \vspace{-1.2ex}
    \caption*{{\tiny ``Yellow mango''}}
  \end{subfigure} \\[1.5ex]
  \begin{subfigure}{0.16\textwidth}
    \centering
    \includegraphics[width=\textwidth]{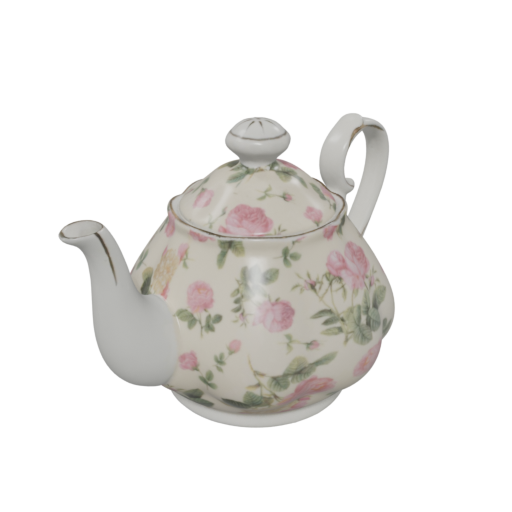}
    \vspace{-1.2ex}
    \caption*{{\tiny ``Floral patterned ceramic teapot''}}
  \end{subfigure} \hfill
  \begin{subfigure}{0.16\textwidth}
    \centering
    \includegraphics[width=\textwidth]{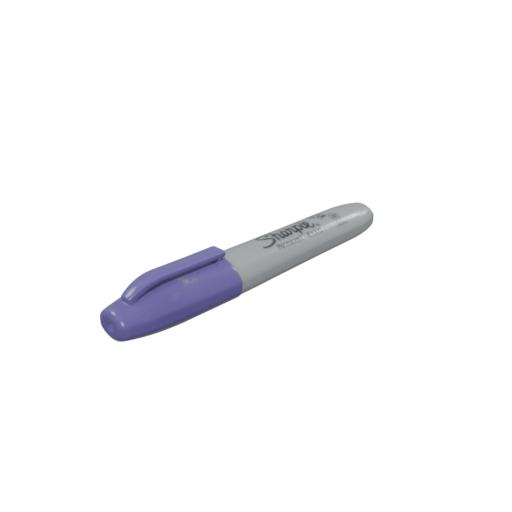}
    \vspace{-1.2ex}
    \caption*{{\tiny ``purple Sharpie marker''}}
  \end{subfigure} \hfill
  \begin{subfigure}{0.16\textwidth}
    \centering
    \includegraphics[width=\textwidth]{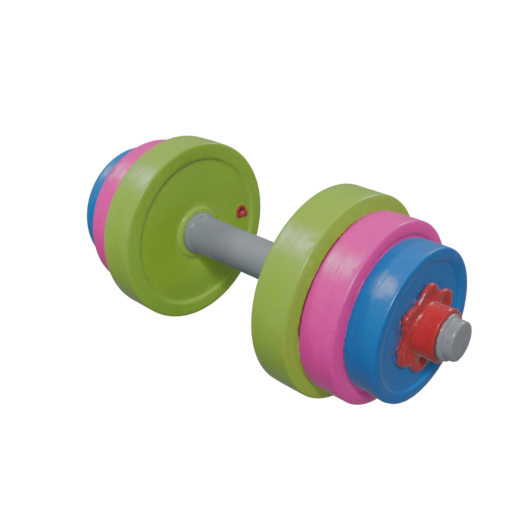}
    \vspace{-1.2ex}
    \caption*{{\tiny ``Colorful plastic dumbbell''}}
  \end{subfigure} \hfill
  \begin{subfigure}{0.16\textwidth}
    \centering
    \includegraphics[width=\textwidth]{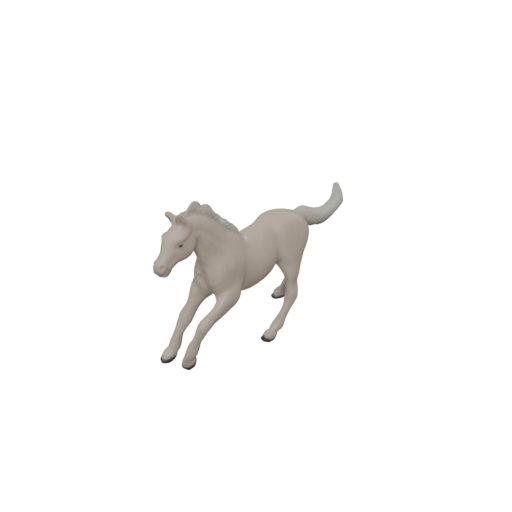}
    \vspace{-1.2ex}
    \caption*{{\tiny ``white horse figurine running''}}
  \end{subfigure} \hfill
  \begin{subfigure}{0.16\textwidth}
    \centering
    \includegraphics[width=\textwidth]{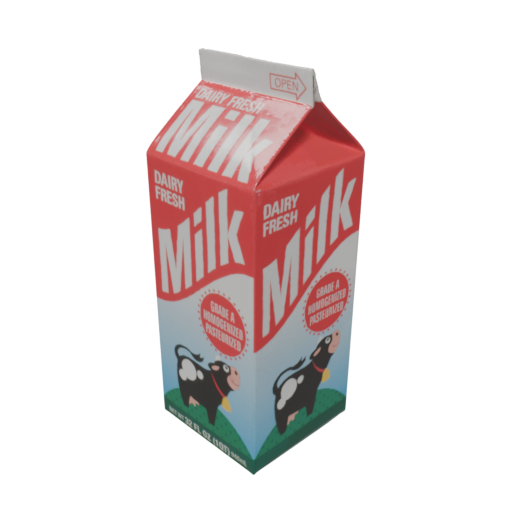}
    \vspace{-1.2ex}
    \caption*{{\tiny ``Toy milk carton''}}
  \end{subfigure} \hfill
  \begin{subfigure}{0.16\textwidth}
    \centering
    \includegraphics[width=\textwidth]{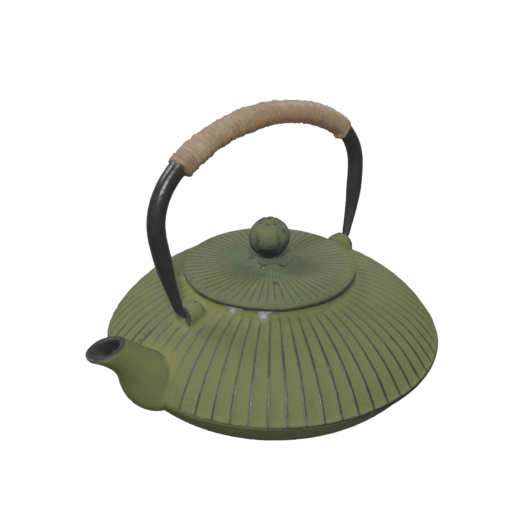}
    \vspace{-1.2ex}
    \caption*{{\tiny ``Green ribbed cast iron teapot''}}
  \end{subfigure} \\[1.5ex]
  \caption{\textbf{External exemplars.} Additional external exemplars are provided by rendering a canonical view of each object in front of a white background. These exemplars are used to evaluate the sensitivity of counting models to the context of the provided visual exemplar.}
  \label{fig:external_exemplars}
\end{figure*}